\documentclass[10pt,twocolumn,letterpaper]{article}

\usepackage[pagenumbers]{wacv} %

\usepackage{graphicx}
\usepackage{amsmath}
\usepackage{amssymb}
\usepackage{booktabs}
\usepackage{mathtools}

\usepackage{multirow}
\usepackage{subcaption}

\usepackage{mathtools}
\usepackage{booktabs}
\usepackage{colortbl}
\usepackage[accsupp]{axessibility}

\usepackage{cuted}
\usepackage{afterpage}

\usepackage[normalem]{ulem}
\useunder{\uline}{\ul}{}
\usepackage{xcolor}
\definecolor{ForestGreen}{RGB}{34,139,34}
\definecolor{pinkcarino}{RGB}{255,0,255}

\newcommand{\giulia}[1]{{\color{black}{#1}}}

\newcommand{\pietro}[1]{{\color{black}{#1}}}

\usepackage{amsfonts}
\usepackage{pifont}
\newcommand{\cmark}{\ding{51}}%

\usepackage{adjustbox}
\usepackage{arydshln}

\usepackage[pagebackref,breaklinks,colorlinks]{hyperref}

\usepackage[capitalize]{cleveref}
\crefname{section}{Sec.}{Secs.}
\Crefname{section}{Section}{Sections}
\Crefname{table}{Table}{Tables}
\crefname{table}{Tab.}{Tabs.}

\def\wacvPaperID{16} %
\def\confName{WACV}
\def\confYear{2024}

\begin{document}

\title{Source-Free Domain Adaptation for RGB-D Semantic\\
Segmentation with Vision Transformers}

\author{Giulia Rizzoli \qquad Donald Shenaj \qquad Pietro Zanuttigh \\
University of Padova, Italy  \\
}

\maketitle

\begin{abstract}
    With the increasing availability of depth sensors, multimodal frameworks that combine color information with depth data are gaining interest. However, ground truth data for semantic segmentation is burdensome to provide, thus making domain adaptation a significant research area. Yet most domain adaptation methods are not able to effectively handle multimodal data.  
    Specifically, we address the challenging source-free domain adaptation setting where the adaptation is performed without reusing source data. We propose \textbf{MISFIT}: \textbf{\underline{M}ult\underline{I}modal \underline{S}ource-\underline{F}ree \underline{I}nformation fusion \underline{T}ransformer}, a depth-aware framework which injects depth data into a segmentation module based on vision transformers at multiple stages, namely at the input, feature and output levels. Color and depth style transfer helps early-stage domain alignment while re-wiring self-attention between modalities creates mixed features, allowing the extraction of better semantic content. Furthermore, a depth-based entropy minimization strategy is also proposed to adaptively weight regions at different distances.
    Our framework, which is also the first approach using RGB-D vision transformers for source-free semantic segmentation, shows noticeable performance improvements with respect to standard strategies.
\end{abstract}

\section{Introduction}
\label{sec:intro}

Semantic segmentation has traditionally been performed employing RGB images, which solely capture color information. Yet, as depth sensors become more widely available, multimodal frameworks that integrate RGB visuals with depth information have emerged \cite{rizzoli2022multimodal}.
This integration offers the potential for improved semantic segmentation performance due to the additional clues provided by depth. 
Depth information proves particularly beneficial in several scenarios, including distinguishing between objects with similar colors but different distances, as well as aiding the segmentation of objects with complex geometries. 
\begin{figure}[t!]
    \centering
    \includegraphics[width=0.5\textwidth]{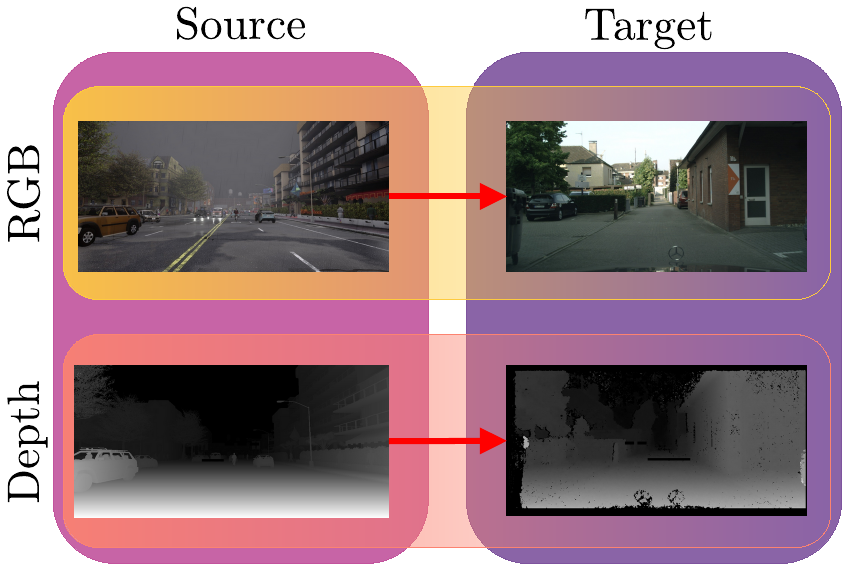} 
    \caption{RGB-D Domain Adaptation: (a) Color domain shift, where variations in lighting conditions, color distribution, and texture affect the RGB images; (b) Depth domain shift, where differences in depth estimation strategies, surface geometry, and objects' scale impact the depth maps.}
    \label{fig:source_vs_target}
\end{figure}
Although state-of-the-art approaches achieve good results on several benchmarks, most multimodal methods do not test the generalization capabilities of the model \cite{valada2020self,liu2022cmx,barbato2022depthformer}.
Domain adaptation is especially crucial in real-world scenarios where environments and conditions may vary, making it challenging to train a single model that performs well across all domains \cite{toldo2020unsupervised}. 
The majority of domain adaptation methods work with color data alone \cite{huang2021model,kundu2021generalize,you2021domain} or focus on the Unsupervised Domain Adaptation (UDA) setting. %
\pietro{However, a more challenging yet realistic setting is Source Free Domain Adaptation, where the pre-trained model undergoes adaptation without accessing the source data, as opposed to the typical joint source supervised and target unsupervised training approach in standard UDA. In this scenario, jointly using color and depth data for pre-training is unexplored.}
Given the recent studies demonstrating the generalization \cite{hoyer2022daformer,park2022dat} and multimodal processing \cite{liu2022cmx,barbato2022depthformer} capabilities of transformer architectures, in this paper we explore the feasibility of employing a transformer architecture for source-free domain adaptation.
Furthermore, we would like to exploit the potential of depth data in guiding the adaptation process.
To this extent, we propose our method, MultImodal Source-Free Information fusion Transformer (MISFIT), which includes the following contributions:
\begin{enumerate}
    \item The introduction of the first RGB-D framework for source-free domain adaptation semantic segmentation, exploiting vision transformers;
    \item The exploration of input-level depth stylization \pietro{in source pre-training} using a fast and simple approach operating in the frequency domain;
    \item The evaluation of  generalization capabilities of RGB-D attention fusion within transformer architectures;
    \item The development of a novel approach that leverages depth data in a self-teaching optimization scheme for source-free domain adaptation.
\end{enumerate}

The proposed approach tackles the multimodal source-free domain adaptation task by introducing several provisions into a vision transformer architecture for semantic segmentation. \giulia{The method involves distinct stages for both pre-training and adaptation. In the preliminary pre-training phase, we employ a domain stylization to tailor the input data, thereby enhancing the adaptability of the model. 
Within the internal network representation, we introduce modifications to the attention module of the transformer to handle the multimodal nature of the data effectively in both stages. %
Finally, during adaptation, our approach integrates a depth-guided self-teaching strategy to refine the segmentation results.}
We validated it on standard RGB-D benchmarks and the employed provisions allow to tackle the source-free domain adaptation task effectively.

After discussing the related works in Section \ref{sec:relworks}, we will introduce the main components of our method in Section \ref{sec:method}, detailing the input (Section \ref{sec:input_level}), feature (Section \ref{sec:output_level}) and output (Section \ref{sec:output_level}) level provisions. Finally, we present the experimental results and ablation studies in Section \ref{sec:results} and draw the conclusions (Section \ref{sec:conclusions}).

\section{Related Works}
\label{sec:relworks}

\textbf{Multimodal  Semantic Segmentation}
Recent studies have highlighted the potential of additional representations, such as depth and thermal data, in extracting semantic cues \cite{valada2020self,seichter2021efficient}. Early multimodal segmentation techniques involved combining RGB data with other modalities into multi-channel representations, which were then fed into standard semantic segmentation networks \cite{rizzoli2022multimodal}. %
This simple fusion strategy fails to comprehensively capture the varied information conveyed by each modality. To address this limitation, current methods employ various fusion strategies at different levels in the deep network. These approaches typically rely on a multi-stream encoder with a network branch for each modality, along with additional network modules that combine modality-specific features into fused ones and carry information across branches  \cite{liu2022cmx,barbato2022depthformer,zhang2023delivering}.

\textbf{Transformer-based Adaptation for Semantic Segmentation}
Several studies investigated the potential of transformers for semantic segmentation in unsupervised domain adaptation (UDA) settings \cite{hoyer2022daformer, park2022dat}. %
They showed the generalization potential of Transformers architectures \cite{kolesnikov2021image,xie2021segformer} compared to the widely used convolutional neural networks.
Hoyer et al. \cite{hoyer2022hrda} propose a multi-resolution training approach for UDA to preserve fine segmentation details and capture long-range context dependencies.
Park et al. \cite{park2022dat} apply an entropy-based re-weighting in the attention module to address domain discrepancy.
Although these methods show the applicability of vision transformers in the UDA scenario, none of them investigates adaptation in the source-free setting data nor domain adaptation with multimodal data.

\textbf{Multimodal Domain Adaptation for Semantic Segmentation}
Hu et al. \cite{hu2023multi} addresses a single-stage input-level fusion, summing the depth after being injected into one attention block. %
xMUDA \cite{jaritz2022cross} proposes an unsupervised domain adaptation scheme for 3D semantic segmentation where the output feature of two distinct networks (2D for RGB and 3D for LiDAR) are fused through mutual mimicking. 
In MM-TTA \cite{shin2022mm}, they investigate the challenge of test-time adaptation for multi-modal 3D semantic segmentation.
\giulia{Due to the additional domain shift introduced by pre-training a model on depth, there is no currently existing framework for source-free domain adaptation exploiting depth on the task of 2D semantic segmentation.}

\textbf{Unsupervised vs Source-Free Domain Adaptation}
In the challenging domain of Source-free domain adaptation (SFDA), the availability of source domain data is limited to an initial pre-training stage, while the subsequent adaptation process relies solely on unlabeled target data. Domain adaptation methods can be classified into two categories: data-level approaches and model-level approaches \cite{yu2023comprehensive}.
Data-level approaches aim to mitigate the domain shift by manipulating the target data to resemble the source domain. This involves aligning various aspects such as the imaging style in the input space \cite{yang2020fda,hoffman2018cycada,tranheden2021dacs}, feature space \cite{luo2019taking,pan2020unsupervised}, output space \cite{tsai2019domain}. 
However, SFDA presents a greater challenge as domain alignment must be achieved without access to the source dataset, making traditional adversarial learning methods used in UDA unsuitable.
On the other hand, \textit{model-level} approaches for domain adaptation include self-training, where the model is used to generate pseudo-labels for the data from the unlabeled target domain \cite{yu2023comprehensive,zou2018unsupervised,zou2019confidence}. %
In standard UDA, entropy minimization approaches enhance the quality of pseudo-labels by minimizing the entropy of the target data or using the entropy map as input for a domain discriminator with an adversarial learning strategy \cite{vu2018advent, chen2019domain}. However, in the absence of labeled source domain data in the source data-free setting, the effectiveness of entropy minimization-based methods can be compromised.

\textbf{Source-Free Domain Adaptation}
Apart from standard domain adaptation, more recent works, investigate unimodal source-free adaptation. Liu et al. \cite{liu2021source} leverage self-supervised learning to learn representations that are robust to domain shift and a knowledge distillation loss function is used to align the representations of the source and target domains. Fleuret et al. \cite{fleuret2021uncertainty} exploit posterior probabilities to estimate uncertainty in the adaptation process.
Huang et al. \cite{huang2021model} uses contrastive category discrimination on pseudo-labels target samples to learn category-discriminative representations.
You et al. \cite{you2021domain} adopted a positive-negative learning strategy in combination with intra-class pseudo-labels thresholding.
Kundu et al. \cite{kundu2021generalize} employ several encoder-level heads which are further pruned to select the optimal one. 
Ye et al. \cite{ye2021source} employ uncertainty and prior distribution-aware domain adaptation techniques, incorporating both adversarial learning and self-training strategies, to create a set of virtual source domain data.
Yang et al. \cite{yang2022source} propose a weight-regularized distribution transfer method, followed by class-balanced thresholding and multi-class negative techniques during the adaptation phase.
Zhao et al. \cite{zhao2023towards} introduce a dynamic teacher update mechanism and a resampling strategy based on training consistency.
Furthermore, to tackle diverse practical contexts, some approaches integrate source-free domain adaption with federated learning \cite{shenaj2023learning}, black box test \cite{peng2022toward}, or robust transfer \cite{agarwal2022unsupervised}.

\begin{figure*}
    \centering
    \includegraphics[width=0.9\textwidth]{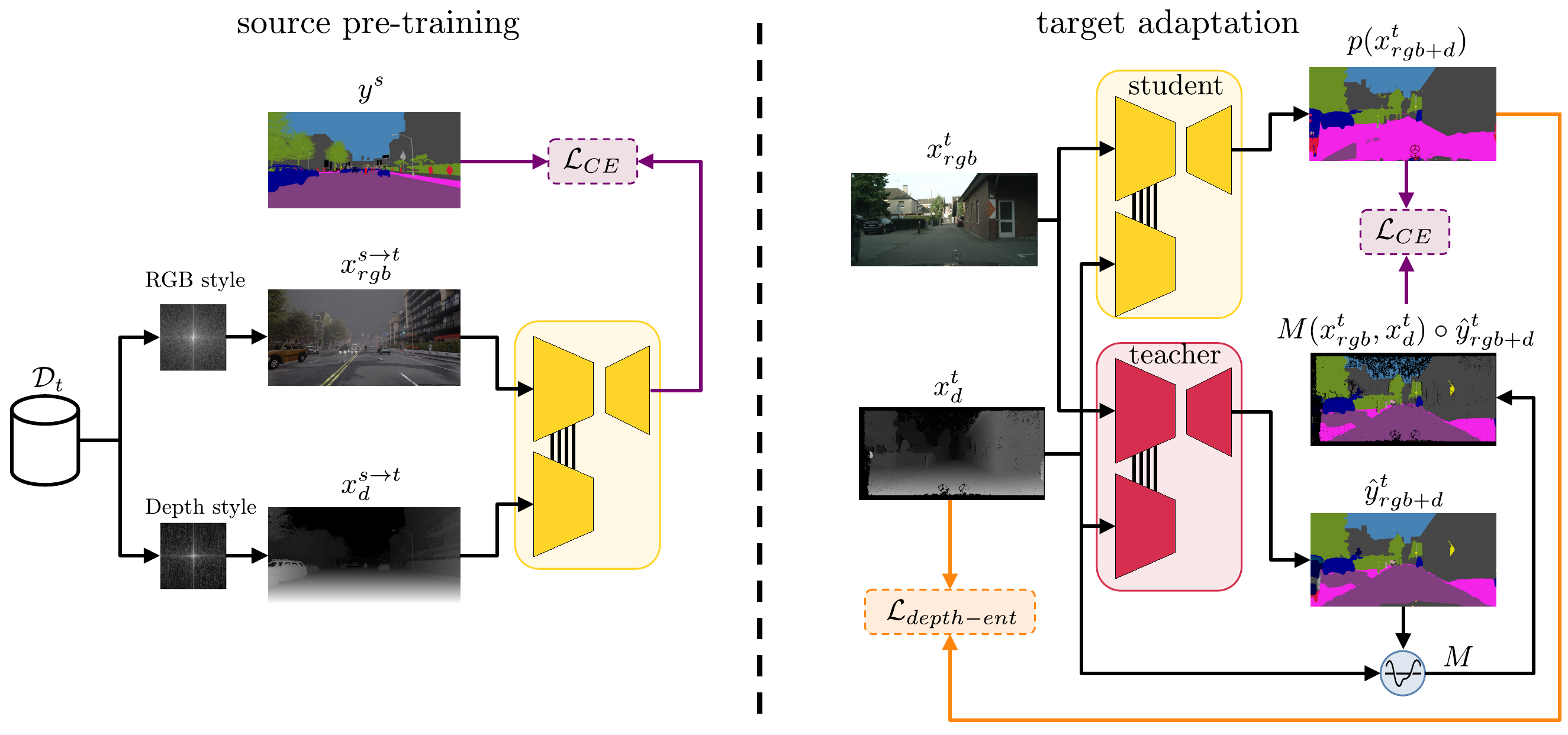}
    \caption{Overview of the training procedure for the proposed method. First, the network is trained on the (synthetic) supervised source dataset, while style transfer is applied both on RGB and depth images (Sec.~\ref{sec:input_level}). Then, the model is trained on the (real) unsupervised target dataset via the masked self-training strategy and depth entropy minimization (Sec.~\ref{sec:output_level}).
    In both steps, the fusion between RGB and depth features is performed with cross-modality attention (Sec.~\ref{sec:feat_adap}).}
    \label{fig:misfit}
\end{figure*}
\section{Method}
\label{sec:method}

In this section, we introduce the three main strategies that we adopt for source-free domain adaptation, organizing them according to the stage at which they are employed: at the input level we exploited  style transfer during pre-training (Section
 \ref{sec:input_level}); at the feature level we tackle the multimodal setting by exchanging information in the attention module of the transformer (Section \ref{sec:feat_adap}); at the output level a depth-based self-teaching strategy is used for domain adaptation (Section \ref{sec:output_level}).  An overview of the framework is shown in Figure \ref{fig:misfit}.

We denote the labeled source domain data samples as $\mathcal{D}_{s} = {((x_{rgb}^{s}, x_{d}^{s}), y^{s})}$, where $y^{s}$ is the  label corresponding to the multimodal input $(x_{rgb}^{s}, x_{d}^{s})$ (as expected $x_{rgb}^{s}$ is the color image and $x_{d}^{s}$ the corresponding depth map). The target domain is unlabeled and drawn from the distribution $\mathcal{D}_{t} = {(x_{rgb}^{t}, x_{d}^{t})}$.  In the source-free setting, we assume that $\mathcal{D}_{s}$ is only available during model pre-training. The target is to assign to each pixel one of the $C$ possible classes. The transformer architecture is constituted by a multi-head attention, which constitutes the encoder part, and a segmentation decoder, in particular, we employed SegFormer \cite{xie2021segformer} as the starting architecture.
The probability output of the network is denoted as $p(x) \coloneqq p \in \mathcal{R}^{C}$.

\begin{figure}[h!]
\begin{subfigure}[b]{0.5\textwidth}
\centering
\begin{subfigure}[b]{0.22\textwidth}
\caption*{Source}
\includegraphics[width=\textwidth]{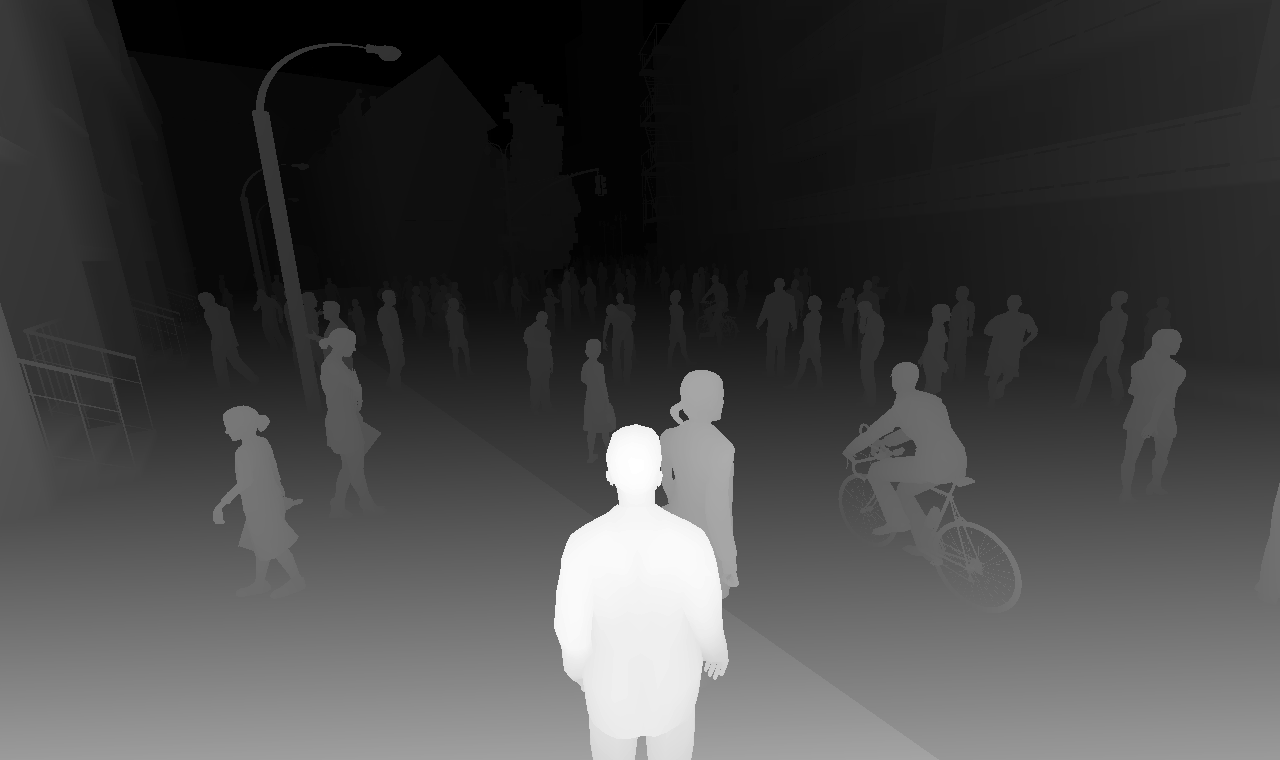}
\end{subfigure} 
\begin{subfigure}[b]{0.22\textwidth}
\caption*{$\beta = 0.01$}
\includegraphics[width=\textwidth]{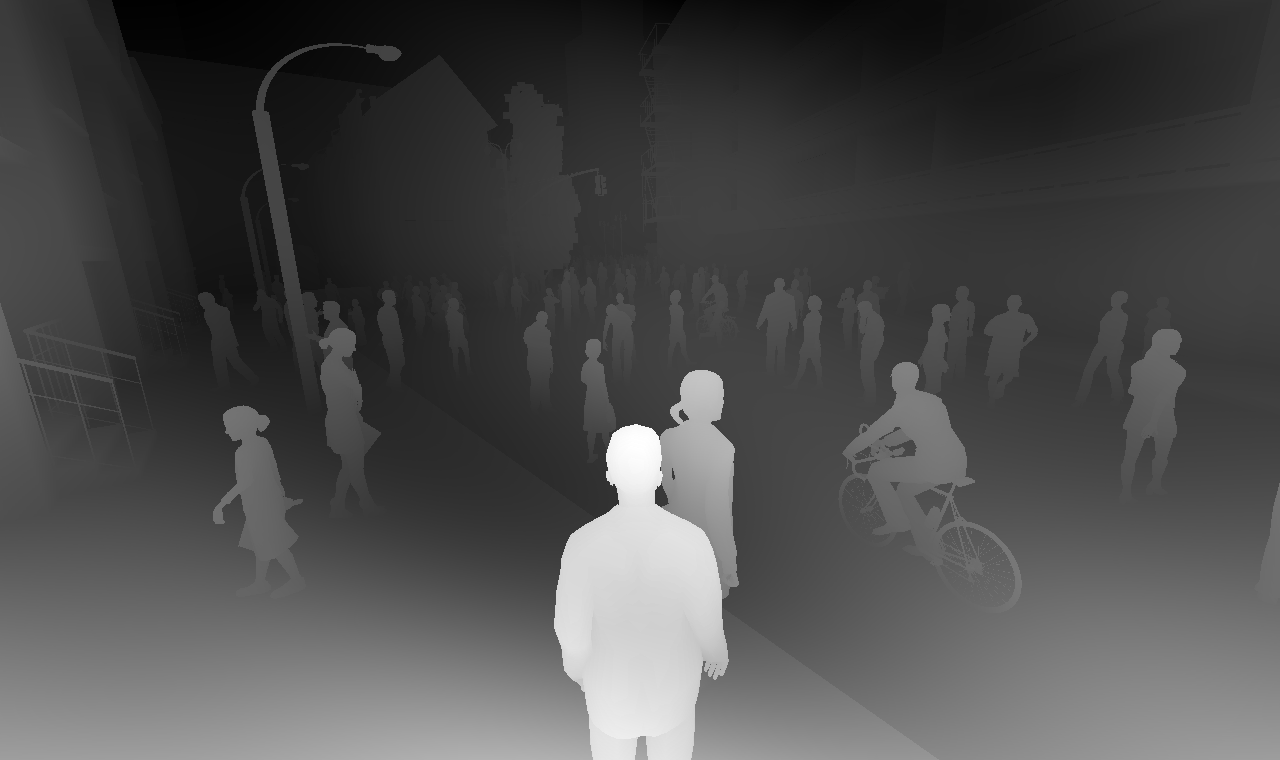}
\end{subfigure} 
\begin{subfigure}[b]{0.22\textwidth}
\caption*{$\beta = 0.03$}
\includegraphics[width=\textwidth]{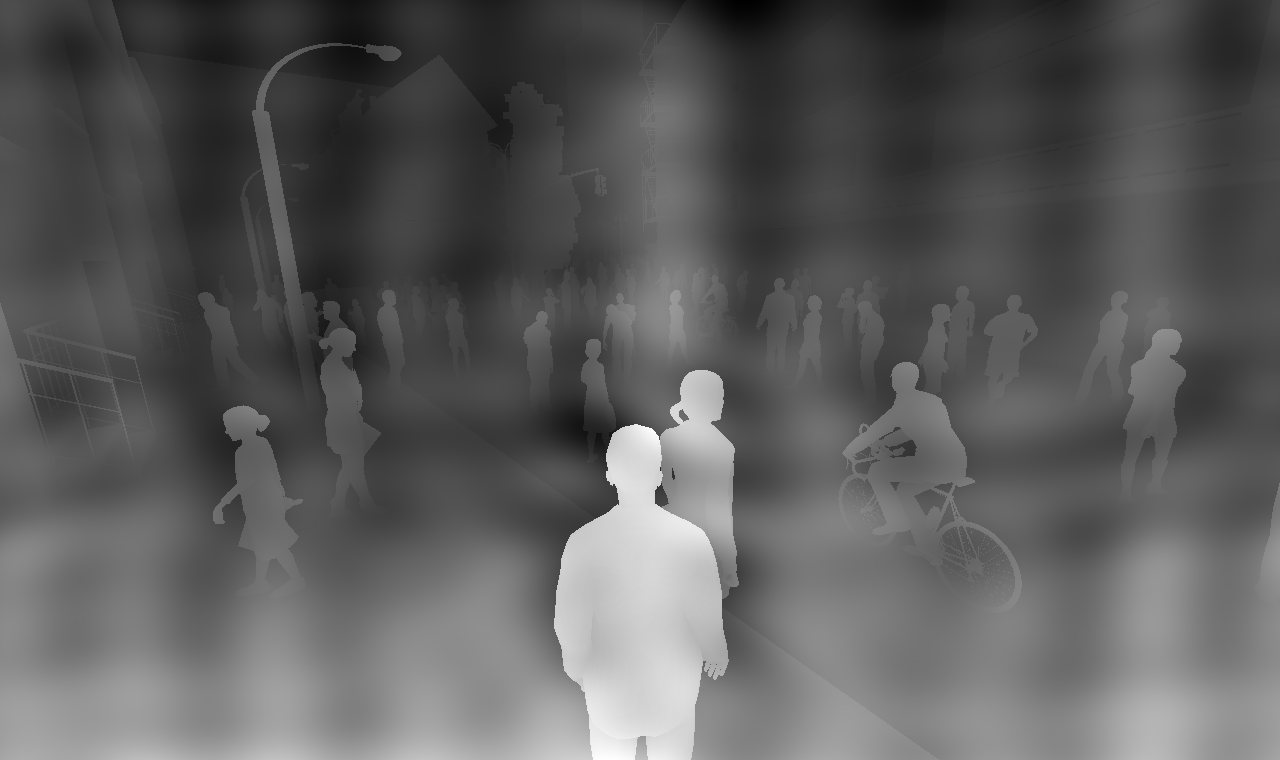}
\end{subfigure}
\begin{subfigure}[b]{0.22\textwidth}
\caption*{$\beta = 0.05$}
\includegraphics[width=\textwidth]{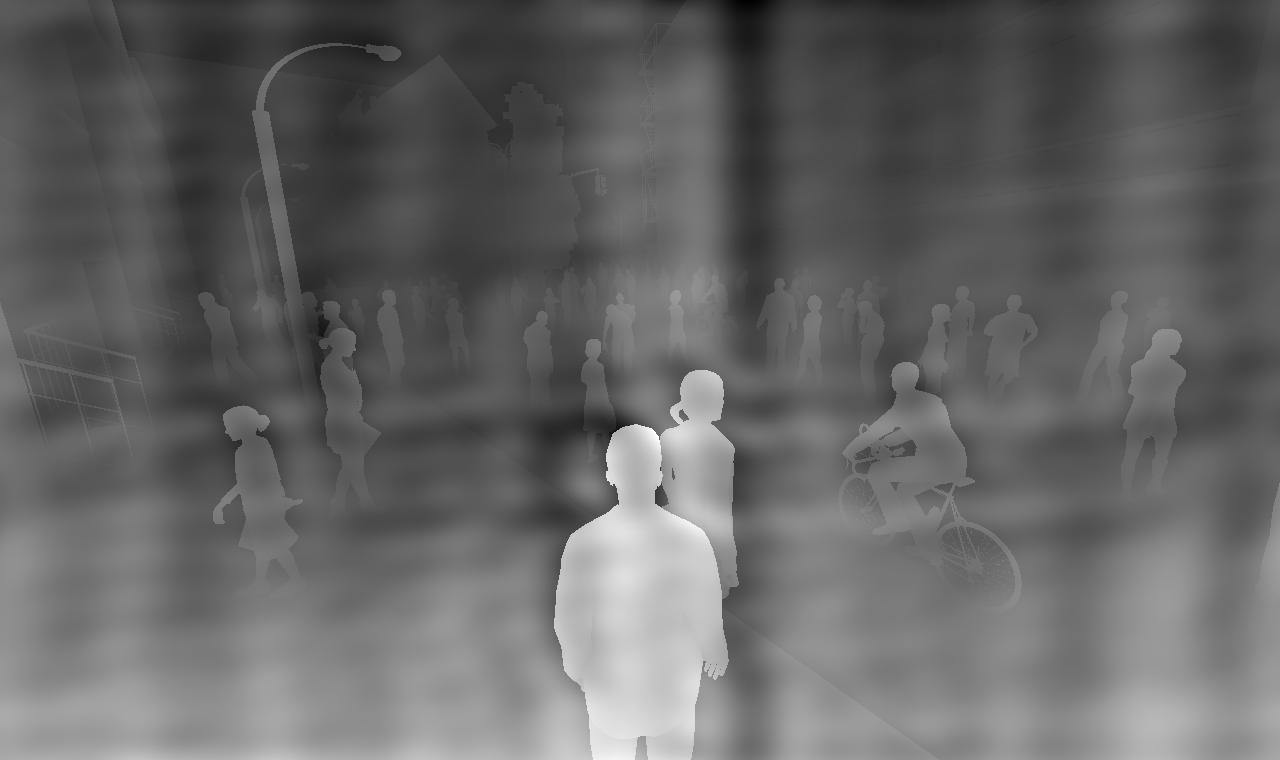}
\end{subfigure}
\end{subfigure}

\begin{subfigure}[b]{0.5\textwidth}
\centering
\begin{subfigure}[b]{0.22\textwidth}
\includegraphics[width=\textwidth]{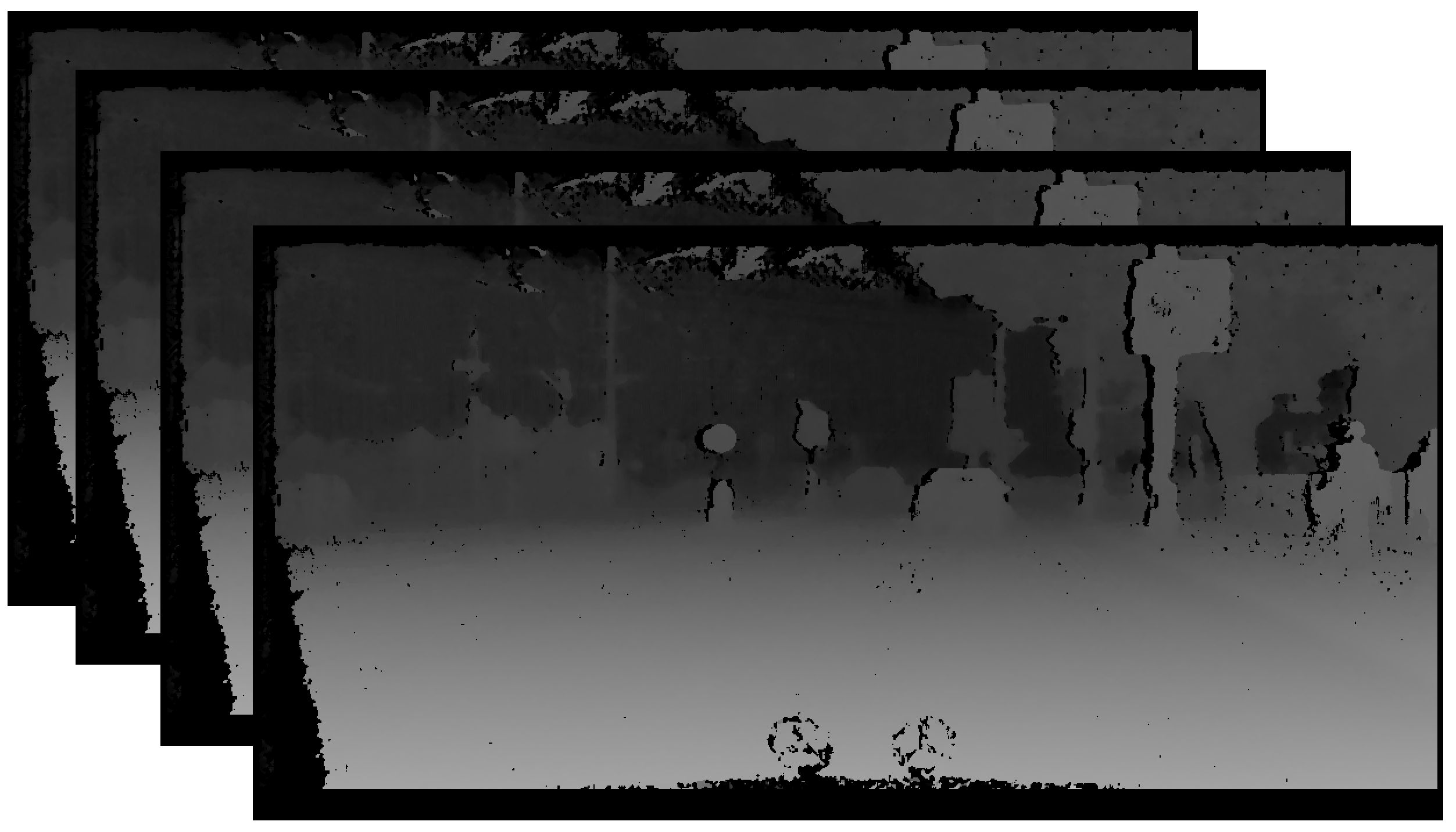}
\caption*{Target} %
\end{subfigure} 
\begin{subfigure}[b]{0.22\textwidth}
\includegraphics[width=\textwidth]{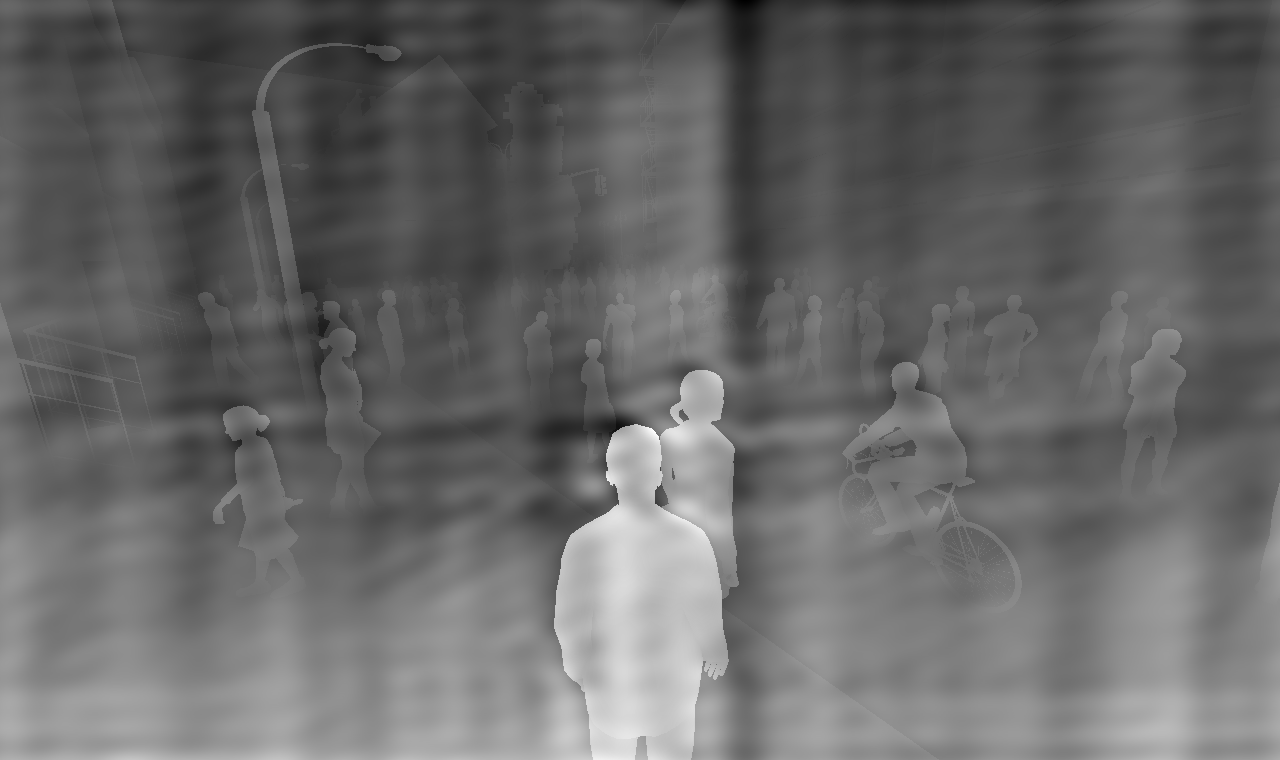}
\caption*{$\beta = 0.07$}
\end{subfigure} 
\begin{subfigure}[b]{0.22\textwidth}
\includegraphics[width=\textwidth]{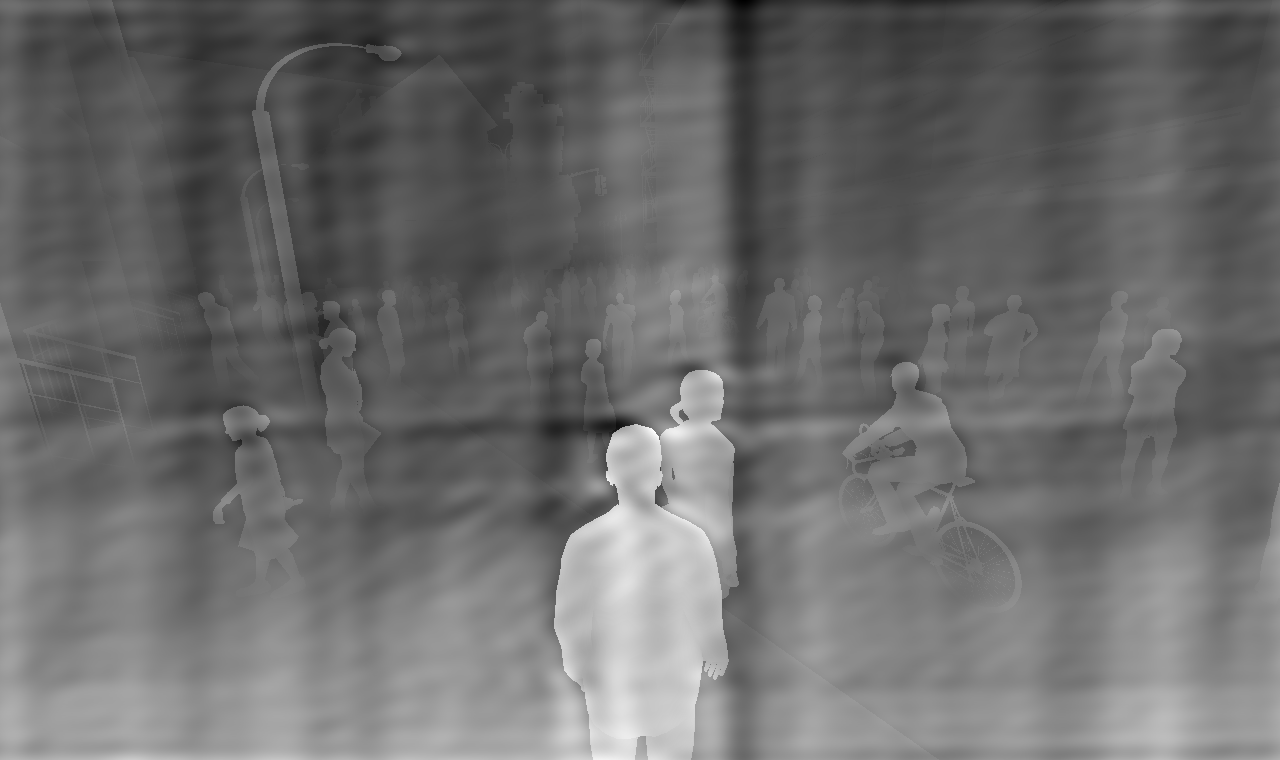}
\caption*{$\beta = 0.09$}
\end{subfigure}
\begin{subfigure}[b]{0.22\textwidth}
\includegraphics[width=\textwidth]{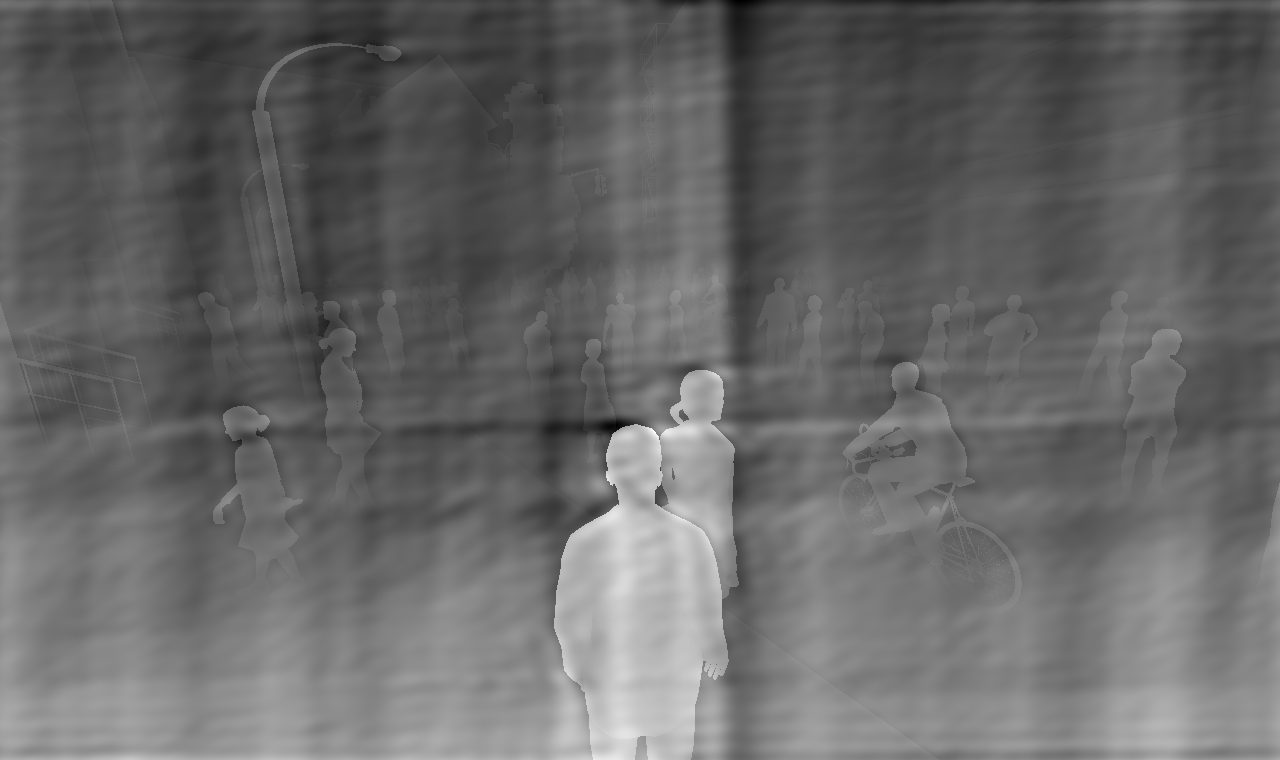}
\caption*{$\beta = 0.12$}
\end{subfigure}
\end{subfigure}
\caption{Effect of the Fourier domain style transfer applied on depth images, whereas $\beta$ = 0 is equivalent to no transfer and $\beta$ = 1 to the transfer of the full target amplitude.}
\label{fig:fda_synthia}
\end{figure}

\subsection{Input Level Pre-training Adaptation} %
\label{sec:input_level}
Style transfer techniques allow the alignment of the visual appearance of samples from the source to the target domain, thus increasing the model's generalization capabilities. These techniques have been widely used on color data, but their applicability to depth representations has not been explored.

In our setting, the segmentation network is pre-trained by applying image-to-image translation on both the color and depth data of the input samples from $\mathcal{D}_{s}$. 
In particular, we opted for a frequency domain image translation algorithm for style transfer, i.e., FDA \cite{yang2020fda}, preserving the advantage of using a simple module that does not require training complex adversarial deep networks modules for image translation. This allows for avoiding additional computational complexity at inference time and for keeping simpler the training procedure.
We retrieve the frequency space representation of a sample $x_{i} \in \mathbb{R}^{HxWxC}$ through Fast Fourier Transform (FFT) \cite{yang2020fda} as:
\begin{equation}
    \mathcal{F}(x_{i})(u,v,c) = \sum_{h=0}^{H-1} \sum_{w=0}^{W-1} x_{i}(h,w,c) e^{-j2\pi(\frac{h}{H}u + \frac{w}{W}v)}
\end{equation}
The frequency space signal $\mathcal{F}(x_{i})$ can be decomposed into an amplitude spectrum $\mathcal{A}(x_{i}) \in \mathbb{R}^{HxWxC}$ and a phase angle $\mathcal{P}(x_{i}) \in \mathbb{R}^{HxWxC}$. Low-level distributions reflect the image style, thus replacing lower frequencies in the source spectrum with the target ones  - in our case we used the average of a small set of target samples - can improve the domain adaptation performances: %
\begin{equation}
    x^{s \rightarrow t} = \mathcal{F}^{-1}(\mathcal{A}_{low}(x^{t}) + \mathcal{A}_{high}(x^{s}), \mathcal{P}(x^{s}))
\end{equation}
where $x$ can be both the color image $x_{rgb}$ or the depth map $x_{d}$. The fraction of the replaced low-level details is set by the parameter $\beta$, which controls the amplitude window.

In particular, different choices of $\beta$ affect the source representation: a larger $\beta$ increases the domain translation effect but also introduces visual artifacts.

The predicted semantics should not be influenced by the sensor's properties or other causes of variation linked to the acquisition procedure. Yet the generalization ability of the network is affected by these aspects \cite{yang2020fda}. Hence, in domain adaptation settings, perceptually minor changes in the low-level data might result in a considerable decline in the trained model's performance. Depth maps are influenced not only by the acquisition sensor, which can be based on completely different technologies, e.g., time-of-flight, active or passive-stereo, etc., leading to very different frequency responses, but also by the characteristics of the scene and by the camera viewpoint.
Performing alignment of low-frequency coefficients allows getting a better invariance to the characteristics of the sensors and to the depth values distribution due to the employed camera.
Furthermore, when performing synthetic-to-real adaptation the difference between noise-free ideal synthetic depth and the inaccurate data of most real-world depth acquisition strategies is another critical challenge. It can be mitigated by the frequency domain adaptation forcing the network to focus more on the semantic structure of the scene than on acquisition device peculiarities.

As we can see from Figure \ref{fig:fda_synthia}, the approach allows to better align the depth ranges and makes the synthetic source data less ``clean'', matching the fact that real-world target data computed with stereo vision has more artifacts and a less sharp distribution (see Section \ref{sec:implementation} for more details on the employed datasets). On the other side, using an excessively large $\beta$ introduces visual artifacts that can affect the network's performance.

\subsection{Feature Level Adaptation}
\label{sec:feat_adap}
\textbf{Cross-Modalities Attention} 
In order to perform feature exchange, the multi-head attention module is shared between the two input modalities $x_{rgb}$ and $x_{d}$. 
Cross-modal attention was proposed to provide latent adaptation across modalities in visual-text multimodal tasks \cite{tsai2019multimodal}:
\begin{equation}
    \text{Cross-Att}_{\beta \rightarrow \alpha}(Q_{\alpha},K_{\beta},V_{\beta}) = softmax\left(\frac{Q_{\alpha}K_{\beta}^{T}}{\sqrt{d_{head}}}\right)V_{\beta}
\end{equation}
where $Q$ is the query, $K$ is the key, $V$ is the value and $d_{head}$ is the dimension of the head \cite{vaswani2017attention}.
Transformer attention can be seen as an information retrieval mechanism: the generated query is specified from a key that returns a value. %
Differently from previous Transformer-based fusion approaches which considered feature-fusion at the end of each attention head \cite{liu2022cmx}, following the idea introduced in \cite{barbato2022depthformer},  the proposed framework acts directly at the core of the architecture by swapping the keys as in the following equation:
\begin{equation}
    \text{Att}(Q_{rgb},K_{d},V_{rgb}) = softmax\left(\frac{Q_{rgb}K_{d}^{T}}{\sqrt{d_{head}}}\right)V_{rgb}
\label{eq:cross-att}
\end{equation}
Unlike \cite{barbato2022depthformer} where they investigated the effect of the interaction across each modality, we focused on the impact of this operation on the generalization ability of the network, \giulia{i.e., multi-modal pre-training}. %
We assume the interaction between the two modalities should be consistent across different data distributions, as proved by the ablation studies in Section \ref{sec:abl_crossatt}.
The multi-head mixed attention feature $x_{rgb+d}$ (Eq. \ref{eq:cross-att}) is served at the decoder level as in \cite{xie2021segformer}.

\subsection{Output Level Adaptation}
\label{sec:output_level}

\textbf{Self-Training} 
During target adaptation, pseudo-labels $\hat{y}^t_{rgb+d}$ are assigned to unlabeled target data by the model through a self-training procedure. These labels are not always accurate, and filtering them can improve the performance of the model. The adopted filtering function uses a combination of probability thresholds and top-k filtering to select high-quality pseudo-labeled data points for training and discard the unlabeled ones. Following \cite{yang2020fda}, we considered valid only the predictions with a confidence score above $0.9$ or the ones that are within the top-$66\%$ confidence values.

Furthermore, we took under consideration that in the model pre-trained on the source dataset -thus synthetic data- depth maps are typically ground truth rendered maps free from noise, artifacts and missing points.
Real-world depth maps, especially if obtained through stereo-matching, as in the case of the cityscapes dataset, are corrupted by noise and have missing disparity values due to occlusions or to limitations of the stereo-matching strategy. 
In a vanilla multimodal approach, color and depth features equally contribute to the loss term.
In our setting, the depth information does not directly produce the semantic data estimation but contributes to the attention mechanism employed to construct the actual feature. Nevertheless, masking pixels with missing or corrupted depth data during the computation of pseudo-labels has the potential to significantly aid in the adaptation process.

The loss driven by the self-teaching module is thus computed as:
\begin{equation}
    \mathcal{L}_{pseudo} = L_{CE}(p(x^{t}_{rgb+d}),M(x^{t}_{rgb}, x^{t}_{d}) \circ \hat{y}^t_{rgb+d} )
\end{equation}
where the pseudo-label selection mask is:
\begin{equation}
    M(x_{rgb}, x_{d}) = \begin{cases*}
1 & \textrm{if $x_{d}$ is valid and} \\  \vspace{1mm}& \textrm{[$p > 0.9$ or $p\in$ Top-$66\%$] } \\ 
0  & \textrm{otherwise}
\end{cases*}
\end{equation}

Notice how the resulting binary mask is a combination of the probability-based and top-$k$ masks with the depth validity constraint.
By employing depth as an indicator of the model's uncertainty, it becomes possible to enhance the model without the need for additional hyperparameters.

\noindent\textbf{Depth Entropy minimization} 
Pseudo-labels allow network training on unlabeled target data imitating the label's existence. However, it can be experimentally noticed that, as training progresses, after a certain point the learning curve begins to decline \cite{fleuret2021uncertainty}.
Initially, self-training serves as a means to narrow the discrepancy between the knowledge obtained from the source dataset $\mathcal{D}_{s}$ and that required for satisfactory performance on the target dataset $\mathcal{D}_{t}$. However, as the training proceeds, the network becomes overly self-assured in its predictions, thereby diminishing its efficacy and resulting in increased misclassification errors.

Under the assumption that real-world depth data often contains inconsistencies, we have chosen to assign greater weight to images captured at shorter distances, on the basis that disparity values for distant objects are more prone to error. Furthermore, close objects have a higher resolution in terms of pixels in the image and thus are better represented and easier to be properly classified also in color data. 

To exploit this, we developed an entropy minimization strategy exploiting distance information through the disparity map. Recall that the disparity map, which is the typical output of stereo vision methods, is inversely proportional to depth data. Therefore a smaller disparity corresponds to objects that are captured with a lower spatial resolution by pinhole cameras and have less reliable depth values. %

We started from the standard entropy minimization target proposed in \cite{vu2018advent} to aid the domain adaptation task:
\begin{equation}
    \mathcal{L}_{ent}(x) = -\sum_{c}^{C}p(x)^{(c)}log p(x)^{(c)}
\end{equation}
We modified the loss by adding a weighting term that depends on the distance from the camera giving more relevance to close points. We found that simply weighting the entropy loss with the disparity values $x_{disp}^{(w,h)}$ led to the best performances: 
\begin{equation}
    \mathcal{L}_{depth-ent} = \sum_{h}^{H} \sum_{w}^{W} \mathcal{L}_{ent}(x_{rgb+d})^{(w,h)} * (x_{disp}^{(w,h)})
\end{equation}

\begin{table*}[htbp]
  \begin{adjustbox}{width=\textwidth}
\begin{tabular}{ll|llllllllllllllll|ll}
    \toprule
 \textbf{Method}     & \textbf{Backbone}   &  
\rotatebox{90}{road} & \rotatebox{90}{side.} & \rotatebox{90}{build.} & \rotatebox{90}{wall*} & \rotatebox{90}{fence*} & \rotatebox{90}{pole*} & \rotatebox{90}{light} & \rotatebox{90}{sign} & \rotatebox{90}{vege.} & \rotatebox{90}{sky} & \rotatebox{90}{pers.} & \rotatebox{90}{rider} & \rotatebox{90}{car} & \rotatebox{90}{bus} & \rotatebox{90}{motor} & \rotatebox{90}{bike} &
 $\textrm{\textbf{mIoU}}_{16}$  & $\textrm{\textbf{mIoU}}_{13}$ \\
\toprule
SFDA \cite{liu2021source}       & ResNet-50   & 81.9 & 44.9 & 81.7 & 4.0 & 0.5 & 26.2 & 3.3 & 10.7 & 86.3 & 89.4 & 37.9 & 13.4 & 80.6 & 25.6 & 9.6 & 31.3 & 39.2 & 45.9  \\
URMA \cite{fleuret2021uncertainty}      & ResNet-101  & 59.3 & 24.6 & 77.0 & 14.0 & 1.8 & 31.5 & 18.3 & 32.0 & 83.1 & 80.4 & 46.3 & 17.8 & 76.7 & 17.0 & 18.5 & 34.6 & 39.6  & 45.0    \\
DT+AC \cite{yang2022source} & ResNet-101 & 77.5 & 37.4 & 80.5 & 13.5 & 1.7 & 30.5 & 24.8 & 19.7 & 79.1 & 83.0 & 49.1 & 20.8 & 76.2 & 12.1 & 16.5 & 46.1 & 41.8 & 47.9 \\
LD \cite{you2021domain}        & ResNet-101 & 77.1 & 33.4 & 79.4 & 5.8 & 0.5 & 23.7 & 5.2 & 13.0 & 81.8 & 78.3 & 56.1 & 21.6 & 80.3 & 49.6 & 28.0 & 48.1 & 42.6  & 50.1    \\
HCL  \cite{huang2021model}       & ResNet-101 & 80.9 & 34.9 & 76.7 & 6.6 & 0.2 & 36.1 & 20.1 & 28.2 & 79.1 & 83.1 & 55.6 & 25.6 & 78.8 & 32.7 & 24.1 & 32.7 & 43.5  & 50.2     \\
SFUDA \cite{ye2021source} & ResNet-101 & 90.9 & 45.5 & 80.8 & 3.6 & 0.5 & 28.6 & 8.5 & 26.1 & 83.4 & 83.6 & 55.2 & 25.0 & 79.5 & 32.8 & 20.2 & 43.9 & 44.2 & 51.9 \\
DT-ST \cite{zhao2023towards} & ResNet-101 & 88.9 & 45.8 & 83.3 & 13.7 & 0.8 & 32.7 & 31.6 & 20.8 & 85.7 & 82.5 & 64.4 & 27.8 & 88.1 & 50.9 & 37.6 & 57.3 & 50.7 & 58.8 \\
SOMAN+cPAE \cite{kundu2021generalize} & ResNet-101  &  90.5 & 50.0 & 81.6 & 13.3 & 2.8 & 34.7 & 25.7 & 33.1 & 83.8 & 89.2 & 66.0 & 34.9 & 85.3 & 53.4 & 46.1 & 46.6  & 52.0  & 60.1   \\
\midrule
Source Only RGB & MiT-B5 & 28.5 & 19.7 & 56.7 & 3.4 & 0.2 & 39.1 & 34.9 & 18.0 & 81.0 & 86.1 & 64.0 & 11.6 & 82.6 & 28.2 & 7.5 & 29.4 & 36.9 & 42.2 \\
RGB + FDA\cite{yang2020fda} & MiT-B5 & 41.1  & 27.2  & 60.6  & 6.3 & 0.3 & 42.7 & 31.0 & 27.2 & 82.2 & 87.8 & 65.8 & 15.0 & 61.4 & 38.9 & 9.0 & 30.8 &   39.2 & 44.5 \\
MISFIT (Ours) & MiT-B5 & 80.2 & 38.5 & 85.9 & 30.3 & 1.2 & 52.3& 56.8 & 29.0 & 89.9 &88.3 &68.1 & 10.8& 92.1& 69.0& 26.3& 52.6&  $\textbf{54.5}$  & $\textbf{60.6}$\\
\bottomrule
\end{tabular}
\end{adjustbox}
\caption{Semantic segmentation results for the SYNTHIA-to-Cityscapes source-free adaptation task. $mIoU_{13}$ denotes performance over 13 classes excluding those marked with *.}
\label{tab:synthia}
\end{table*}

\begin{table*}[htbp]
  \begin{adjustbox}{width=\textwidth}
\begin{tabular}{l|lllllllllllllllllll|ll}
    \toprule
 \textbf{Method}   &   
\rotatebox{90}{road} & \rotatebox{90}{side.} & \rotatebox{90}{build.} & \rotatebox{90}{wall} & \rotatebox{90}{fence} & \rotatebox{90}{pole} & \rotatebox{90}{light} & \rotatebox{90}{sign} & \rotatebox{90}{vege.} & \rotatebox{90}{terr.*} & \rotatebox{90}{sky} & \rotatebox{90}{pers.} & \rotatebox{90}{rider} & \rotatebox{90}{car} & \rotatebox{90}{truck*} &  \rotatebox{90}{bus} & \rotatebox{90}{train*} & \rotatebox{90}{motor} & \rotatebox{90}{bike} &
 $\textrm{\textbf{mIoU}}_{19}$   & $\textrm{\textbf{mIoU}}_{16}$  \\
\toprule
Source Only RGB &  70.9 & 45.7 & 71.2 & 12.4 & 7.8 & 37.4 & 37.5 & 35.4 & 84.8 & 24.8 & 81.7 & 65.9 & 23.4 & 65.7& 11.5& 21.5& 2.8& 41.0 & 45.7 & 41.4 &  46.8\\
RGB + FDA \cite{yang2020fda} & 64.0 & 47.2 & 60.0 & 7.1 & 6.8 & 41.2 & 38.6 & 43.9 & 84.1 & 20.3 & 79.0 & 66.8 & 23.7 & 73.4 & 18.6 & 34.7 & 4.1& 37.5 & 39.8  & 41.6  & 46.7 \\
MISFIT  (w/o $\mathcal{L}_{depth-ent}$) & 74.2 & 61.8 & 66.0 & 8.6 & 15.9 & 51.3 & 55.2 & 60.7 & 87.0 & 24.4 & 73.0 & 72.2 & 32.7 & 86.8 & 31.6 & 62.9 & 0.0 & 39.9  & 57.1 & 50.6 & 56.6\\
MISFIT (Ours) & 76.2 & 63.2 & 68.7 & 5.6 & 13.7 & 50.9 & 57.2 & 60.8 & 87.2 & 21.6 & 89.8 & 72.2 & 33.3 & 86.6 & 30.0 & 54.8 & 7.8 & 43.9 & 58.2 & $\textbf{51.7}$ & $\textbf{57.6}$\\
\bottomrule
\end{tabular}
\end{adjustbox}
\caption{Semantic segmentation results for the SELMA-to-Cityscapes source-free adaptation task. $mIoU_{16}$ denotes performance over 16 classes - corresponding to SYNTHIA classes - excluding those marked with *.}
\label{tab:selma}
\end{table*}

\begin{figure*}[h!]
\begin{subfigure}[b]{\textwidth}
\centering
\begin{subfigure}[b]{0.15\textwidth}
\caption*{RGB}
\includegraphics[width=\textwidth]{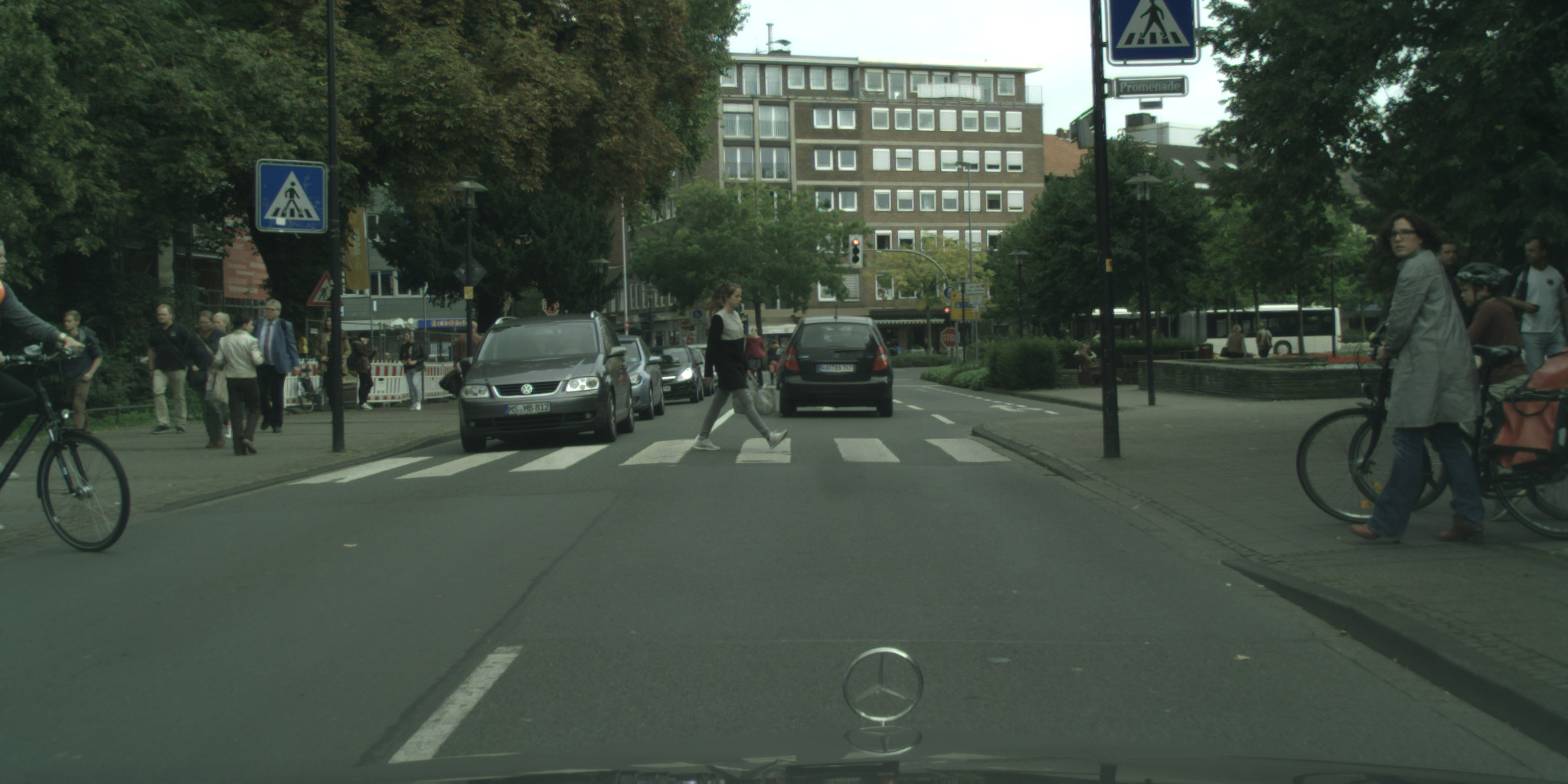}
\end{subfigure} 
\begin{subfigure}[b]{0.15\textwidth}
\caption*{Depth}
\includegraphics[width=\textwidth]{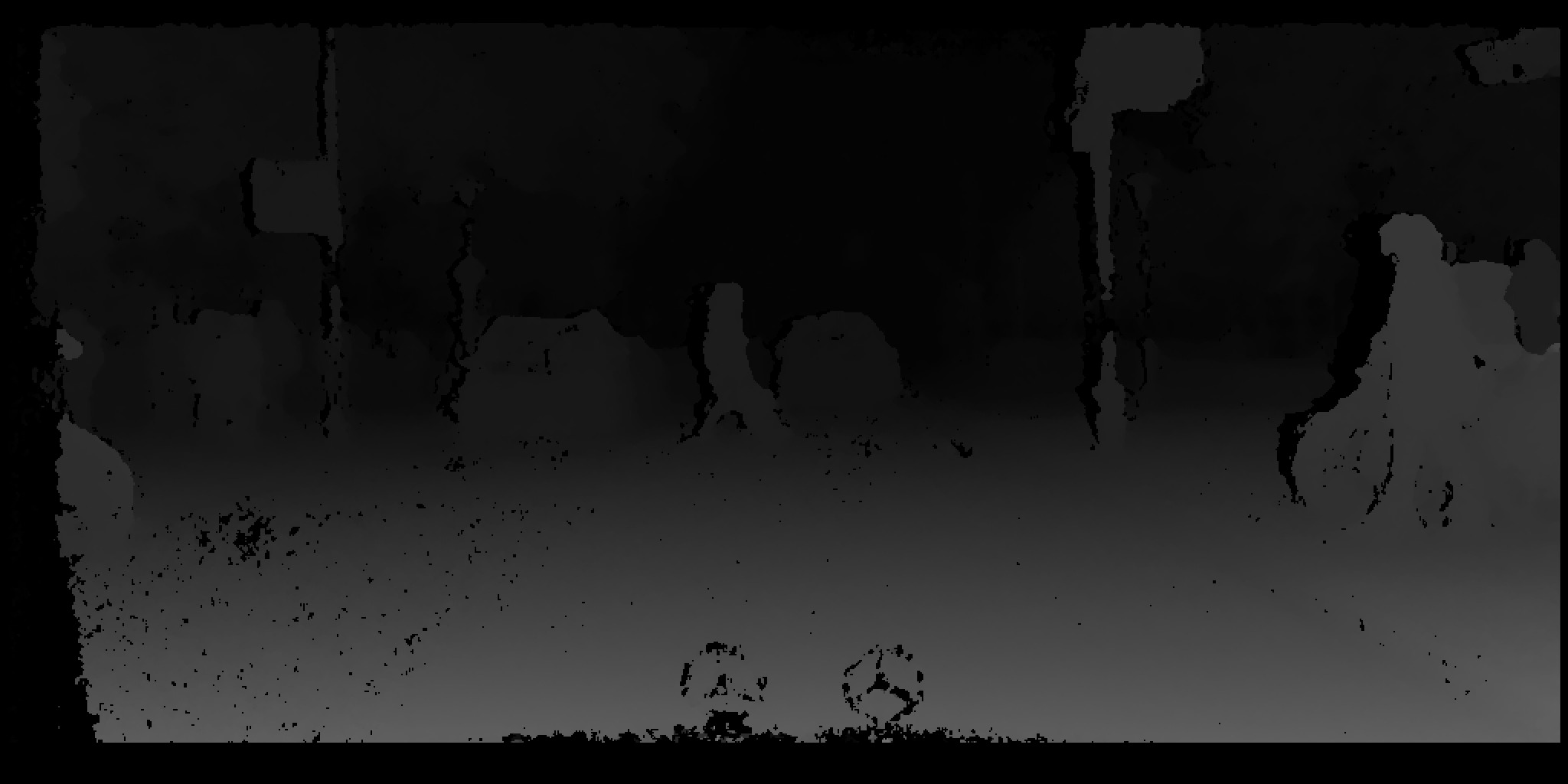}
\end{subfigure} 
\begin{subfigure}[b]{0.15\textwidth}
\caption*{GT}
\includegraphics[width=\textwidth]{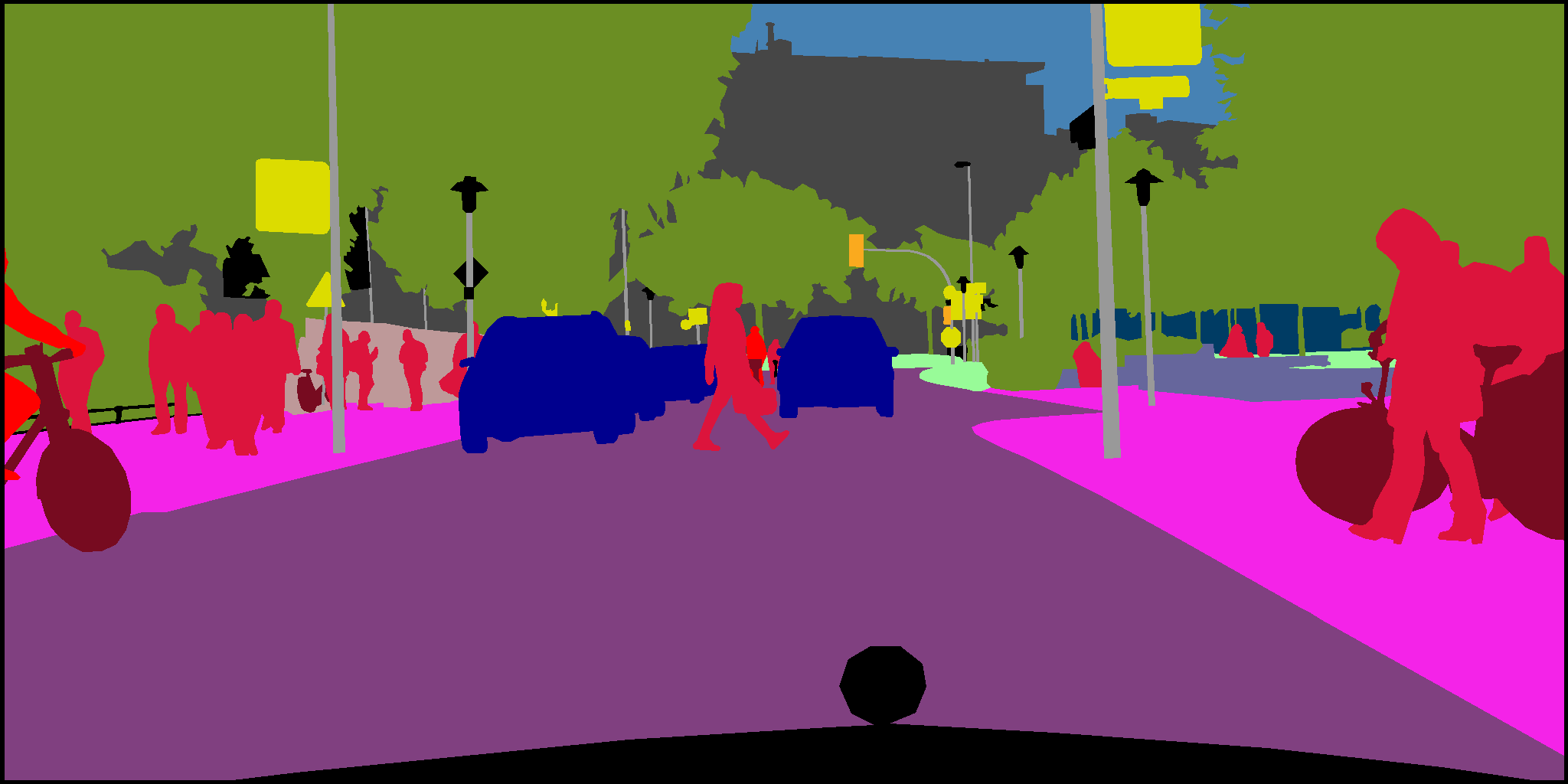}
\end{subfigure}
\begin{subfigure}[b]{0.15\textwidth}
\caption*{RGB+FDA\cite{yang2020fda}}
\includegraphics[width=\textwidth]{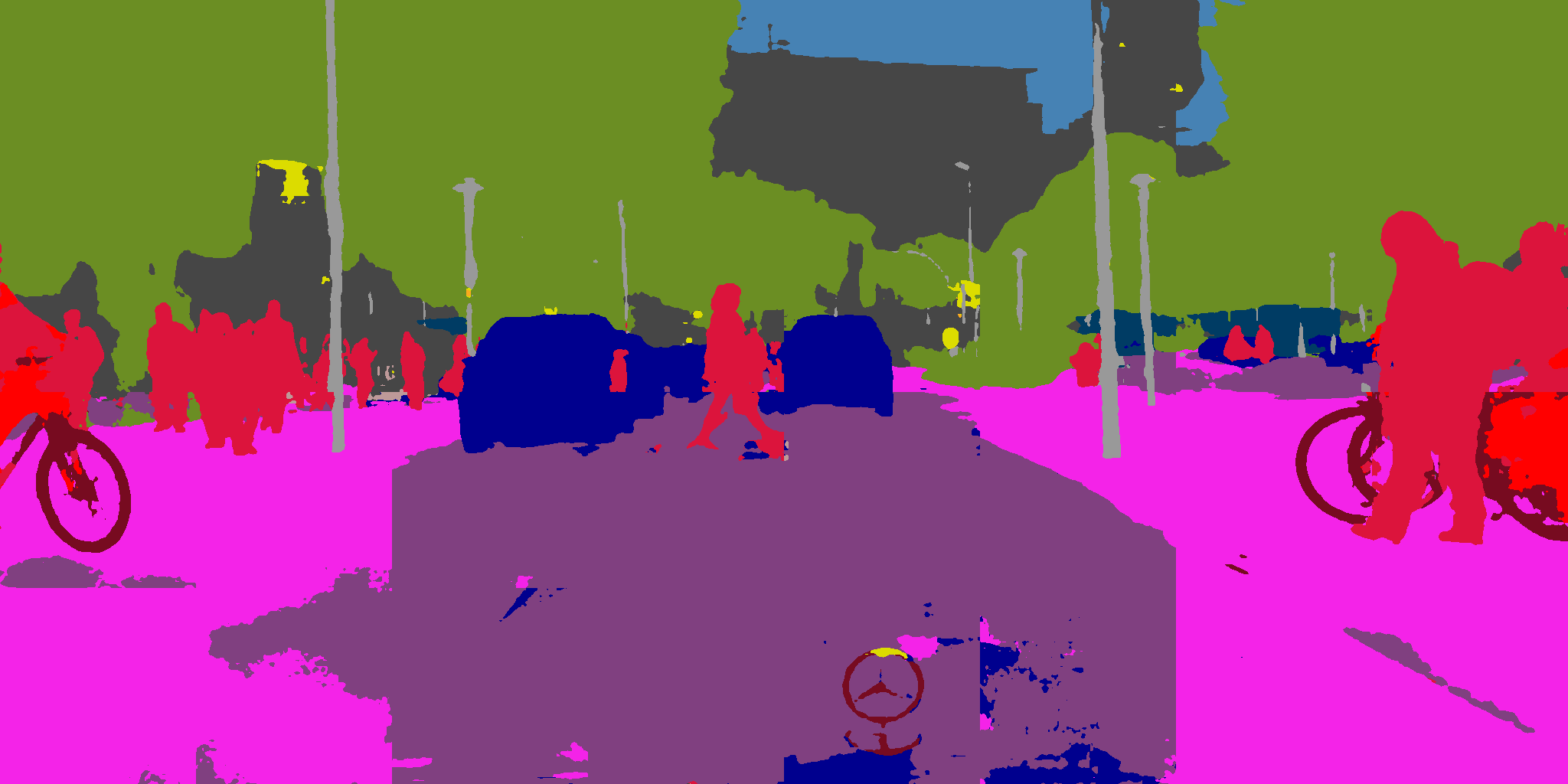}
\end{subfigure}
\begin{subfigure}[b]{0.15\textwidth}
\caption*{MISFIT\tiny{w/o$\mathcal{L}_{depth-ent}$}}
\includegraphics[width=\textwidth]{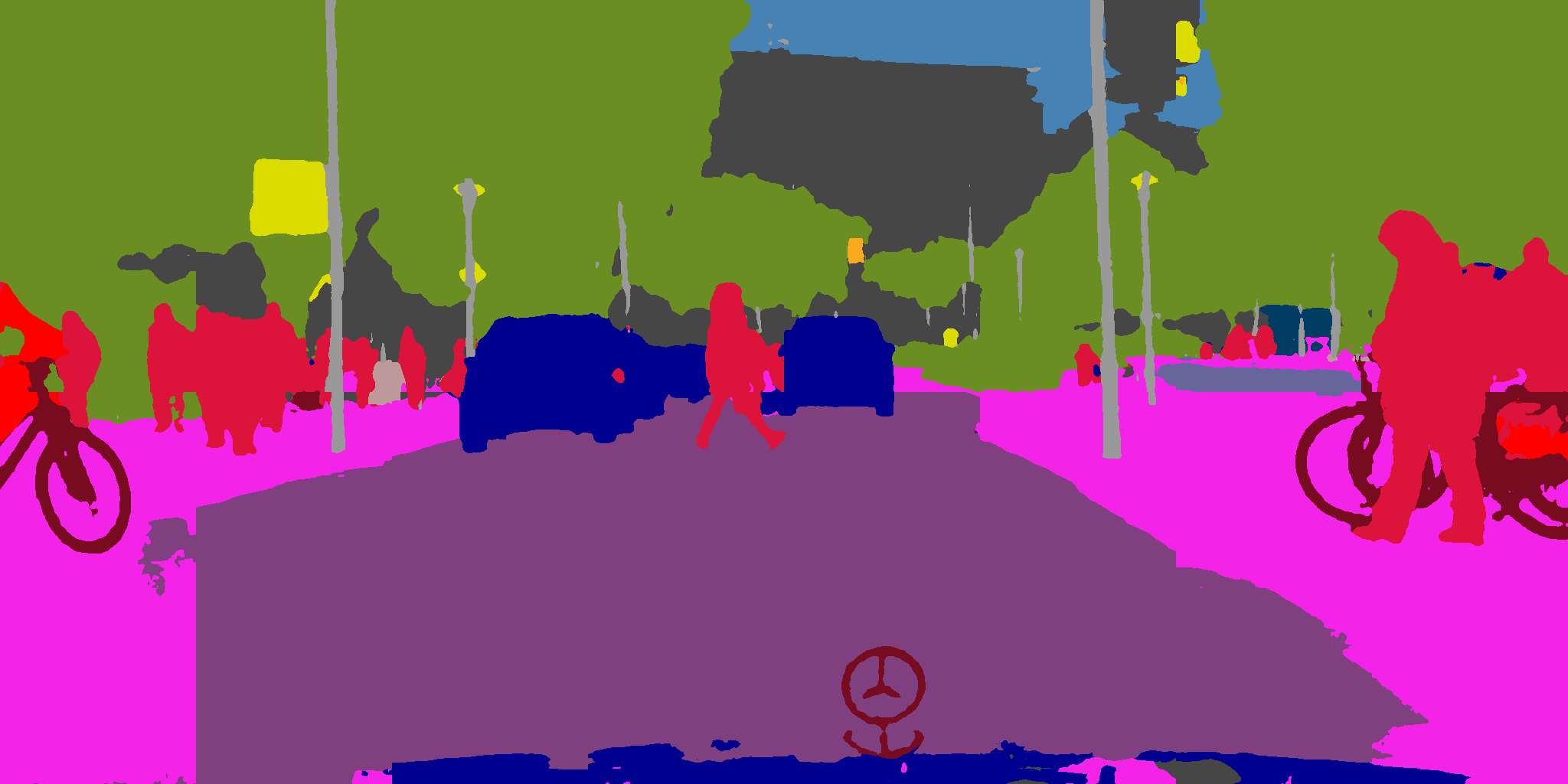}
\end{subfigure}
\begin{subfigure}[b]{0.15\textwidth}
\caption*{MISFIT}
\includegraphics[width=\textwidth]{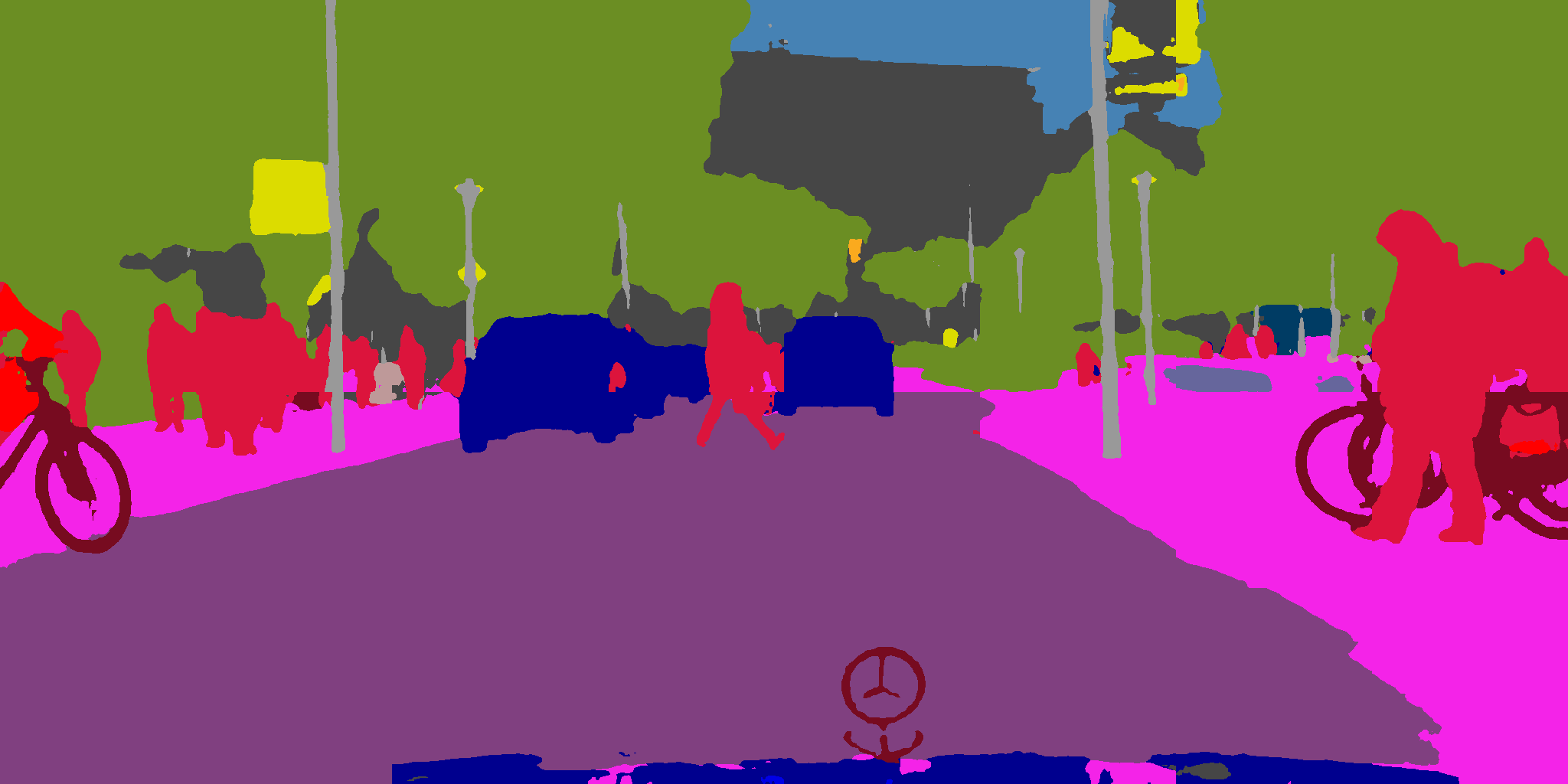}
\end{subfigure}
\end{subfigure}
\begin{subfigure}[b]{\textwidth}
\centering
\begin{subfigure}[b]{0.15\textwidth}
\includegraphics[width=\textwidth]{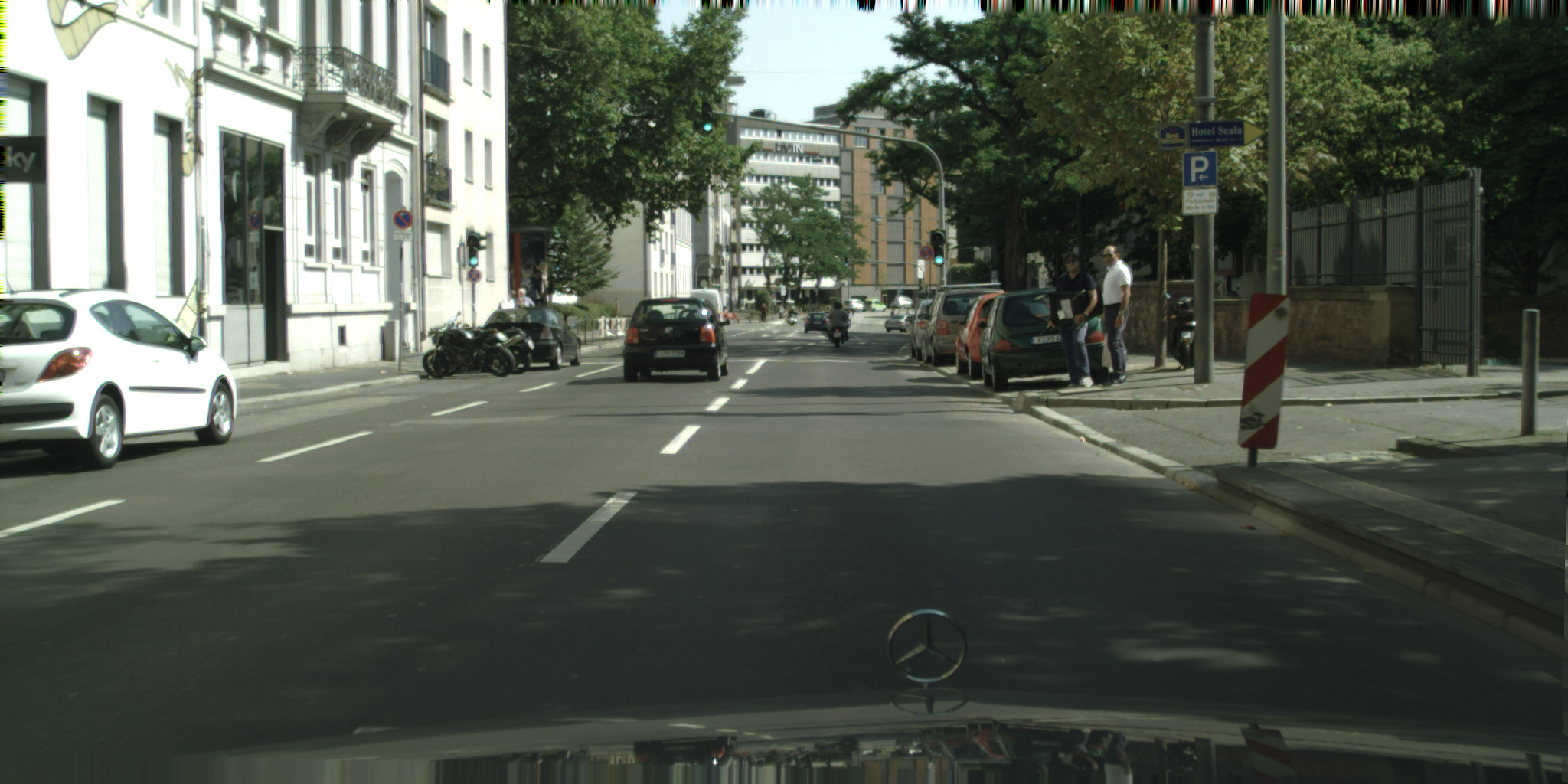}
\end{subfigure} 
\begin{subfigure}[b]{0.15\textwidth}
\includegraphics[width=\textwidth]{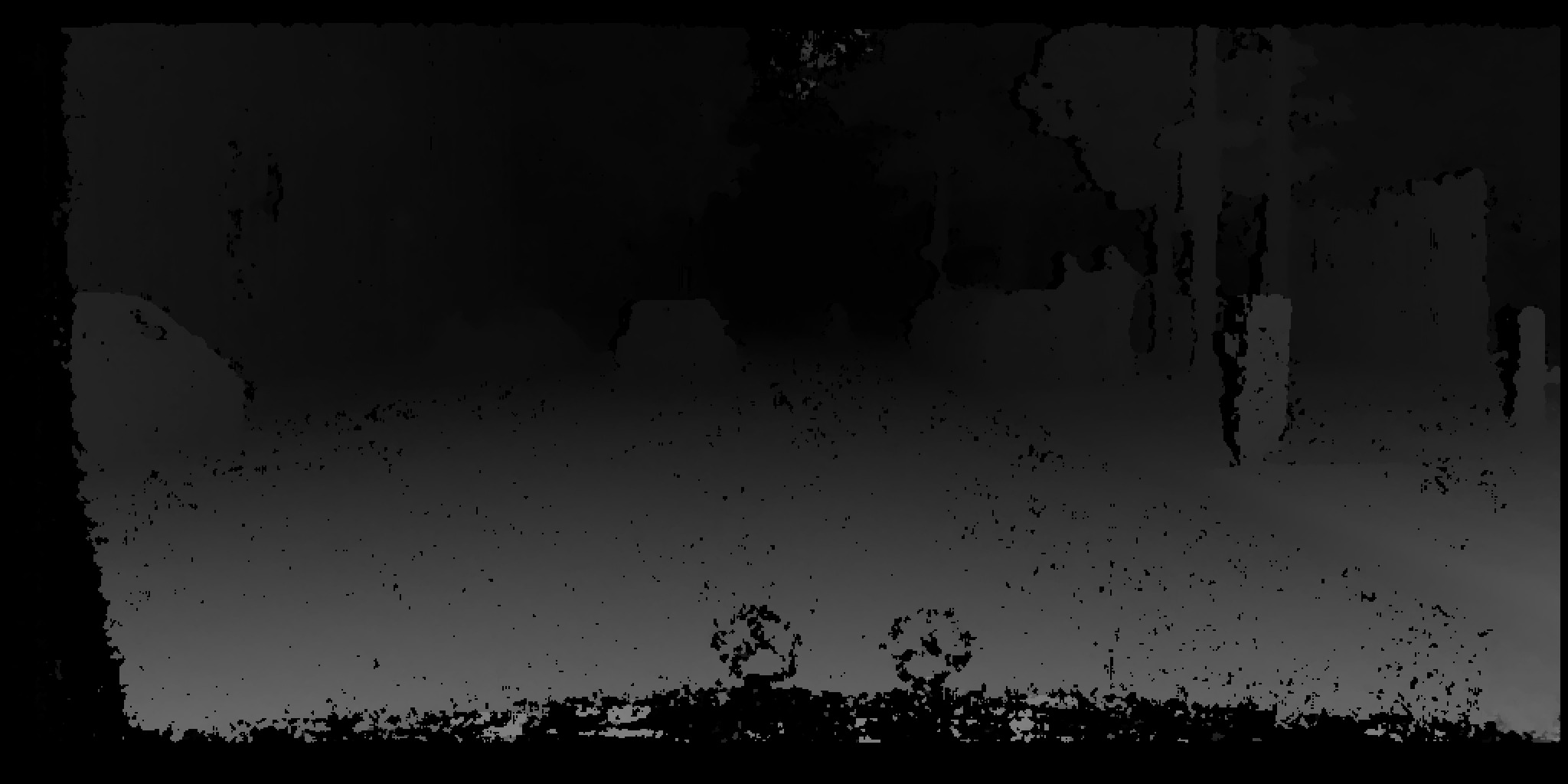}
\end{subfigure} 
\begin{subfigure}[b]{0.15\textwidth}
\includegraphics[width=\textwidth]{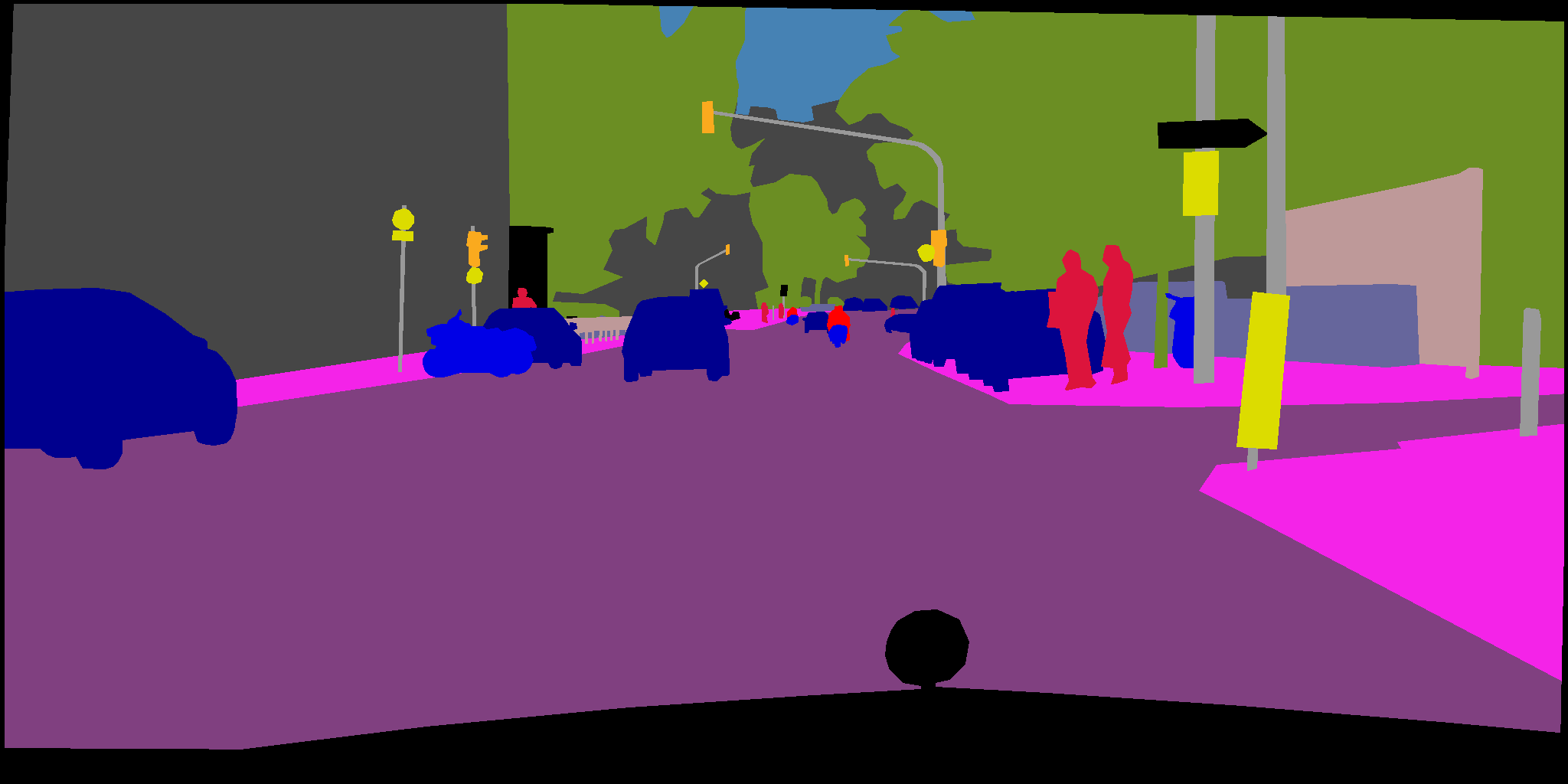}
\end{subfigure}
\begin{subfigure}[b]{0.15\textwidth}
\includegraphics[width=\textwidth]{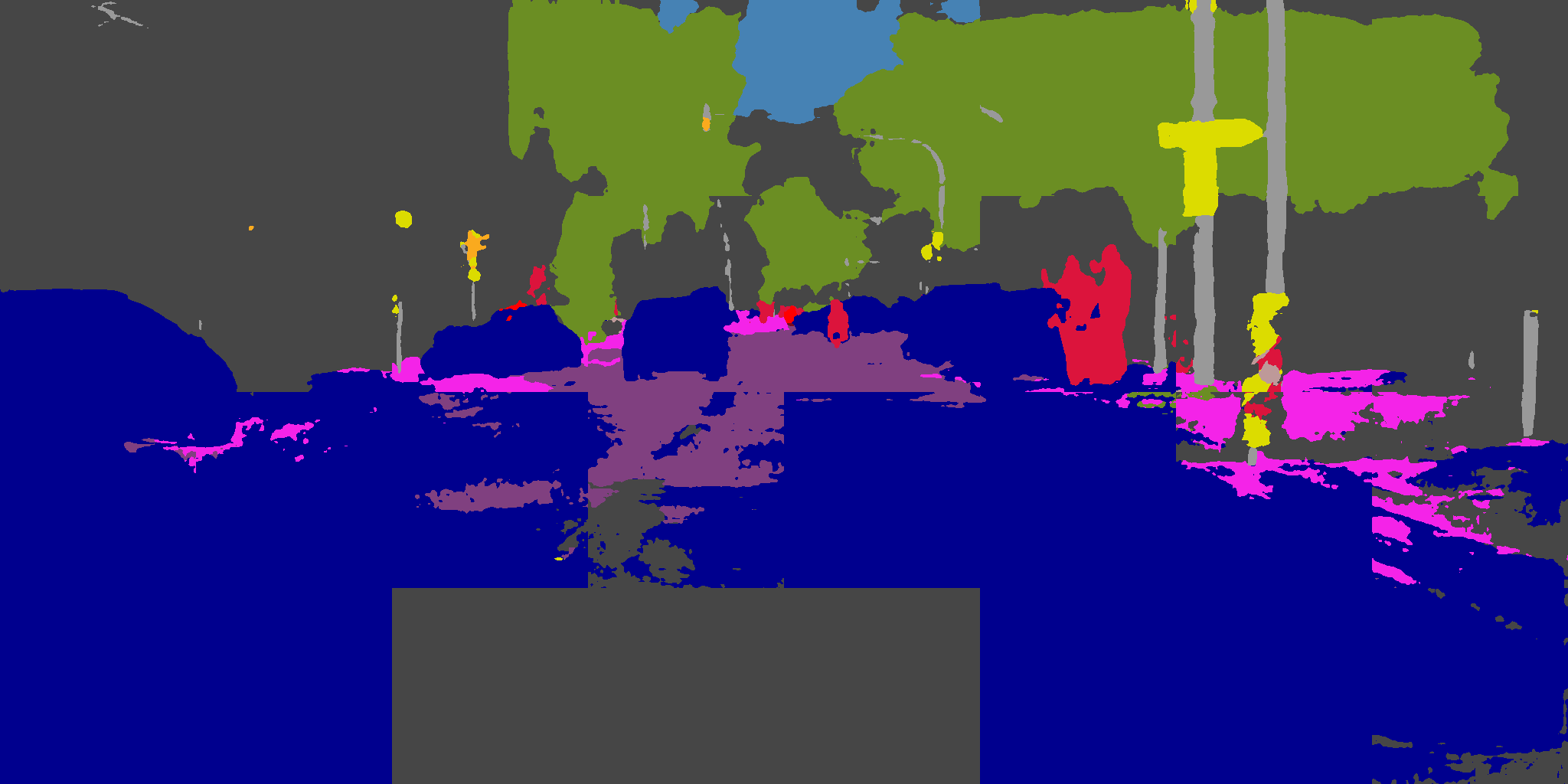}
\end{subfigure}
\begin{subfigure}[b]{0.15\textwidth}
\includegraphics[width=\textwidth]{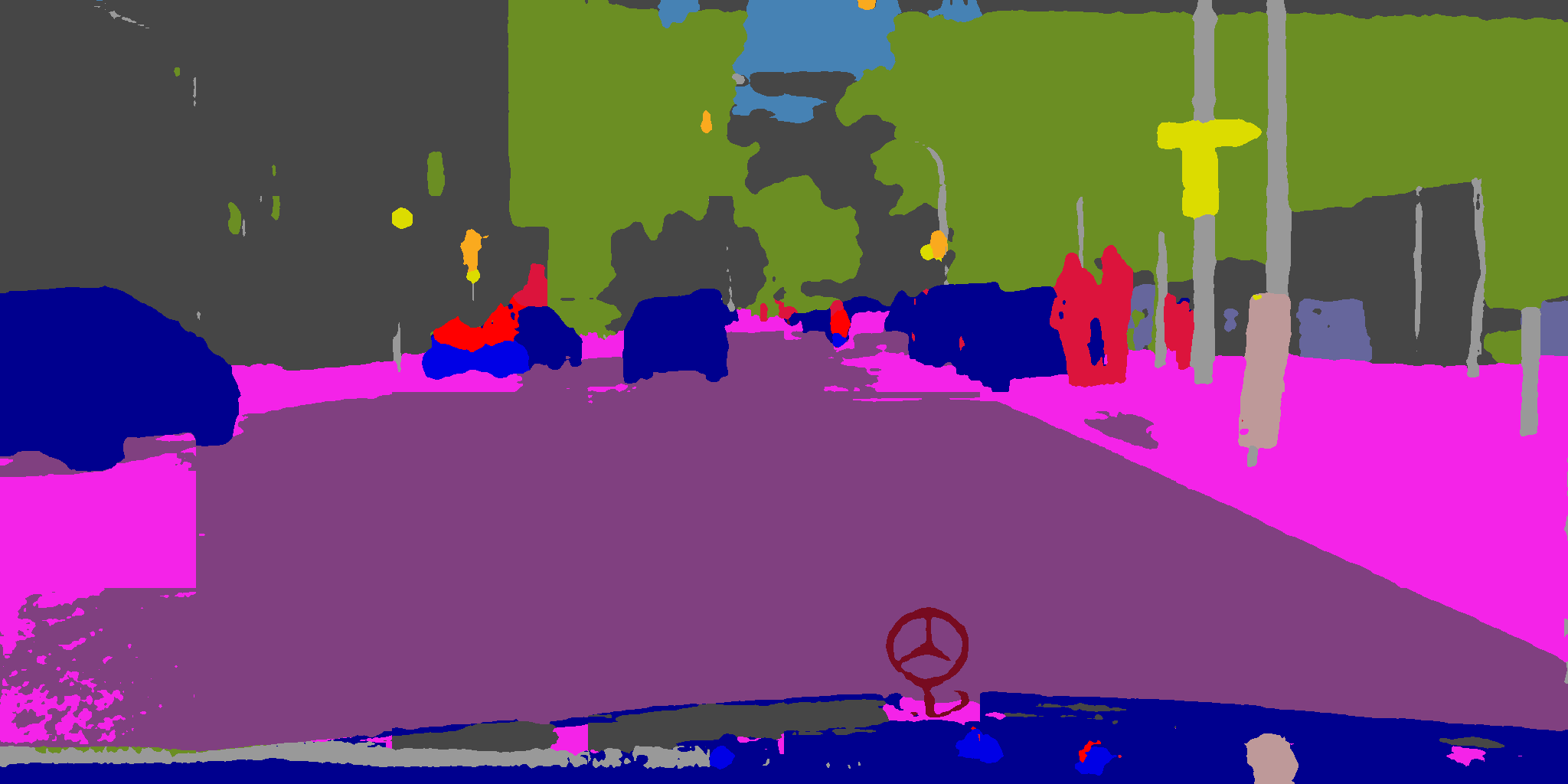}
\end{subfigure}
\begin{subfigure}[b]{0.15\textwidth}
\includegraphics[width=\textwidth]{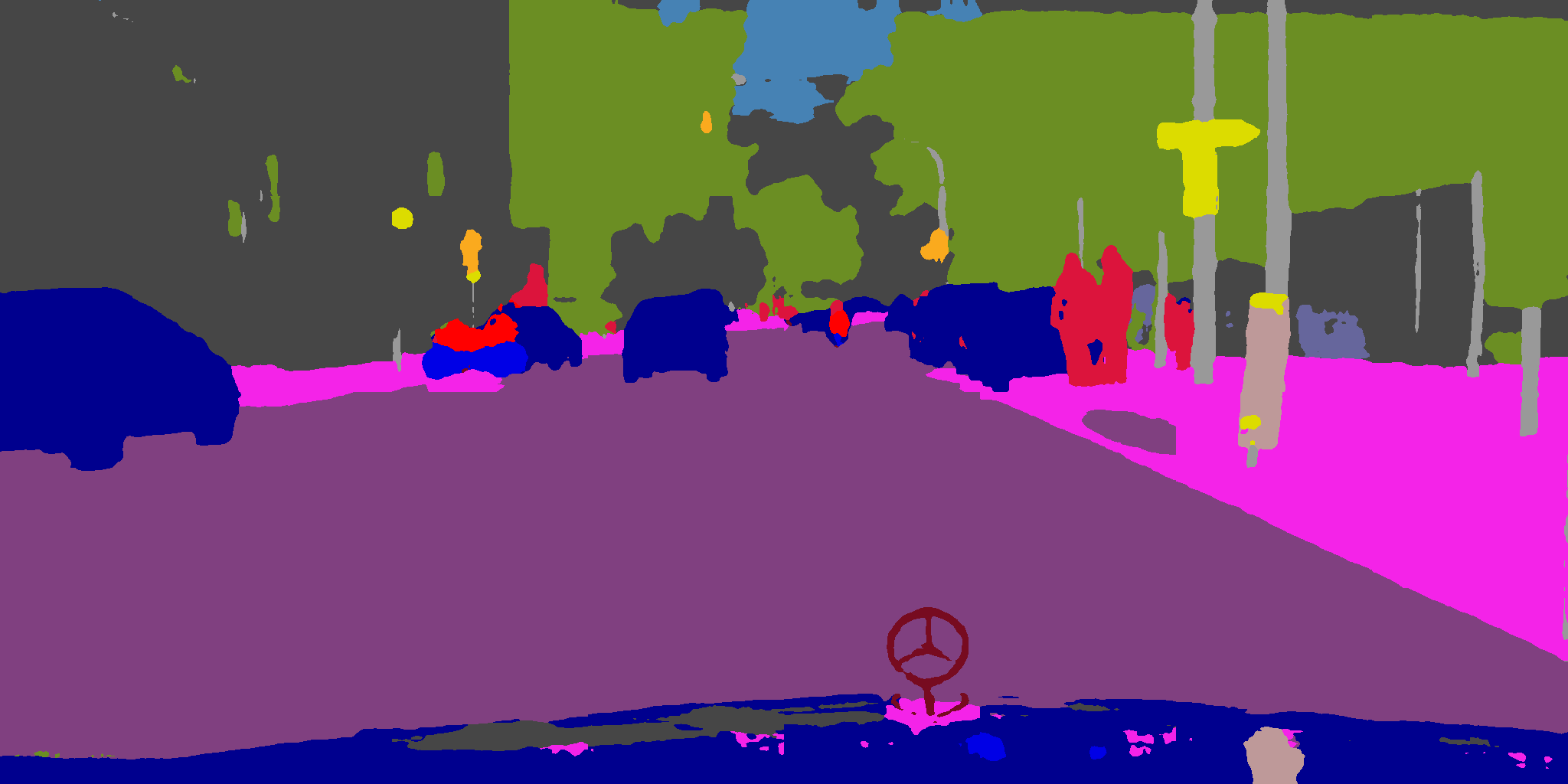}
\end{subfigure}
\end{subfigure}
\begin{subfigure}[b]{\textwidth}
\centering
\begin{subfigure}[b]{0.15\textwidth}
\includegraphics[width=\textwidth]{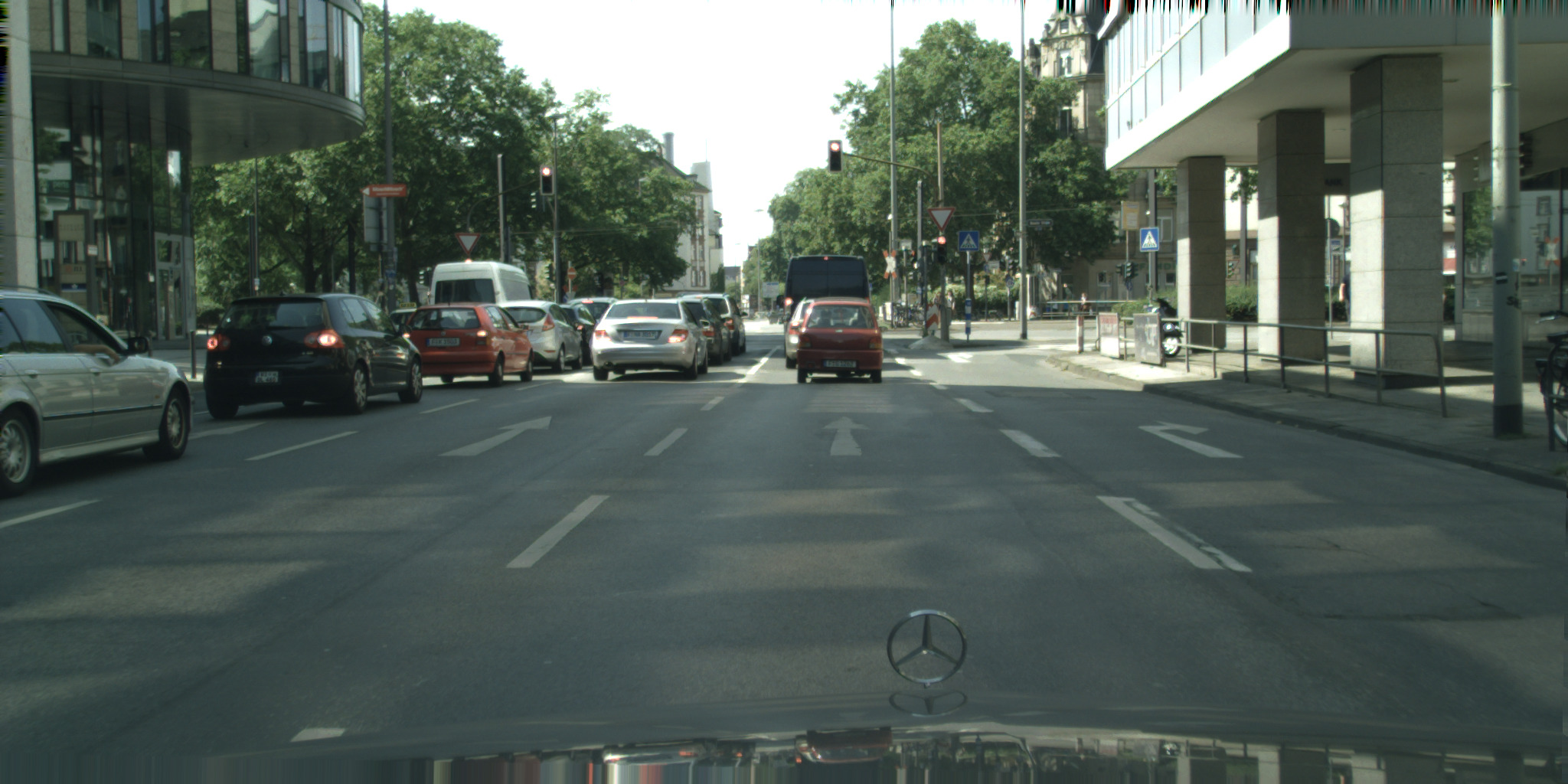}
\end{subfigure} 
\begin{subfigure}[b]{0.15\textwidth}
\includegraphics[width=\textwidth]{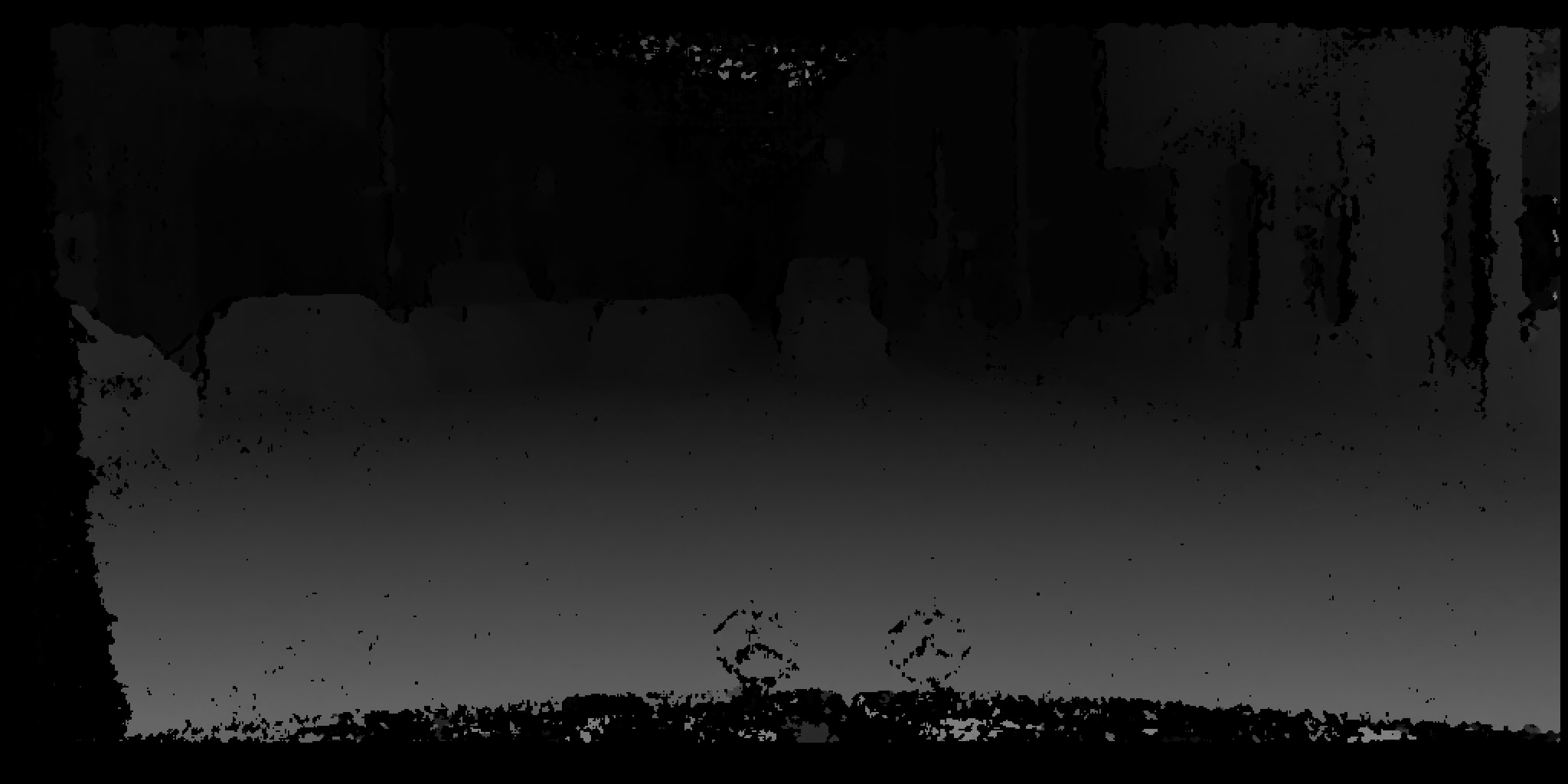}
\end{subfigure} 
\begin{subfigure}[b]{0.15\textwidth}
\includegraphics[width=\textwidth]{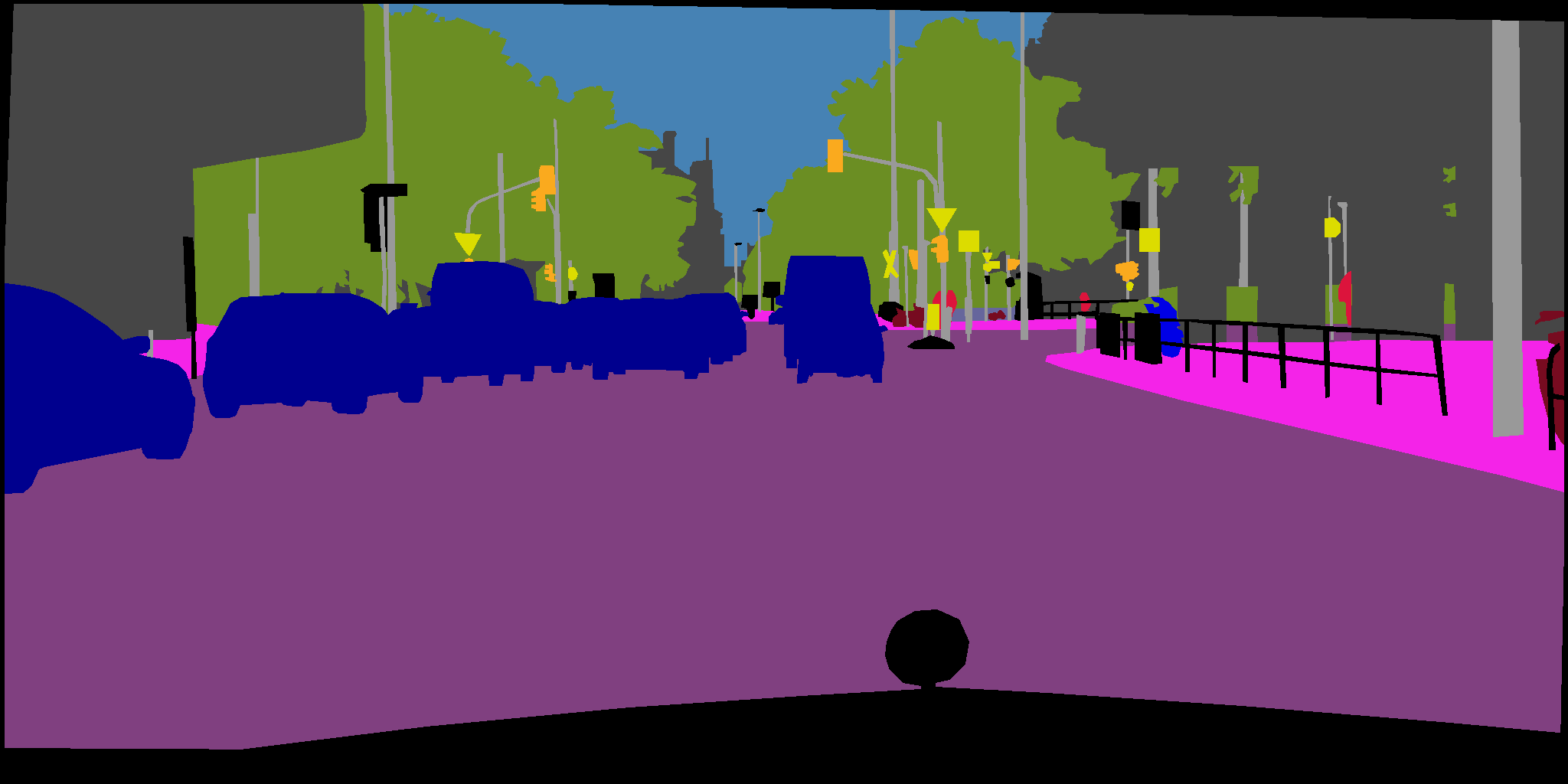}
\end{subfigure}
\begin{subfigure}[b]{0.15\textwidth}
\includegraphics[width=\textwidth]{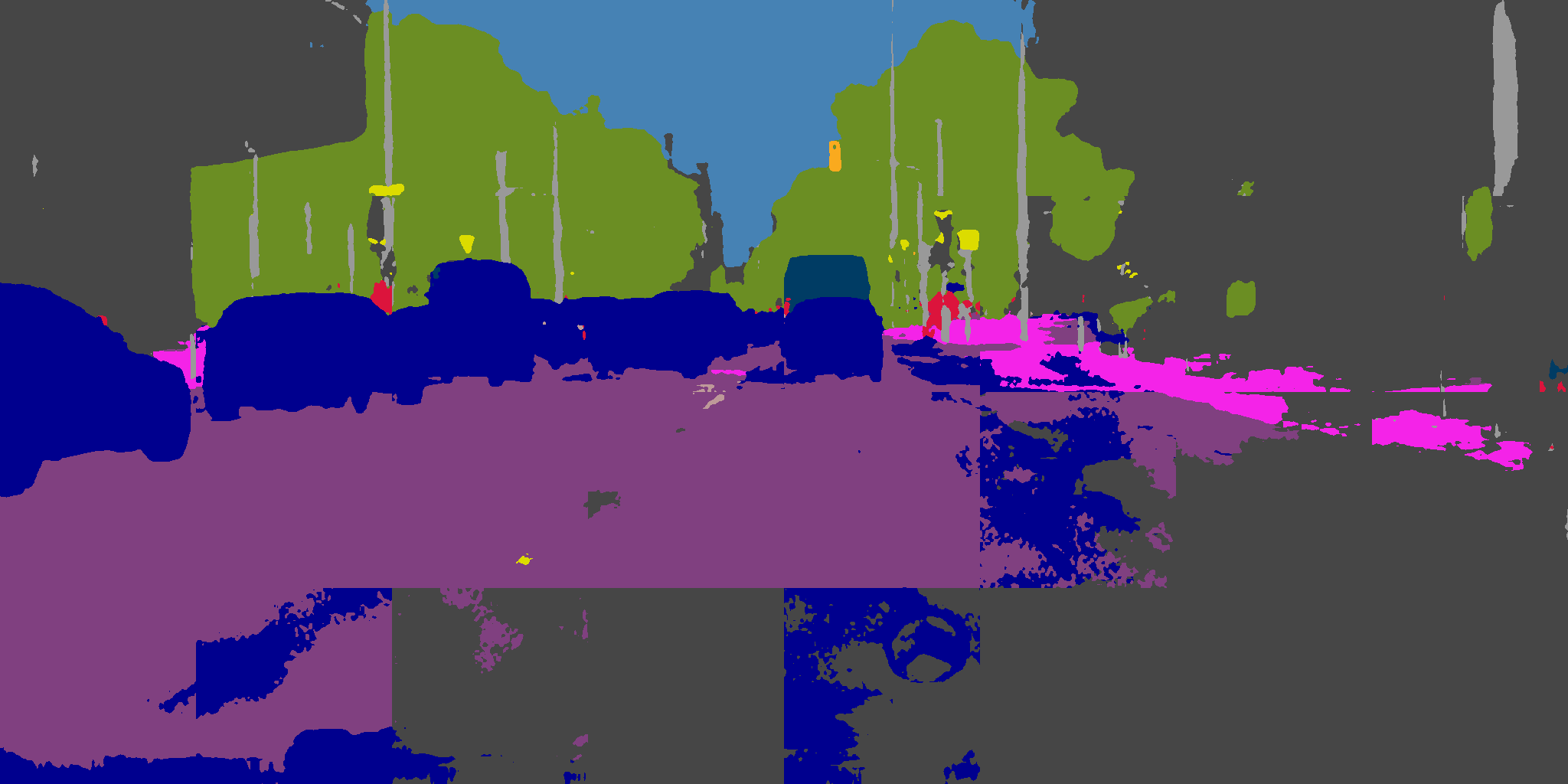}
\end{subfigure}
\begin{subfigure}[b]{0.15\textwidth}
\includegraphics[width=\textwidth]{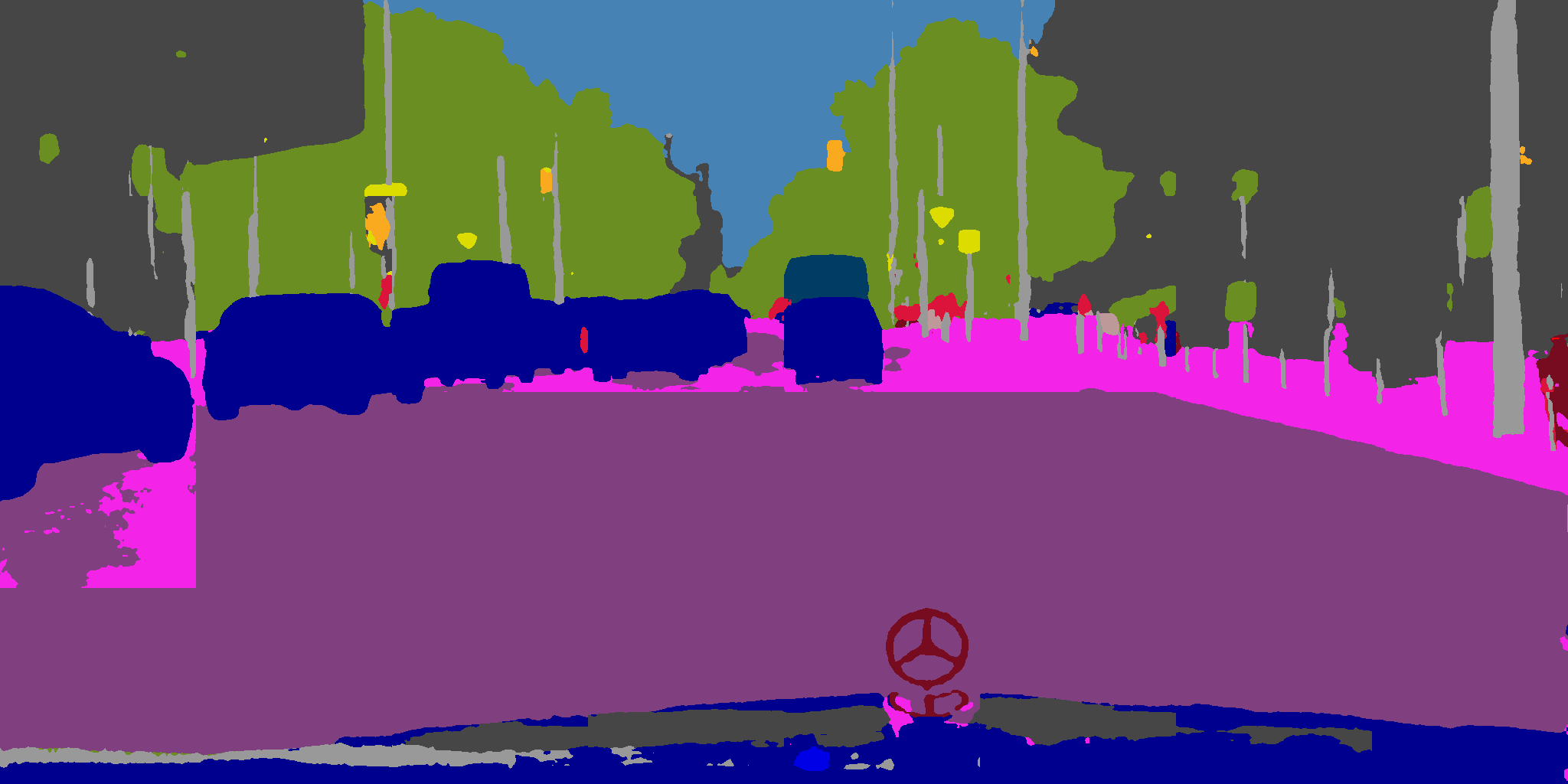}
\end{subfigure}
\begin{subfigure}[b]{0.15\textwidth}
\includegraphics[width=\textwidth]{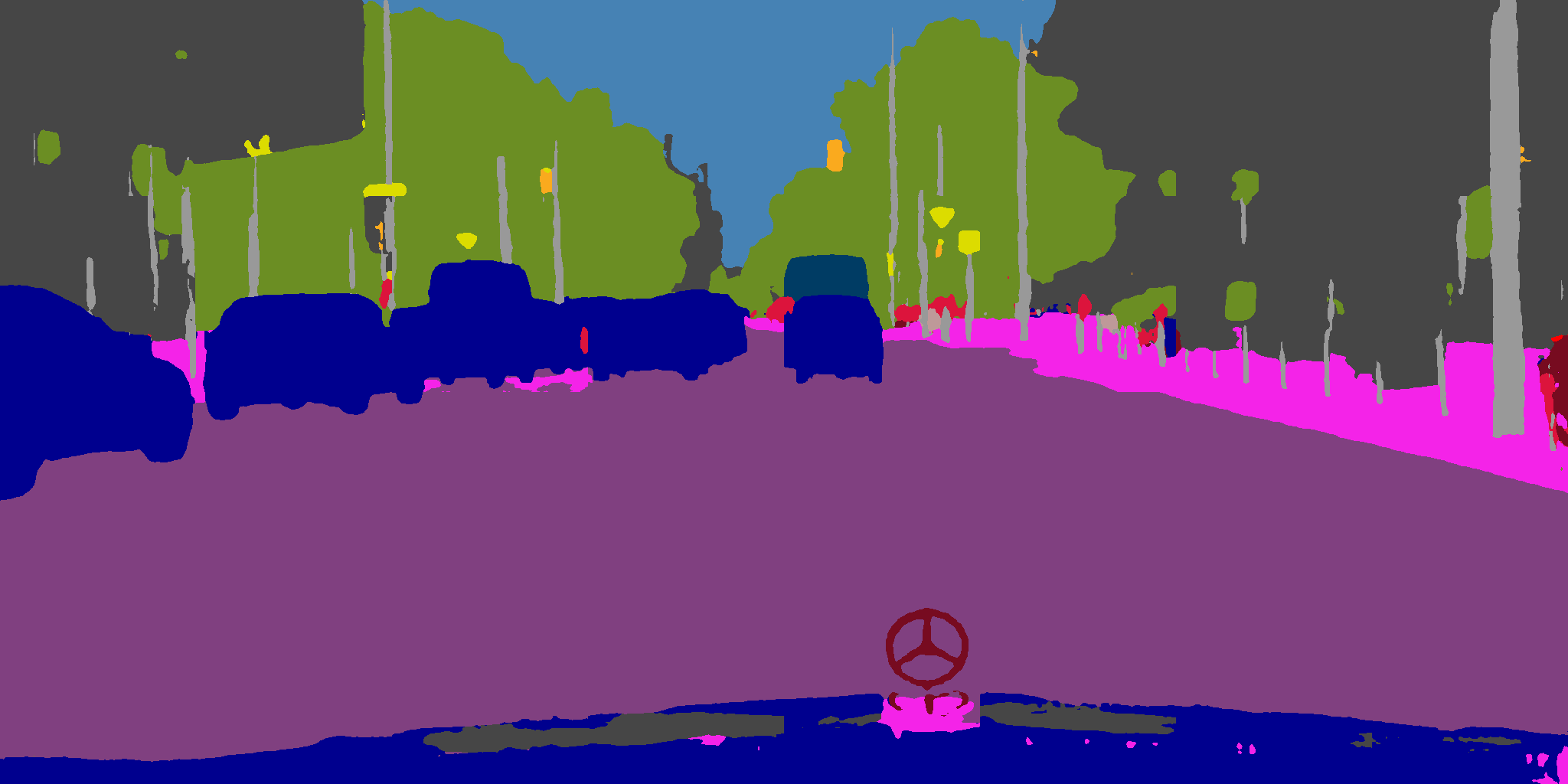}
\end{subfigure}
\end{subfigure}
\caption{Qualitative semantic segmentation results for the SYNTHIA-to-Cityscapes adaptation task.}
\label{fig:syn_qualitative}
\end{figure*}

\section{Results}
\label{sec:results}

In this section, we introduce the experimental framework and the employed datasets, then we present the numerical results obtained by our method. Finally we present some ablation studies to evaluate the impact of the different components of the approach.

\subsection{Datasets}
We evaluate our method on two synthetic-to-real road scene segmentation scenarios: (a) SYNTHIA-to-Cityscapes and (b) SELMA-to-Cityscapes.
For the supervised pre-training on source data, we employed the widely used  SYNTHIA dataset \cite{ros2016synthia} that contains 9400 total samples with a resolution of 1280×760. Furthermore, we made some tests also with the more recent SELMA dataset \cite{testolina2023selma}, which comprises 31k scenes with a resolution of 1280x640 in a wide range of different acquisition conditions. While well-known domain adaptation datasets like GTAV \cite{richter2016playing} (Synthetic) or BDD100K \cite{yu2020bdd100k} (Real) cannot be utilized in our approach due to the absence of depth maps, employing recently available datasets like SELMA offers a distinct advantage. The primary benefit lies in SELMA's provision of all the 19 classes present in Cityscapes, enabling a better matching between the datasets. As a target real-world dataset we used Cityscapes \cite{cordts2016cityscapes}, which is the most common benchmark for semantic segmentation in the driving environment. It includes 2975 training samples and 500 validation ones. Each image is provided with the associated depth map computed with stereo vision, while the resolution is 2048×1024. 
Notice that depth maps are the result of a stereo matching algorithm and consequently present many issues and artifacts, differently from the ones of SYNTHIA and SELMA, that contain ground truth data extracted from the rendering engine. This makes the domain adaptation task more challenging since it must adapt both from synthetic to real data and from ground truth depth to stereo vision data. 

\subsection{Implementation Details}
\label{sec:implementation}
We adopted SegFormer \cite{xie2021segformer} as the basic segmentation framework since it is a widespread well-performing approach based on vision transformers. Furthermore, previous studies have demonstrated its small generalization gap \cite{park2022dat}. In our framework, the encoder is shared between the two modalities thus reducing the number of parameters to be estimated and at the same time supporting depth attention processing and multimodal fusion as described in Section \ref{sec:feat_adap}.
The architecture is pre-trained on source data using the Adam optimizer for $40$ epochs (160k iterations) with batch size $4$ and learning rate starting from $6e-5$ with a weight decay rate of $0.01$. 
For the unsupervised target adaptation, where only target data is used, the batch size was set equal to 2. We employed the same data augmentation and input resolution ($512 \times 512$) as those in \cite{xie2021segformer}. 
For generating the target pseudo-labels, we utilize the depth-masked self-training strategy of Section \ref{sec:output_level}, where the teacher model is updated every $100$ steps with update momentum $0.99$.
The FFT style transfer parameters are empirically chosen as $\beta=0.01$ for color and $\beta=0.09$ for depth.

\selectcolormodel{rgb}
\definecolor{road}{rgb}{.502,.251,.502}
\definecolor{sidewalk}{rgb}{.957,.137,.910}
\definecolor{building}{rgb}{.275,.275,.275}
\definecolor{wall}{rgb}{.4,.4,.612}
\definecolor{fence}{rgb}{.745,.6,.6}
\definecolor{pole}{rgb}{.6,.6,.6}
\definecolor{tlight}{rgb}{.980,.667,.118}
\definecolor{tsign}{rgb}{.863,.863,0}
\definecolor{vegetation}{rgb}{.420,.557,.137}
\definecolor{terrain}{rgb}{.596,.984,.596}
\definecolor{sky}{rgb}{0,.510,.706}
\definecolor{person}{rgb}{.863,.078,.235}
\definecolor{rider}{rgb}{1,0,0}
\definecolor{car}{rgb}{0,0,.557}
\definecolor{truck}{rgb}{0,0,.275}
\definecolor{bus}{rgb}{0,.235,.392}
\definecolor{train}{rgb}{0,.314,.392}
\definecolor{motorbike}{rgb}{0,0,.902}
\definecolor{bicycle}{rgb}{.467,.043,.125}
\definecolor{unlabelled}{rgb}{0,0,0}
\definecolor{backgroud}{rgb}{0.494,0.545,0.655}
\definecolor{static_object}{rgb}{0.635,0.561,0.447}
\definecolor{moving_object}{rgb}{0.678,0.102,0.396}
\definecolor{signage}{rgb}{0.922,0.769,0.082}
\definecolor{barrier}{rgb}{0.596,0.510,0.604}
\definecolor{twowheels}{rgb}{0.329,0.027,0.643}
\definecolor{personaltransport}{rgb}{0.000,0.000,0.439}
\definecolor{publictransport}{rgb}{0.000,0.275,0.392}
\definecolor{pavement}{rgb}{0.761,0.200,0.733}
\definecolor{ground_m}{rgb}{0.514,0.796,0.431}
\definecolor{thinobject}{rgb}{0.776,0.686,0.427}
\definecolor{structure}{rgb}{0.463,0.408,0.467}
\definecolor{human}{rgb}{0.933,0.055,0.165}
\definecolor{vehicle}{rgb}{0.231,0.137,0.537}

\begin{figure}[tbp]
  \centering
  \begin{subfigure}[b]{0.235\textwidth}
    \caption*{RGB}
    \vspace{-1mm}
    \includegraphics[width=\textwidth]{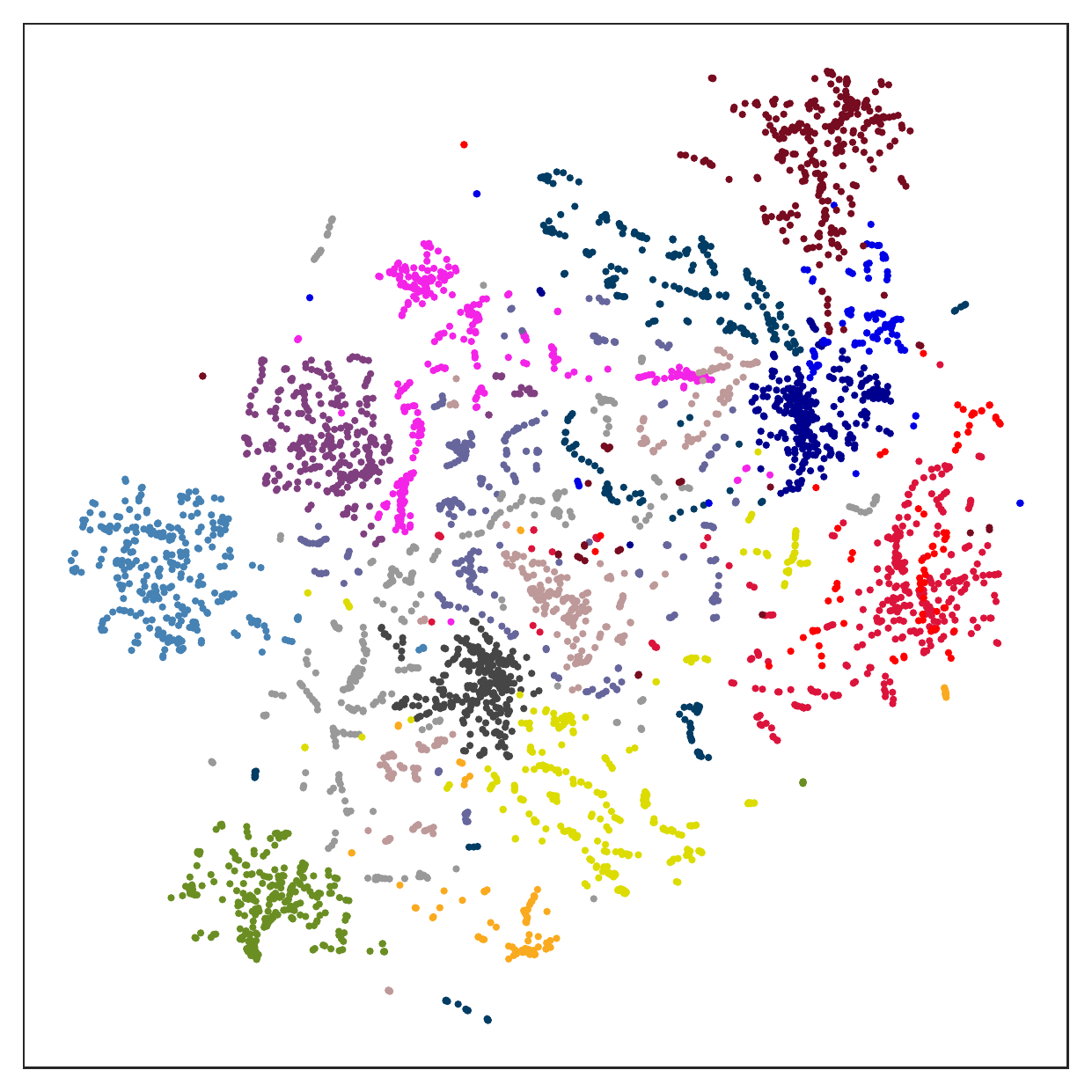}
  \end{subfigure}
  \hfill
  \begin{subfigure}[b]{0.235\textwidth}
    \caption*{RGB-D}
        \vspace{-1mm}
    \includegraphics[width=\textwidth]{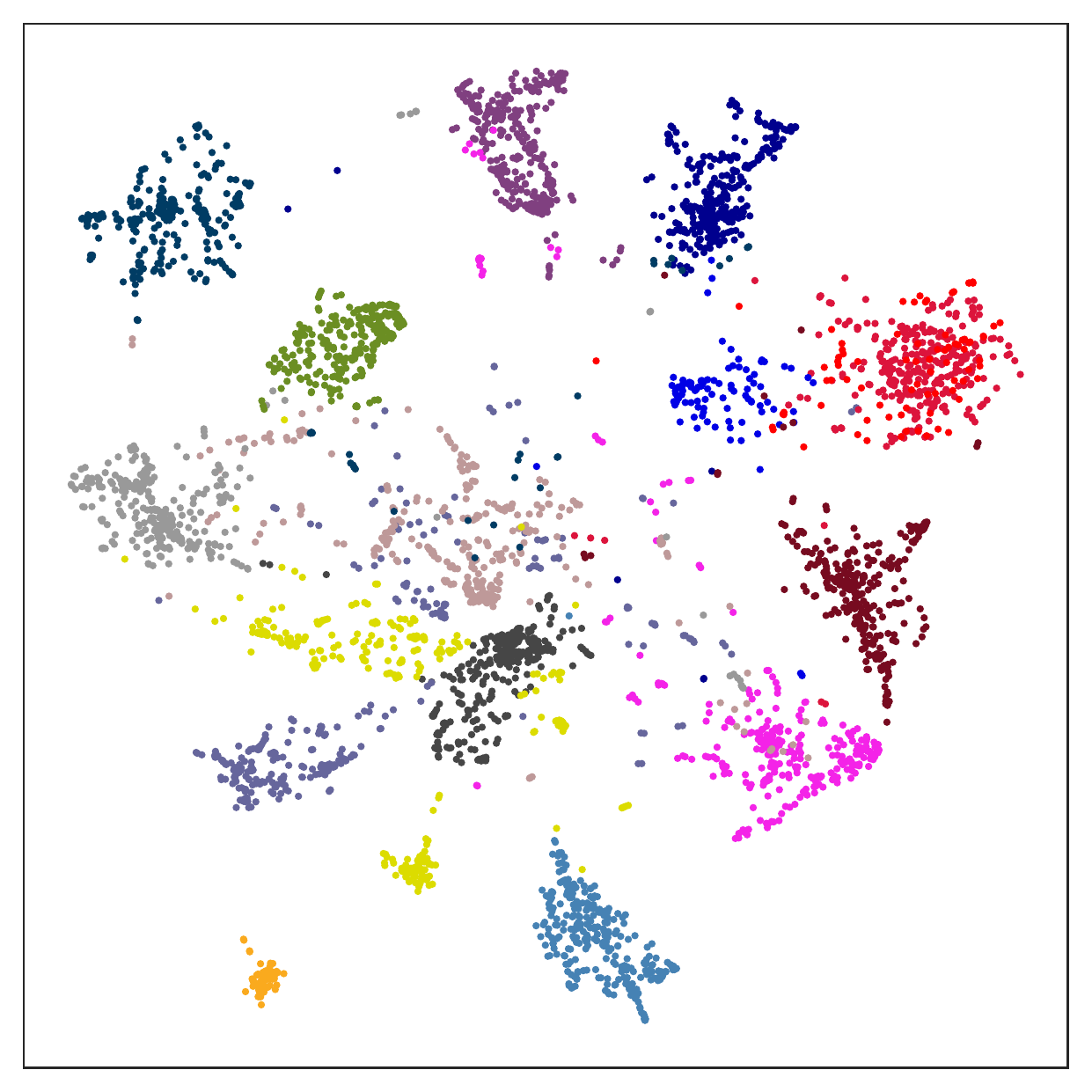}
  \end{subfigure}
\begin{subfigure}{0.467\textwidth}
    \tiny
        \vspace{-1mm}
    \begin{tabular}{cccccccc}
\cellcolor{road} \textcolor{white}{road} & \cellcolor{sidewalk} \textcolor{white}{sidewalk}  & \cellcolor{building} \textcolor{white}{building} & \cellcolor{wall} \textcolor{white}{wall} & \cellcolor{fence} \textcolor{black}{fence} &  \cellcolor{pole} \textcolor{white}{pole} & \cellcolor{tlight} \textcolor{black}{traffic light} & \cellcolor{tsign} \textcolor{black}{traffic sign} \\ \cellcolor{vegetation} \textcolor{white}{vegetation} & \cellcolor{sky} \textcolor{white}{sky} &  \cellcolor{person} \textcolor{white}{person} & \cellcolor{rider} \textcolor{white}{rider} & \cellcolor{car} \textcolor{white}{car} & \cellcolor{bus} \textcolor{white}{bus} &  \cellcolor{motorbike} \textcolor{white}{mbike} & \cellcolor{bicycle} \textcolor{white}{bicycle} 
\end{tabular}
\end{subfigure}
  \caption{T-SNE on the SYNTHIA-to-Cityscapes setting: (Left) RGB-only adaptation (Right) RGB-D adaptation from our method MISFIT.}
  \label{fig:tsne}
\end{figure}

\subsection{SYNTHIA-to-Cityscapes adaptation results}
Table \ref{tab:synthia} shows the performances of our approach in the SYNTHIA-to-Cityscapes benchmark and compares it with the other source-free domain adaptation approaches from the literature. To showcase the effectiveness of our method, we compared it with several recent convolutional architecture-based methods. The table presents the state-of-the-art results for source-free domain adaptation, highlighting the superiority of our proposed approach.

In this setting, training on source color data leads to a relatively low accuracy of approximately 37\%. Even when incorporating FDA-style transfer, there is only a marginal improvement observed, reaching up to 39.2\%.
On the other hand, our approach, which leverages multimodal domain adaptation techniques, achieves remarkable results, as evidenced by an impressive mIoU score of 54.5\%. This performance surpasses most of the competing methods by a significant margin of over 10\%. %
Furthermore, our method demonstrates consistent and outstanding performance across different classes, particularly excelling in classes such as bus, pole, and wall. Notably, our scores in the wall class are more than double the best scores achieved by our competitors.

The visual results align with the numerical evaluation, as evident in Figure \ref{fig:syn_qualitative}, where challenging objects such as the bike with the rider and the poles are accurately identified. At the same time, the domain shift causes issues on textured areas like the road or the sidewalk that need all the components of the proposed approach to be correctly handled. Notably, as observed in the Figure, the regions in question encounter difficulties when only certain components are employed.
This outcome is reinforced by the improved disambiguation observed between the road and sidewalk classes in the T-SNE representation (see Figure \ref{fig:tsne}). Moreover, the numerical results from Table \ref{tab:synthia} validate the improved class separation observed for the driving classes such as bus, motorbike, and bike.

\subsection{SELMA-to-Cityscapes adaptation results}
\label{sec:selma}

The evaluation of the SELMA-to-Cityscapes benchmark is presented in Table \ref{tab:selma}. On this recent dataset there are no results for the source-free setting from previous works, so we can compare the performances of our approach only with some baselines. 
The training on source color data in this setting leads to an accuracy of $41.4\%$ and the FDA style transfer on color data has a very limited impact (the gain is $0.2\%$). %
The higher performance observed when training on the source data, combined with the limited impact of style transfer, indicates that the dataset, compared to other benchmarks, exhibits a higher visual quality that closely aligns with real-world data.
Our multimodal domain adaptation approach allows to increase the mIoU to $51.7\%$ with a remarkable gain of more than $10\%$ over the RGB baseline. Furthermore, it is worth mentioning that the depth-driven entropy minimization loss exhibits a slightly larger impact. This is particularly significant considering that while the performance may be comparatively lower in the SYNTHIA-to-Cityscapes scenario, SELMA encompasses all 19 classes present in Cityscapes.

\begin{figure}[t]
\vspace{-3mm}
    \centering
    \includegraphics[width=0.4\textwidth]{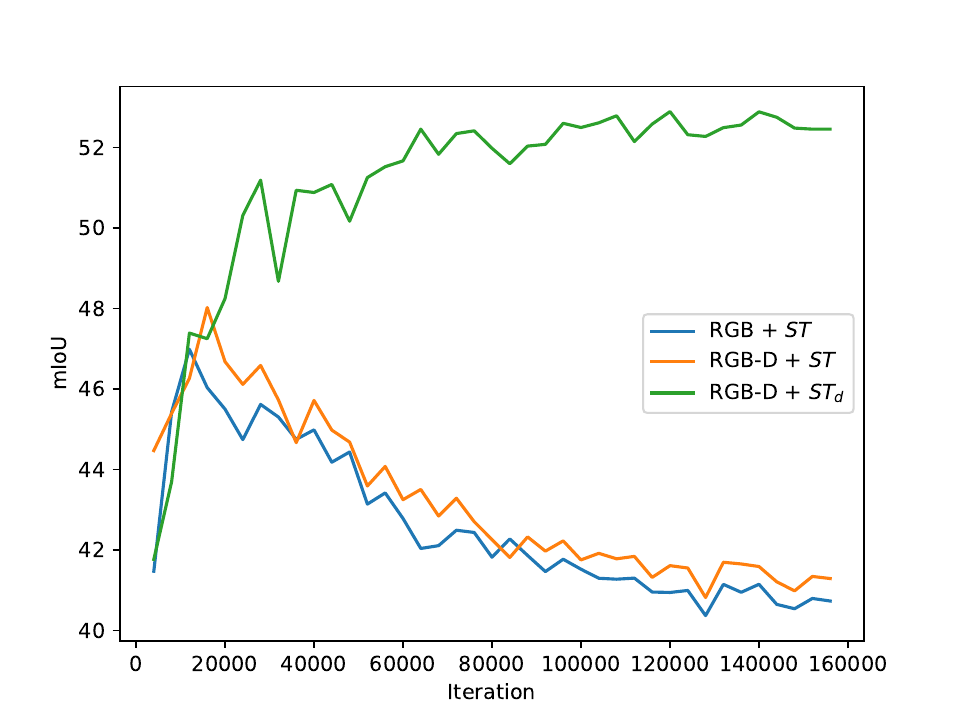}
    \caption{Comparison of learning curves for various self-training setups: standard pseudo-label masking based on confidence scores %
    and $ST_d$ employing depth-based masking on top.}
    \label{fig:plot_ST}
\end{figure}

\subsection{Ablation Studies} 
\label{sec:ablation}

\textbf{Modules of the proposed framework}
First of all, we performed some ablation studies on the SYNTHIA-to-Cityscapes benchmark to evaluate the impact of the various modules of the proposed approach on the final performances. Results are shown in Table \ref{tab:ablation_adaptation}: 
as already pointed out simply training on source color data and testing on the target dataset leads to an accuracy of around $37\%$, which represents the starting point. Adding a self-teaching step with target color data allows improving performances of almost $4\%$.
Moving to the multimodal setting, the source-only training accuracy exploiting depth data is $39.8\%$. The depth-aware self-teaching scheme proposed in Section \ref{sec:output_level} allows for an impressive improvement up to $52.5\%$ of mIoU. In particular, as visible in Figure \ref{fig:plot_ST}, incorporating depth-based masking serves as a regularization method, enhancing training performance compared to the ones on confidence only (refer to the \textit{Supplementary Material} for details). During standard self-training, the approach initially learns effectively, reaching an accuracy of $48.0\%$. Nevertheless, it becomes excessively confident in inaccurate predictions, leading to a decline in the learning curve —a trend consistent with observations in \cite{fleuret2021uncertainty}. Further adding the input level style transfer boosts performances to $54\%$. Finally adding also the entropy minimization target allows us to get the full model accuracy of $54.5\%$.

\begin{table}[htbp]
    \centering
    \begin{tabular}{c|cccc|c}
    \toprule
         \textbf{Mode}  & \textbf{ST} & $\textrm{\textbf{ST}}_{d}$ & 
    \textbf{Style}  &  \textbf{Entropy} & \textbf{mIoU} \\
        \midrule
        RGB &  &  &  &  & 36.93\\
        RGB & \cmark &  &  &  & 40.59\\
        \midrule
        RGB+D &  &  &  & & 39.79\\
        RGB+D &  & \cmark & & & 52.50\\
        RGB+D &  &  \cmark & \cmark & & 54.00\\
        RGB+D &  & \cmark & \cmark & \cmark & \textbf{54.52}\\
    \bottomrule
    \end{tabular}
        \caption{Impact of the various proposed adaptation framework modules, performed on the SYNTHIA-to-Cityscapes setting. When all modules are disabled, it corresponds to source-only. %
        }
    \label{tab:ablation_adaptation}
\end{table}

\textbf{Cross-Modality Attention} 
\label{sec:abl_crossatt}
We conducted tests on the impact of crossing between the two modalities (i.e., color and depth) in the source-only setting. %
Supported by the fact that previous works exploit asymmetric feature fusion \cite{rizzoli2022multimodal}, we tested the transformer cross-attention swap in the direction of color \cite{tsai2019multimodal} and the key-swap algorithm \cite{barbato2022depthformer}.
The key-swapping strategy achieves better results and for this reason, it has been selected for our approach. 
More in detail, the informative content of the depth keys is able to gain a $2.86\%$ over the use of color alone in the source-only setup (see Table \ref{tab:ablation_swap}).

\begin{table}[tbp]
    \centering
    \begin{tabular}{c|c|c}
    \toprule
         \textbf{Mode}  & \textbf{Cross-Attention} &\textbf{mIoU} \\
        \midrule
        RGB &  & 36.93\\
        \midrule
        RGB+D &  RGB$\rightarrow$D  &  29.83 \\ %
        RGB+D &  D$\rightarrow$RGB &  38.02 \\ %
        RGB+D &  Key Swap & \textbf{39.79} \\ %
    \bottomrule
    \end{tabular}
        \caption{Ablation on source-only generalization ability of cross-modalities attention, performed on the SYNTHIA-to-Cityscapes setting.%
        }
    \label{tab:ablation_swap}
    \vspace{-5pt}
\end{table}

\begin{table}[tbp]
\centering
\begin{tabular}{c|c|c}
\toprule
 \textbf{Mode} & $\textrm{\textbf{Style-Transfer}}$ & \textbf{mIoU} \\ 
\toprule
   RGB &  & 36.9 \\ 
   RGB & \checkmark & 39.2 \\ %
   \hdashline
   RGB+D &  & 39.8 \\  %
   RGB+D & \checkmark & \textbf{41.0} \\    %
 \bottomrule
\end{tabular}
\caption{Ablation on source-only generalization ability with different input-level style-transfers, performed on the SYNTHIA-to-Cityscapes setting. %
}
\label{tab:abl_fft}
\end{table}

\textbf{Input Depth Style}
In order to prove the effectiveness of the style transfer method we compare the performances of the algorithm described in Section~\ref{sec:input_level} when applying it to color data or to both modalities in the SYNTHIA-to-Cityscapes benchmark. %
The target style is transferred to the source domain to perform network pre-training. %
Results are shown in Table \ref{tab:abl_fft}: if working with color data alone the style transfer allows for a gain of around $2\%$. Multimodal data allows for a higher starting point and adding the style transfer on both modalities allows to further boost performances from $39.8$ to $41\%$.

\section{Conclusions}
\label{sec:conclusions}
Ultimately, the use of multimodal information for semantic segmentation is a relevant area of research that can help to address the challenges of adapting segmentation models to new domains. However, the domain adaptation capabilities of \giulia{pre-trained} multimodal schemes have seldom been explored, especially in conjunction with vision transformer architectures that represent the current state-of-the-art in many vision tasks.
By leveraging multiple adaptation strategies driven by the complementary information provided by depth data, the proposed multimodal framework allows for improving the robustness and generalization ability of segmentation models, enabling them to be used in a wider range of applications. Experimental results show how it achieves state-of-the-art performance in the challenging source-free domain adaptation setting.

Further research will be devoted to improving the exploitation of depth data in transformer-based segmentation models and to the development of domain adaptation strategies explicitly targeted at the inconsistencies between ground truth and estimated depth data.

\subsection*{Acknowledgment}
This work was supported in part by the European Union through the Italian National Recovery and Resilience Plan (NRRP) of NextGenerationEU, partnership on “Telecommunications of the Future” (Program “RESTART”) under Grant PE0000001.

{\small
\bibliographystyle{ieee_fullname}
\bibliography{egbib}

\begin{thebibliography}{10}\itemsep=-1pt

\bibitem{agarwal2022unsupervised}
Peshal Agarwal, Danda~Pani Paudel, Jan-Nico Zaech, and Luc Van~Gool.
\newblock Unsupervised robust domain adaptation without source data.
\newblock In {\em Proceedings of the IEEE/CVF Winter Conference on Applications
  of Computer Vision}, pages 2009--2018, 2022.

\bibitem{barbato2022depthformer}
Francesco Barbato, Giulia Rizzoli, and Pietro Zanuttigh.
\newblock Depthformer: Multimodal positional encodings and cross-input
  attention for transformer-based segmentation networks.
\newblock In {\em ICASSP 2023-2023 IEEE International Conference on Acoustics,
  Speech and Signal Processing (ICASSP)}, pages 1--5. IEEE, 2023.

\bibitem{chen2019domain}
Minghao Chen, Hongyang Xue, and Deng Cai.
\newblock Domain adaptation for semantic segmentation with maximum squares
  loss.
\newblock In {\em Proceedings of the IEEE/CVF International Conference on
  Computer Vision}, pages 2090--2099, 2019.

\bibitem{cordts2016cityscapes}
Marius Cordts, Mohamed Omran, Sebastian Ramos, Timo Rehfeld, Markus Enzweiler,
  Rodrigo Benenson, Uwe Franke, Stefan Roth, and Bernt Schiele.
\newblock The cityscapes dataset for semantic urban scene understanding.
\newblock In {\em Proceedings of the IEEE conference on computer vision and
  pattern recognition}, pages 3213--3223, 2016.

\bibitem{fleuret2021uncertainty}
Francois Fleuret et~al.
\newblock Uncertainty reduction for model adaptation in semantic segmentation.
\newblock In {\em Proceedings of the IEEE/CVF Conference on Computer Vision and
  Pattern Recognition}, pages 9613--9623, 2021.

\bibitem{hoffman2018cycada}
Judy Hoffman, Eric Tzeng, Taesung Park, Jun-Yan Zhu, Phillip Isola, Kate
  Saenko, Alexei Efros, and Trevor Darrell.
\newblock Cycada: Cycle-consistent adversarial domain adaptation.
\newblock In {\em International conference on machine learning}, pages
  1989--1998. Pmlr, 2018.

\bibitem{hoyer2022daformer}
Lukas Hoyer, Dengxin Dai, and Luc Van~Gool.
\newblock Daformer: Improving network architectures and training strategies for
  domain-adaptive semantic segmentation.
\newblock In {\em Proceedings of the IEEE/CVF Conference on Computer Vision and
  Pattern Recognition}, pages 9924--9935, 2022.

\bibitem{hoyer2022hrda}
Lukas Hoyer, Dengxin Dai, and Luc Van~Gool.
\newblock Hrda: Context-aware high-resolution domain-adaptive semantic
  segmentation.
\newblock In {\em Computer Vision--ECCV 2022: 17th European Conference, Tel
  Aviv, Israel, October 23--27, 2022, Proceedings, Part XXX}, pages 372--391.
  Springer, 2022.

\bibitem{hu2023multi}
Sijie Hu, Fabien Bonardi, Samia Bouchafa, and D{\'e}sir{\'e} Sidib{\'e}.
\newblock Multi-modal unsupervised domain adaptation for semantic image
  segmentation.
\newblock {\em Pattern Recognition}, page 109299, 2023.

\bibitem{huang2021model}
Jiaxing Huang, Dayan Guan, Aoran Xiao, and Shijian Lu.
\newblock Model adaptation: Historical contrastive learning for unsupervised
  domain adaptation without source data.
\newblock {\em Advances in Neural Information Processing Systems},
  34:3635--3649, 2021.

\bibitem{jaritz2022cross}
Maximilian Jaritz, Tuan-Hung Vu, Raoul De~Charette, {\'E}milie Wirbel, and
  Patrick P{\'e}rez.
\newblock Cross-modal learning for domain adaptation in 3d semantic
  segmentation.
\newblock {\em IEEE Transactions on Pattern Analysis and Machine Intelligence},
  45(2):1533--1544, 2022.

\bibitem{kolesnikov2021image}
Alexander Kolesnikov, Alexey Dosovitskiy, Dirk Weissenborn, Georg Heigold,
  Jakob Uszkoreit, Lucas Beyer, Matthias Minderer, Mostafa Dehghani, Neil
  Houlsby, Sylvain Gelly, et~al.
\newblock An image is worth 16x16 words: Transformers for image recognition at
  scale, 2021.

\bibitem{kundu2021generalize}
Jogendra~Nath Kundu, Akshay Kulkarni, Amit Singh, Varun Jampani, and
  R~Venkatesh Babu.
\newblock Generalize then adapt: Source-free domain adaptive semantic
  segmentation.
\newblock In {\em Proceedings of the IEEE/CVF International Conference on
  Computer Vision}, pages 7046--7056, 2021.

\bibitem{liu2022cmx}
Huayao Liu, Jiaming Zhang, Kailun Yang, Xinxin Hu, and Rainer Stiefelhagen.
\newblock Cmx: Cross-modal fusion for rgb-x semantic segmentation with
  transformers.
\newblock {\em arXiv preprint arXiv:2203.04838}, 2022.

\bibitem{liu2021source}
Yuang Liu, Wei Zhang, and Jun Wang.
\newblock Source-free domain adaptation for semantic segmentation.
\newblock In {\em Proceedings of the IEEE/CVF Conference on Computer Vision and
  Pattern Recognition}, pages 1215--1224, 2021.

\bibitem{luo2019taking}
Yawei Luo, Liang Zheng, Tao Guan, Junqing Yu, and Yi Yang.
\newblock Taking a closer look at domain shift: Category-level adversaries for
  semantics consistent domain adaptation.
\newblock In {\em Proceedings of the IEEE/CVF conference on computer vision and
  pattern recognition}, pages 2507--2516, 2019.

\bibitem{pan2020unsupervised}
Fei Pan, Inkyu Shin, Francois Rameau, Seokju Lee, and In~So Kweon.
\newblock Unsupervised intra-domain adaptation for semantic segmentation
  through self-supervision.
\newblock In {\em Proceedings of the IEEE/CVF Conference on Computer Vision and
  Pattern Recognition}, pages 3764--3773, 2020.

\bibitem{park2022dat}
Jinyoung Park, Minseok Son, Sumin Lee, and Changick Kim.
\newblock Dat: Domain adaptive transformer for domain adaptive semantic
  segmentation.
\newblock In {\em 2022 IEEE International Conference on Image Processing
  (ICIP)}, pages 4183--4187. IEEE, 2022.

\bibitem{peng2022toward}
Qucheng Peng, Zhengming Ding, Lingjuan Lyu, Lichao Sun, and Chen Chen.
\newblock Toward better target representation for source-free and black-box
  domain adaptation.
\newblock {\em arXiv preprint arXiv:2208.10531}, 2022.

\bibitem{richter2016playing}
Stephan~R Richter, Vibhav Vineet, Stefan Roth, and Vladlen Koltun.
\newblock Playing for data: Ground truth from computer games.
\newblock In {\em Computer Vision--ECCV 2016: 14th European Conference,
  Amsterdam, The Netherlands, October 11-14, 2016, Proceedings, Part II 14},
  pages 102--118. Springer, 2016.

\bibitem{rizzoli2022multimodal}
Giulia Rizzoli, Francesco Barbato, and Pietro Zanuttigh.
\newblock Multimodal semantic segmentation in autonomous driving: A review of
  current approaches and future perspectives.
\newblock {\em Technologies}, 10(4):90, 2022.

\bibitem{ros2016synthia}
German Ros, Laura Sellart, Joanna Materzynska, David Vazquez, and Antonio~M
  Lopez.
\newblock The synthia dataset: A large collection of synthetic images for
  semantic segmentation of urban scenes.
\newblock In {\em Proceedings of the IEEE conference on computer vision and
  pattern recognition}, pages 3234--3243, 2016.

\bibitem{seichter2021efficient}
Daniel Seichter, Mona K{\"o}hler, Benjamin Lewandowski, Tim Wengefeld, and
  Horst-Michael Gross.
\newblock Efficient rgb-d semantic segmentation for indoor scene analysis.
\newblock In {\em 2021 IEEE International Conference on Robotics and Automation
  (ICRA)}, pages 13525--13531. IEEE, 2021.

\bibitem{shenaj2023learning}
Donald Shenaj, Eros Fan{\`\i}, Marco Toldo, Debora Caldarola, Antonio Tavera,
  Umberto Michieli, Marco Ciccone, Pietro Zanuttigh, and Barbara Caputo.
\newblock Learning across domains and devices: Style-driven source-free domain
  adaptation in clustered federated learning.
\newblock In {\em Proceedings of the IEEE/CVF Winter Conference on Applications
  of Computer Vision}, pages 444--454, 2023.

\bibitem{shin2022mm}
Inkyu Shin, Yi-Hsuan Tsai, Bingbing Zhuang, Samuel Schulter, Buyu Liu, Sparsh
  Garg, In~So Kweon, and Kuk-Jin Yoon.
\newblock Mm-tta: multi-modal test-time adaptation for 3d semantic
  segmentation.
\newblock In {\em Proceedings of the IEEE/CVF Conference on Computer Vision and
  Pattern Recognition}, pages 16928--16937, 2022.

\bibitem{testolina2023selma}
Paolo Testolina, Francesco Barbato, Umberto Michieli, Marco Giordani, Pietro
  Zanuttigh, and Michele Zorzi.
\newblock Selma: Semantic large-scale multimodal acquisitions in variable
  weather, daytime and viewpoints.
\newblock {\em IEEE Transactions on Intelligent Transportation Systems}, 2023.

\bibitem{toldo2020unsupervised}
Marco Toldo, Andrea Maracani, Umberto Michieli, and Pietro Zanuttigh.
\newblock Unsupervised domain adaptation in semantic segmentation: a review.
\newblock {\em Technologies}, 8(2):35, 2020.

\bibitem{tranheden2021dacs}
Wilhelm Tranheden, Viktor Olsson, Juliano Pinto, and Lennart Svensson.
\newblock Dacs: Domain adaptation via cross-domain mixed sampling.
\newblock In {\em Proceedings of the IEEE/CVF Winter Conference on Applications
  of Computer Vision}, pages 1379--1389, 2021.

\bibitem{tsai2019domain}
Yi-Hsuan Tsai, Kihyuk Sohn, Samuel Schulter, and Manmohan Chandraker.
\newblock Domain adaptation for structured output via discriminative patch
  representations.
\newblock In {\em Proceedings of the IEEE/CVF International Conference on
  Computer Vision}, pages 1456--1465, 2019.

\bibitem{tsai2019multimodal}
Yao-Hung~Hubert Tsai, Shaojie Bai, Paul~Pu Liang, J~Zico Kolter, Louis-Philippe
  Morency, and Ruslan Salakhutdinov.
\newblock Multimodal transformer for unaligned multimodal language sequences.
\newblock In {\em Proceedings of the conference. Association for Computational
  Linguistics. Meeting}, volume 2019, page 6558. NIH Public Access, 2019.

\bibitem{valada2020self}
Abhinav Valada, Rohit Mohan, and Wolfram Burgard.
\newblock Self-supervised model adaptation for multimodal semantic
  segmentation.
\newblock {\em International Journal of Computer Vision}, 128(5):1239--1285,
  2020.

\bibitem{vaswani2017attention}
Ashish Vaswani, Noam Shazeer, Niki Parmar, Jakob Uszkoreit, Llion Jones,
  Aidan~N Gomez, {\L}ukasz Kaiser, and Illia Polosukhin.
\newblock Attention is all you need.
\newblock {\em Advances in neural information processing systems}, 30, 2017.

\bibitem{vu2018advent}
Tuan-Hung Vu, Himalaya Jain, Maxime Bucher, Mathieu Cord, and Patrick
  P{\'e}rez.
\newblock Advent: Adversarial entropy minimization for domain adaptation in
  semantic segmentation.
\newblock In {\em CVPR}, 2019.

\bibitem{xie2021segformer}
Enze Xie, Wenhai Wang, Zhiding Yu, Anima Anandkumar, Jose~M Alvarez, and Ping
  Luo.
\newblock Segformer: Simple and efficient design for semantic segmentation with
  transformers.
\newblock {\em Advances in Neural Information Processing Systems}, 34, 2021.

\bibitem{yang2022source}
Cheng-Yu Yang, Yuan-Jhe Kuo, and Chiou-Ting Hsu.
\newblock Source free domain adaptation for semantic segmentation via
  distribution transfer and adaptive class-balanced self-training.
\newblock In {\em 2022 IEEE International Conference on Multimedia and Expo
  (ICME)}, pages 1--6. IEEE, 2022.

\bibitem{yang2020fda}
Yanchao Yang and Stefano Soatto.
\newblock Fda: Fourier domain adaptation for semantic segmentation.
\newblock In {\em Proceedings of the IEEE/CVF Conference on Computer Vision and
  Pattern Recognition}, pages 4085--4095, 2020.

\bibitem{ye2021source}
Mucong Ye, Jing Zhang, Jinpeng Ouyang, and Ding Yuan.
\newblock Source data-free unsupervised domain adaptation for semantic
  segmentation.
\newblock In {\em Proceedings of the 29th ACM International Conference on
  Multimedia}, pages 2233--2242, 2021.

\bibitem{you2021domain}
Fuming You, Jingjing Li, Lei Zhu, Zhi Chen, and Zi Huang.
\newblock Domain adaptive semantic segmentation without source data.
\newblock In {\em Proceedings of the 29th ACM International Conference on
  Multimedia}, pages 3293--3302, 2021.

\bibitem{yu2020bdd100k}
Fisher Yu, Haofeng Chen, Xin Wang, Wenqi Xian, Yingying Chen, Fangchen Liu,
  Vashisht Madhavan, and Trevor Darrell.
\newblock Bdd100k: A diverse driving dataset for heterogeneous multitask
  learning.
\newblock In {\em Proceedings of the IEEE/CVF conference on computer vision and
  pattern recognition}, pages 2636--2645, 2020.

\bibitem{yu2023comprehensive}
Zhiqi Yu, Jingjing Li, Zhekai Du, Lei Zhu, and Heng~Tao Shen.
\newblock A comprehensive survey on source-free domain adaptation.
\newblock {\em arXiv preprint arXiv:2302.11803}, 2023.

\bibitem{zhang2023delivering}
Jiaming Zhang, Ruiping Liu, Hao Shi, Kailun Yang, Simon Rei{\ss}, Kunyu Peng,
  Haodong Fu, Kaiwei Wang, and Rainer Stiefelhagen.
\newblock Delivering arbitrary-modal semantic segmentation.
\newblock In {\em Proceedings of the IEEE/CVF Conference on Computer Vision and
  Pattern Recognition}, pages 1136--1147, 2023.

\bibitem{zhao2023towards}
Dong Zhao, Shuang Wang, Qi Zang, Dou Quan, Xiutiao Ye, and Licheng Jiao.
\newblock Towards better stability and adaptability: Improve online
  self-training for model adaptation in semantic segmentation.
\newblock In {\em Proceedings of the IEEE/CVF Conference on Computer Vision and
  Pattern Recognition}, pages 11733--11743, 2023.

\bibitem{zou2018unsupervised}
Yang Zou, Zhiding Yu, BVK Kumar, and Jinsong Wang.
\newblock Unsupervised domain adaptation for semantic segmentation via
  class-balanced self-training.
\newblock In {\em Proceedings of the European conference on computer vision
  (ECCV)}, pages 289--305, 2018.

\bibitem{zou2019confidence}
Yang Zou, Zhiding Yu, Xiaofeng Liu, BVK Kumar, and Jinsong Wang.
\newblock Confidence regularized self-training.
\newblock In {\em Proceedings of the IEEE/CVF international conference on
  computer vision}, pages 5982--5991, 2019.

\end{thebibliography}
}

\clearpage

\renewcommand{\thefigure}{S\arabic{figure}}
\renewcommand{\theHfigure}{S\arabic{figure}}

\renewcommand{\thesection}{S\arabic{section}}
\renewcommand{\theHsection}{S\arabic{section}}

\renewcommand{\theequation}{S\arabic{equation}}
\renewcommand{\theHequation}{S\arabic{equation}}

\renewcommand{\thetable}{S\arabic{table}}
\renewcommand{\theHtable}{S\arabic{table}}

\setcounter{equation}{0}
\setcounter{figure}{0}
\setcounter{table}{0}
\setcounter{section}{0}
\clearpage

\begin{strip}
{
\null
\vskip .375in
\begin{center}
    {\Large \bf 
    Source-Free Domain Adaptation for RGB-D Semantic \\Segmentation with Vision Transformers: \\
    \textit{Supplementary Material}
    \par}
      \vspace*{12pt}
  \end{center}
  }
\end{strip}


This document contains supporting material for the paper  \textit{Source-Free Domain Adaptation for RGB-D Semantic Segmentation with Vision Transformers}. Here, we include additional ablation experiments supporting some design choices of the proposed method along with additional qualitative visual results on the SELMA dataset \cite{testolina2023selma}.

\afterpage{
\begin{figure*}[ht]
\begin{subfigure}[b]{\textwidth}
\centering
\begin{subfigure}[b]{0.245\textwidth}
\caption*{Source}
\includegraphics[width=\textwidth]{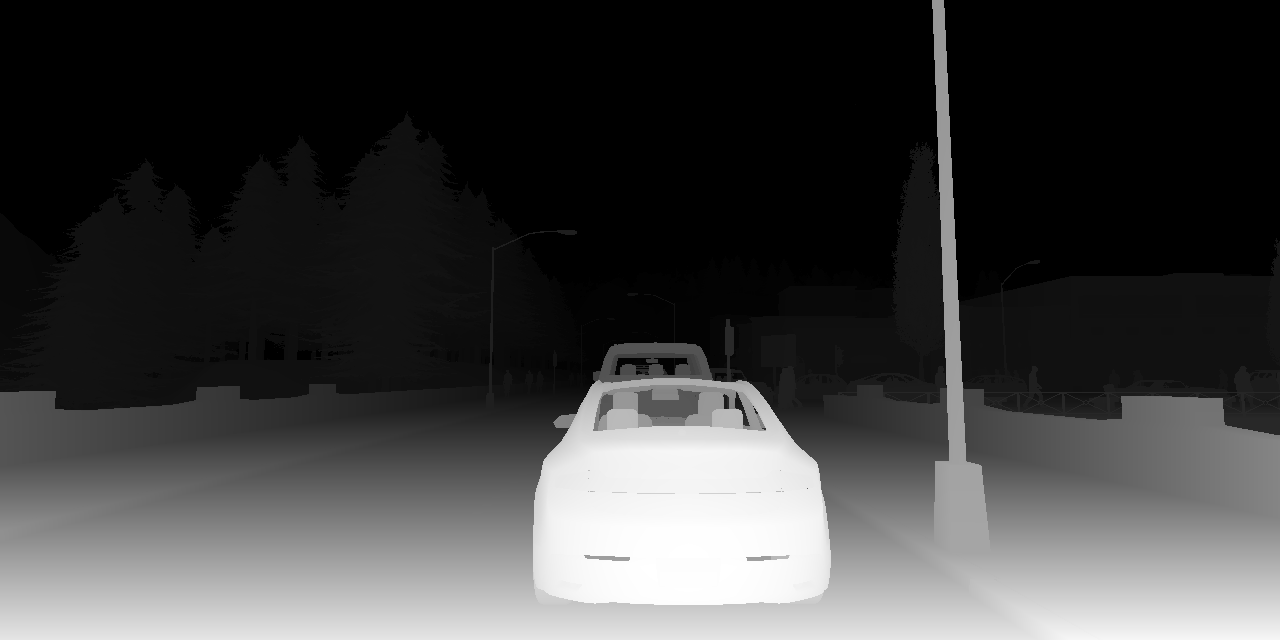}
\end{subfigure} 
\begin{subfigure}[b]{0.245\textwidth}
\caption*{$\beta = 0.01$}
\includegraphics[width=\textwidth]{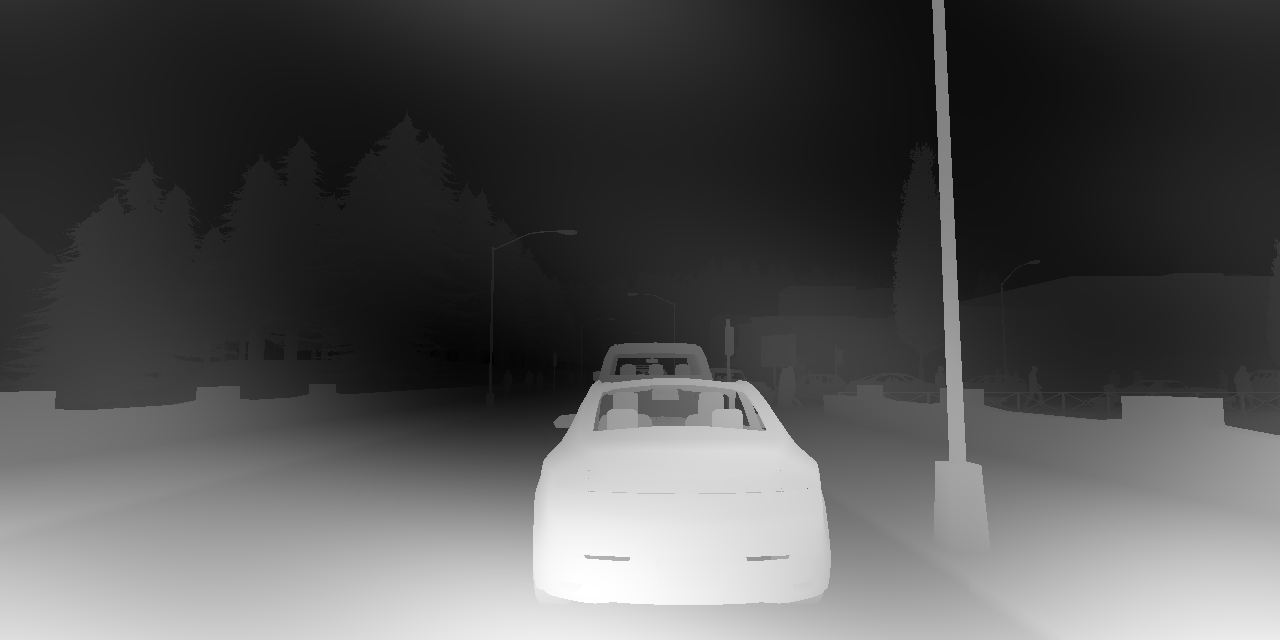}
\end{subfigure} 
\begin{subfigure}[b]{0.245\textwidth}
\caption*{$\beta = 0.03$}
\includegraphics[width=\textwidth]{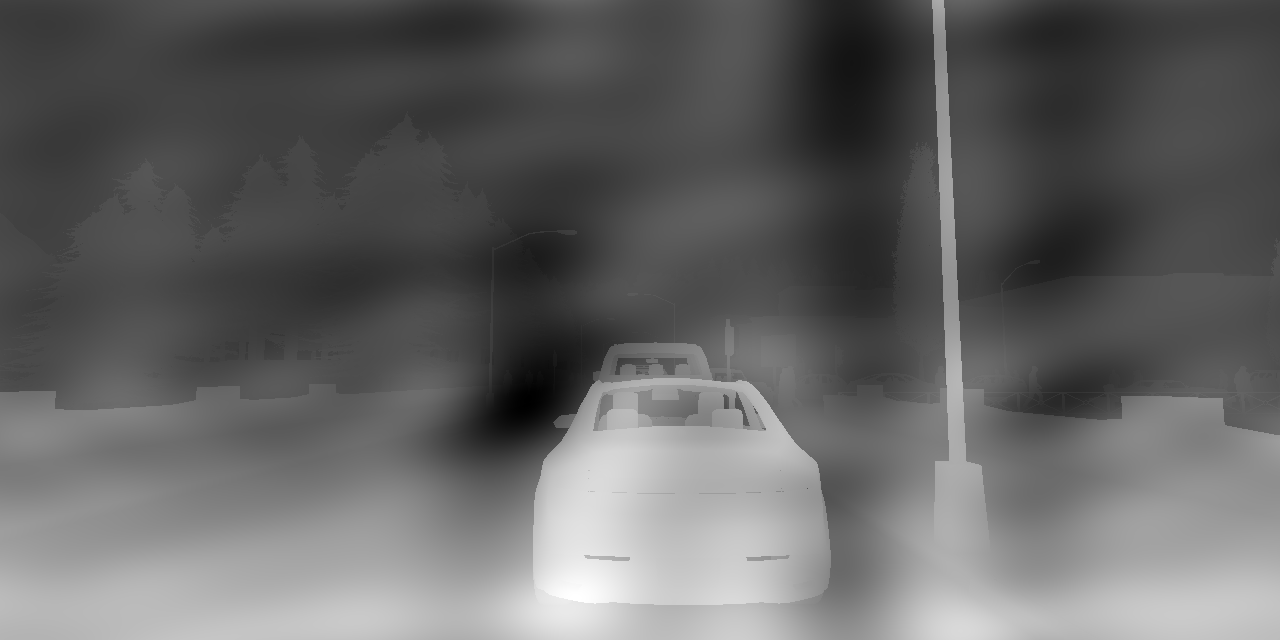}
\end{subfigure}
\begin{subfigure}[b]{0.245\textwidth}
\caption*{$\beta = 0.05$}
\includegraphics[width=\textwidth]{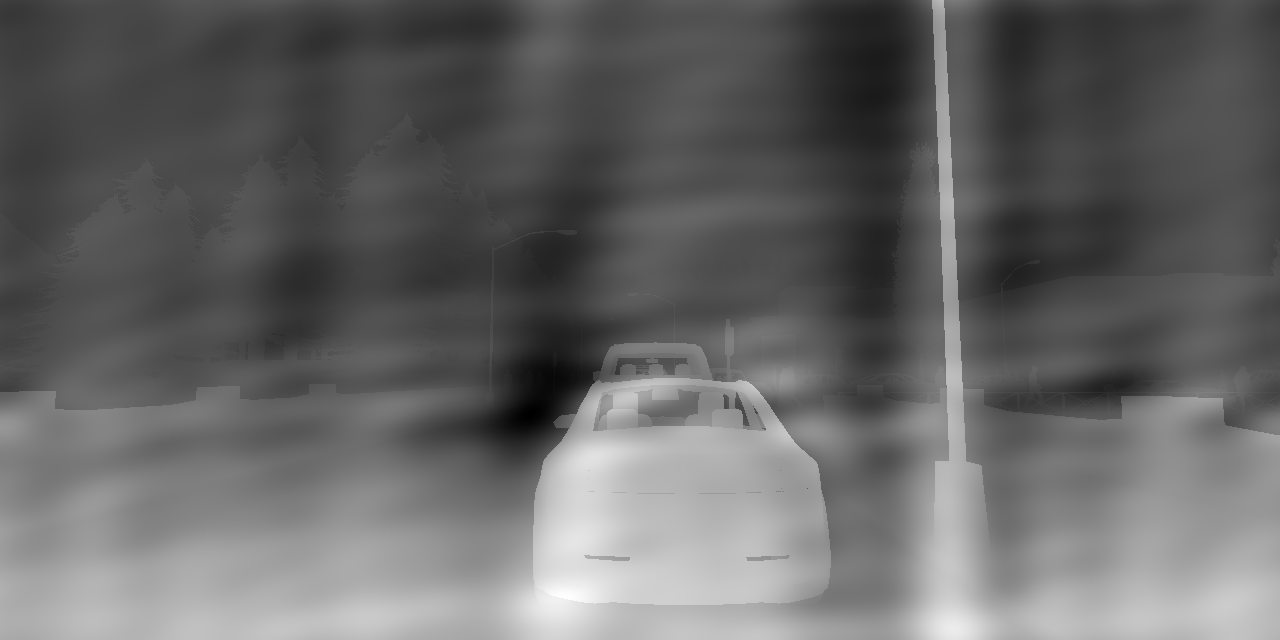}
\end{subfigure}
\end{subfigure}

\begin{subfigure}[b]{\textwidth}
\centering
\begin{subfigure}[b]{0.245\textwidth}
\includegraphics[width=\textwidth]{figures/fda/target_batch_2.png}
\caption*{Target} %
\end{subfigure} 
\begin{subfigure}[b]{0.245\textwidth}
\includegraphics[width=\textwidth]{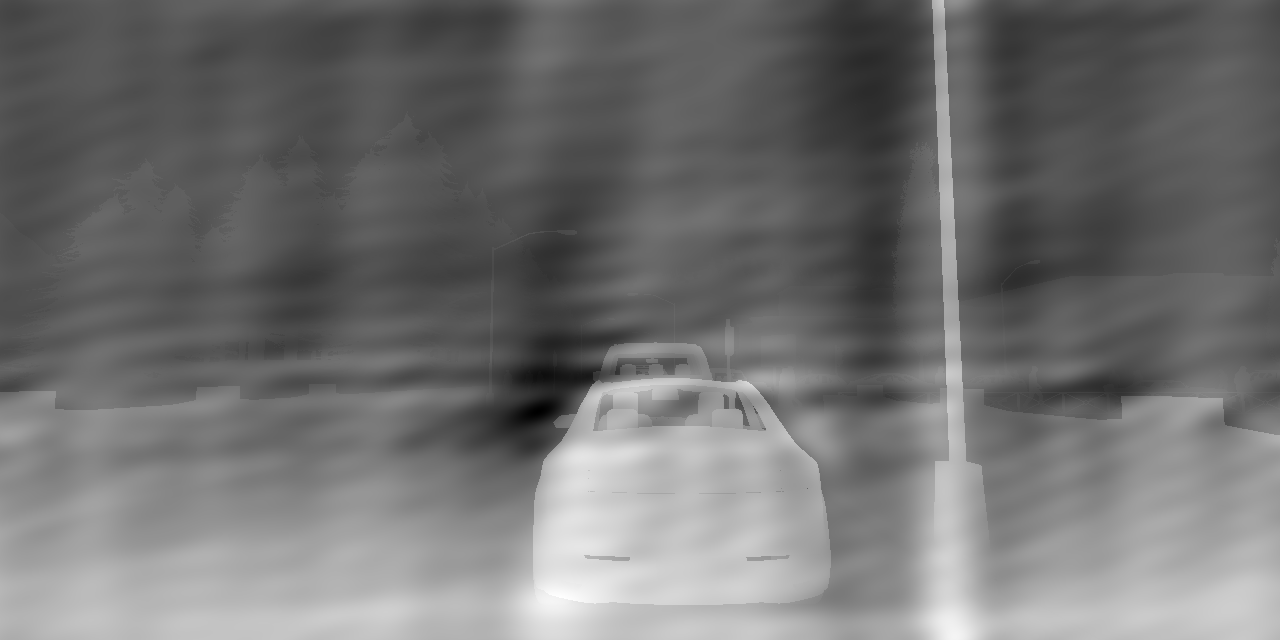}
\caption*{$\beta = 0.07$}
\end{subfigure} 
\begin{subfigure}[b]{0.245\textwidth}
\includegraphics[width=\textwidth]{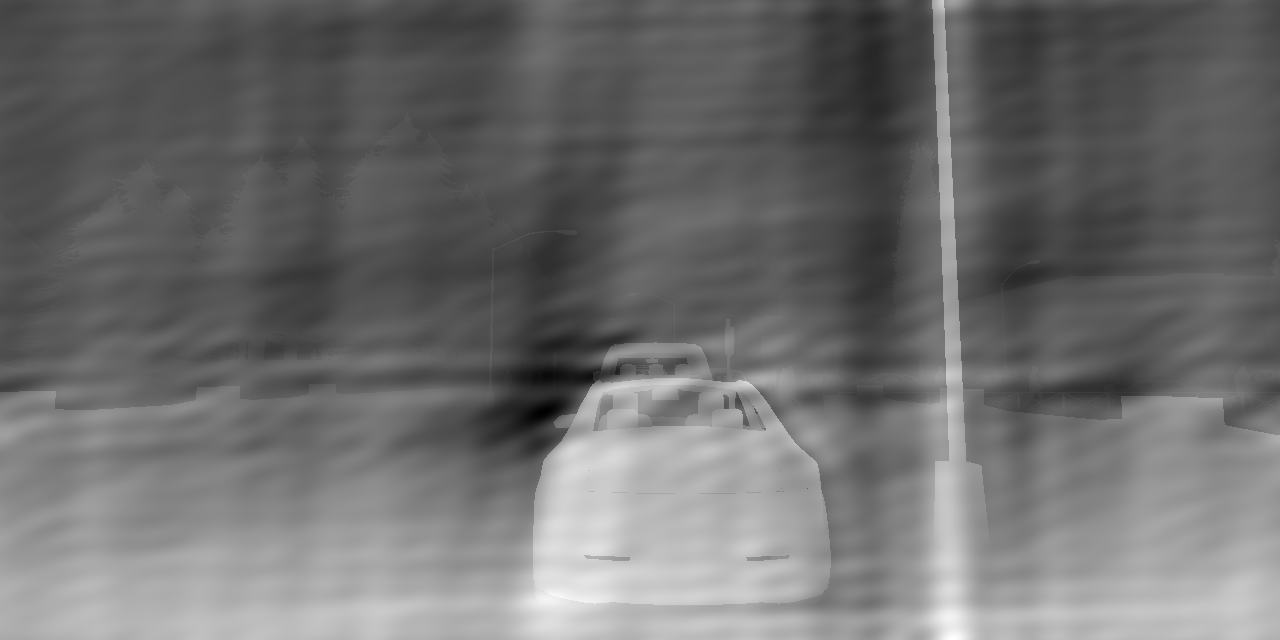}
\caption*{$\beta = 0.09$}
\end{subfigure}
\begin{subfigure}[b]{0.245\textwidth}
\includegraphics[width=\textwidth]{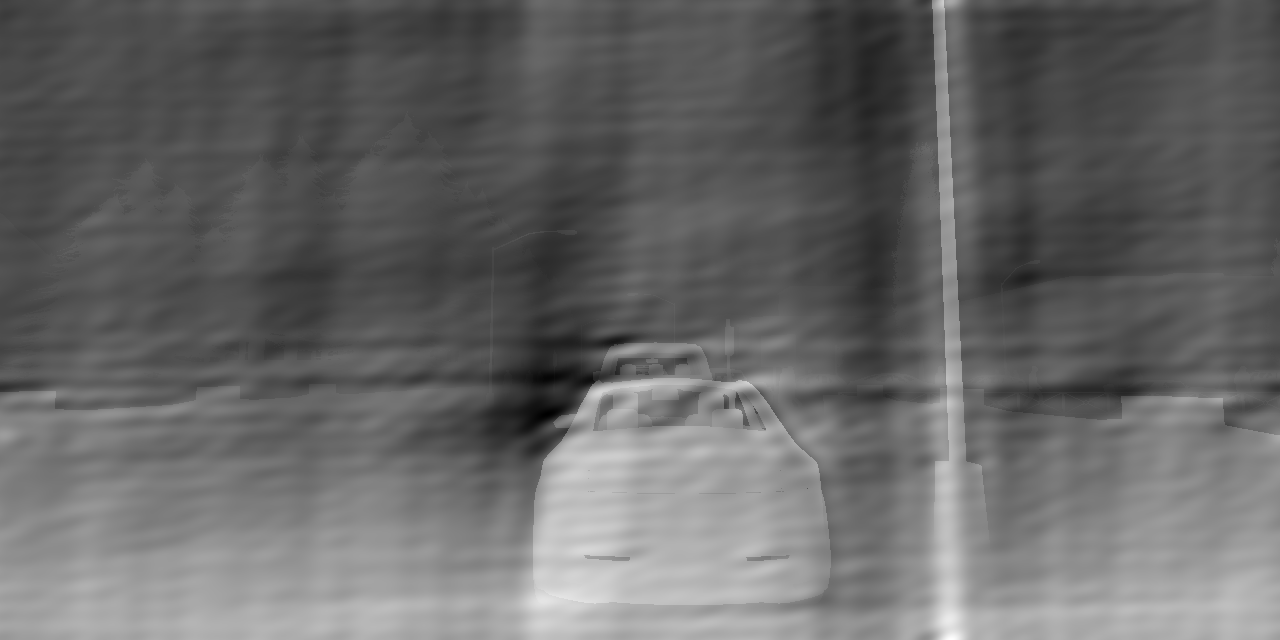}
\caption*{$\beta = 0.12$}
\end{subfigure}
\end{subfigure}
\caption{Effect of the Fourier Domain style transfer applied on depth images from the SELMA dataset, whereas $\beta$ = 0 is equivalent to no transfer and $\beta$ = 1 to the transfer of the full target amplitude.}
\label{fig:fda}
\end{figure*}

}

\section{Tuning on input level adaptation}
In this section, we present some additional results concerning the input level adaptation in the Fourier domain. 

For this strategy, a key parameter is $\beta$ as it selects which region of the amplitude spectrum is going to be replaced with the target style. As pointed out in the paper, larger values of $\beta$ lead to a stronger adaptation effect but also introduce visual distortion. While this aspect has been discussed for standard images in previous works \cite{yang2020fda}, its impact on depth data has never been analyzed. 

The main paper already shows an example relative to the SYNTHIA to Cityscapes adaptation, while Figure \ref{fig:fda} in this document shows an additional example relative to the SELMA to Cityscapes setting.
Notice that, as pointed out in the main paper, the approach allows to better align the depth ranges and matches the fact that real-world target data computed with stereo vision has more artifacts and a less sharp distribution. The example in the image confirms these observations and shows that they are not dataset-dependent. 

Using larger $\beta$  introduces artifacts that are visually disturbing, yet the key question is if they affect also the network performances.
Table \ref{tab:abl_beta} shows the mIoU after the pre-training step for different values of $\beta$. Notice how the relatively large value of $\beta=0.09$ leads to depth maps not very visually appealing but instead effective when used to aid the segmentation model.
The results in Table \ref{tab:abl_beta} refer to employing the average style computed over $2.5k$ patches of the target dataset. Based on the empirical findings, it was determined that employing the mean value of the objective variable across each training step batch leads to a further performance improvement (e.g., to $41.01$ for the SYNTHIA-to-Cityscapes setting). %
The utilization of the per-batch style may explain why it effectively captures the significant variations in depth values based on the patch's position. This strategy was then adopted for all the subsequent tests. %
\newcommand\fs{0.15}

\afterpage{
\begin{figure*}[ht]
\begin{subfigure}[b]{\textwidth}
\centering
\begin{subfigure}[b]{\fs\textwidth}
\caption*{RGB}
\includegraphics[width=\textwidth]{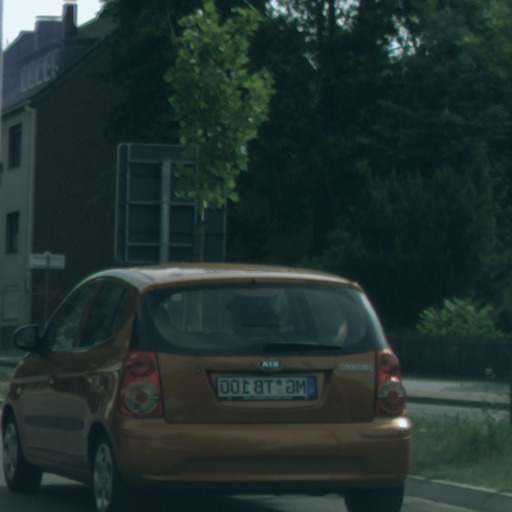}
\end{subfigure} 
\begin{subfigure}[b]{\fs\textwidth}
\caption*{Depth}
\includegraphics[width=\textwidth]{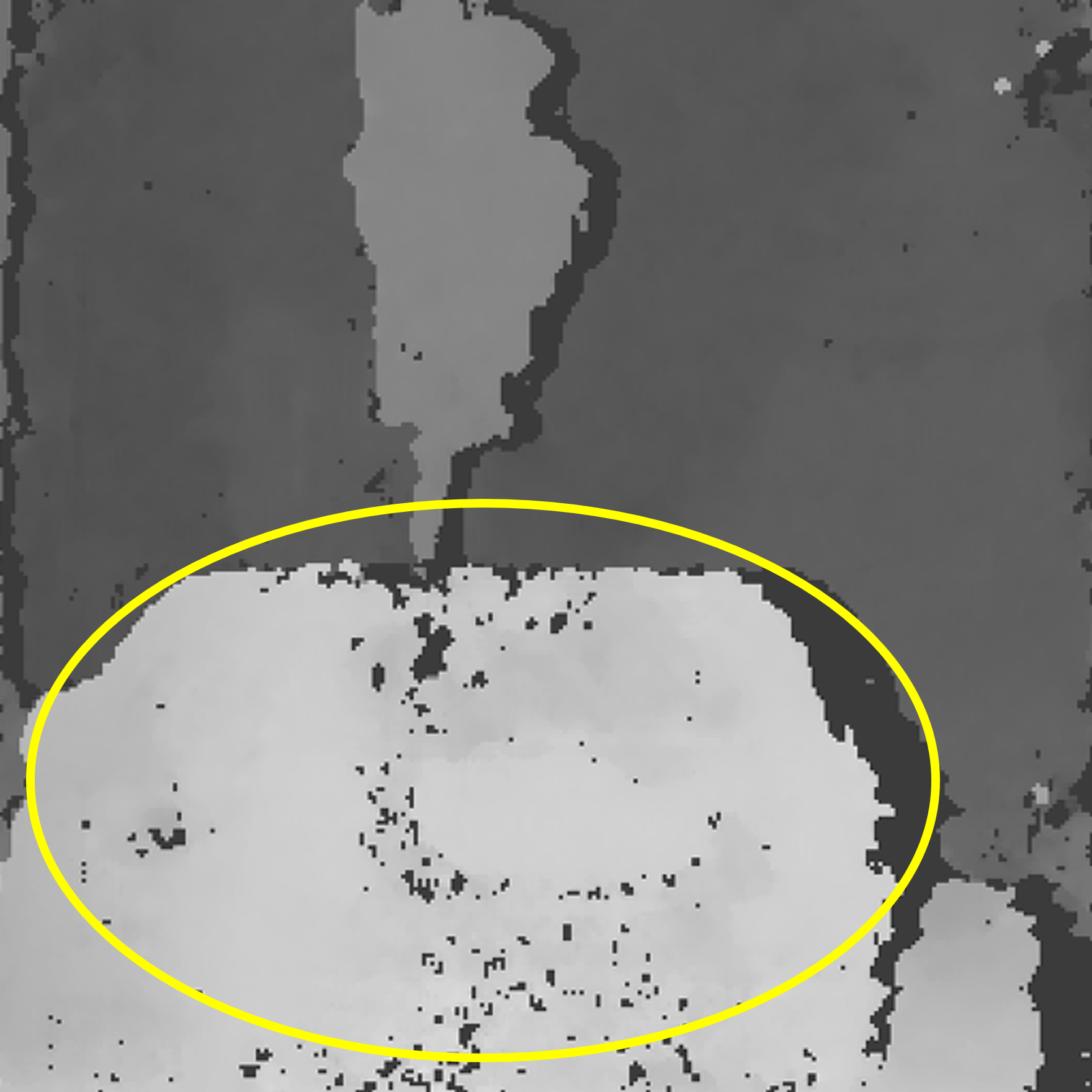}
\end{subfigure} 
\begin{subfigure}[b]{\fs\textwidth}
\caption*{Label}
\includegraphics[width=\textwidth]{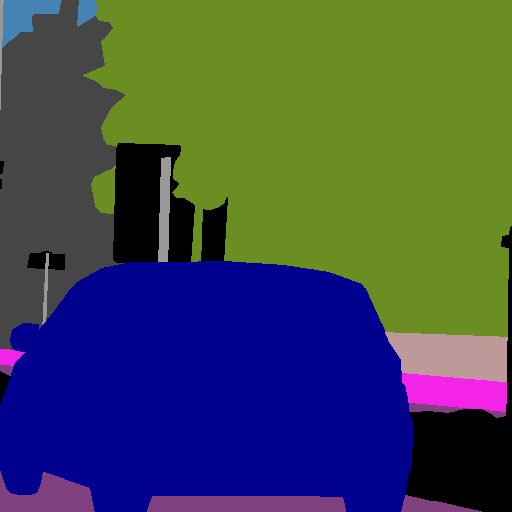}
\end{subfigure}
\begin{subfigure}[b]{\fs\textwidth}
\caption*{Pseudo-label}
\includegraphics[width=\textwidth]{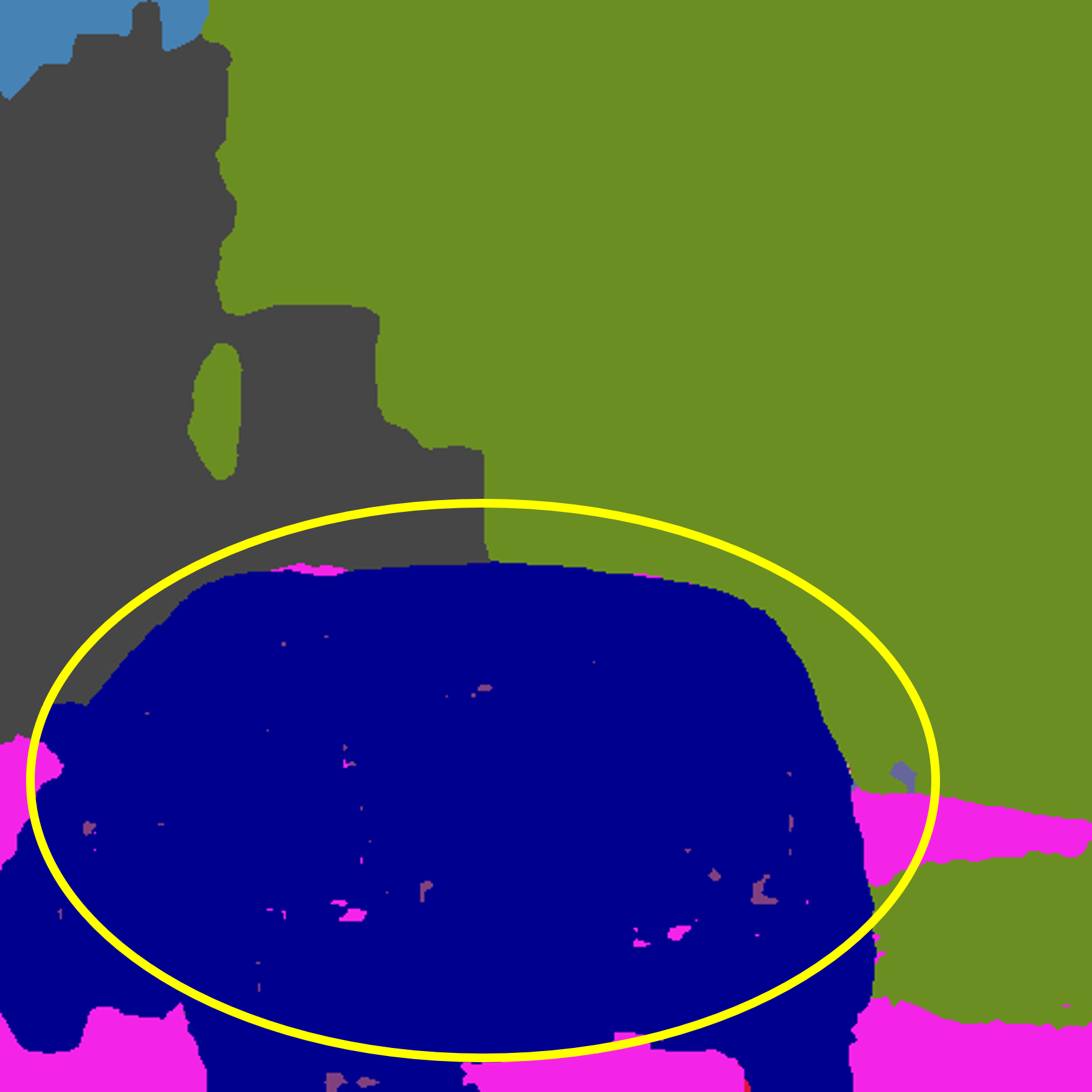}
\end{subfigure}
\begin{subfigure}[b]{\fs\textwidth}
\caption*{Prediction}
\includegraphics[width=\textwidth]{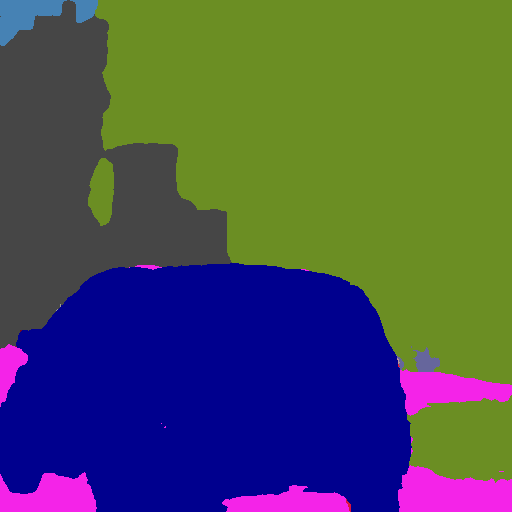}
\end{subfigure}
\end{subfigure}

\begin{subfigure}[b]{\textwidth}
\centering
\begin{subfigure}[b]{\fs\textwidth}
\includegraphics[width=\textwidth]{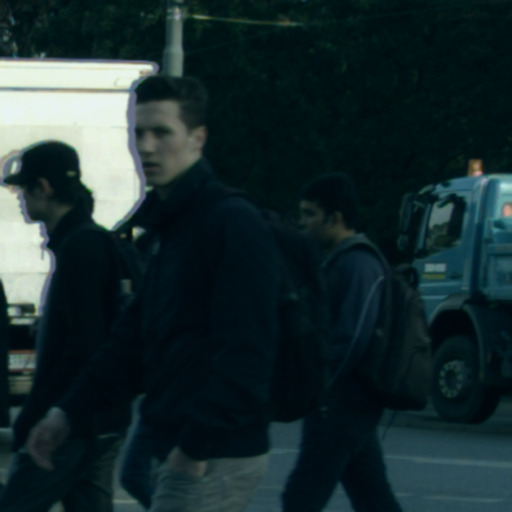}
\end{subfigure} 
\begin{subfigure}[b]{\fs\textwidth}
\includegraphics[width=\textwidth]{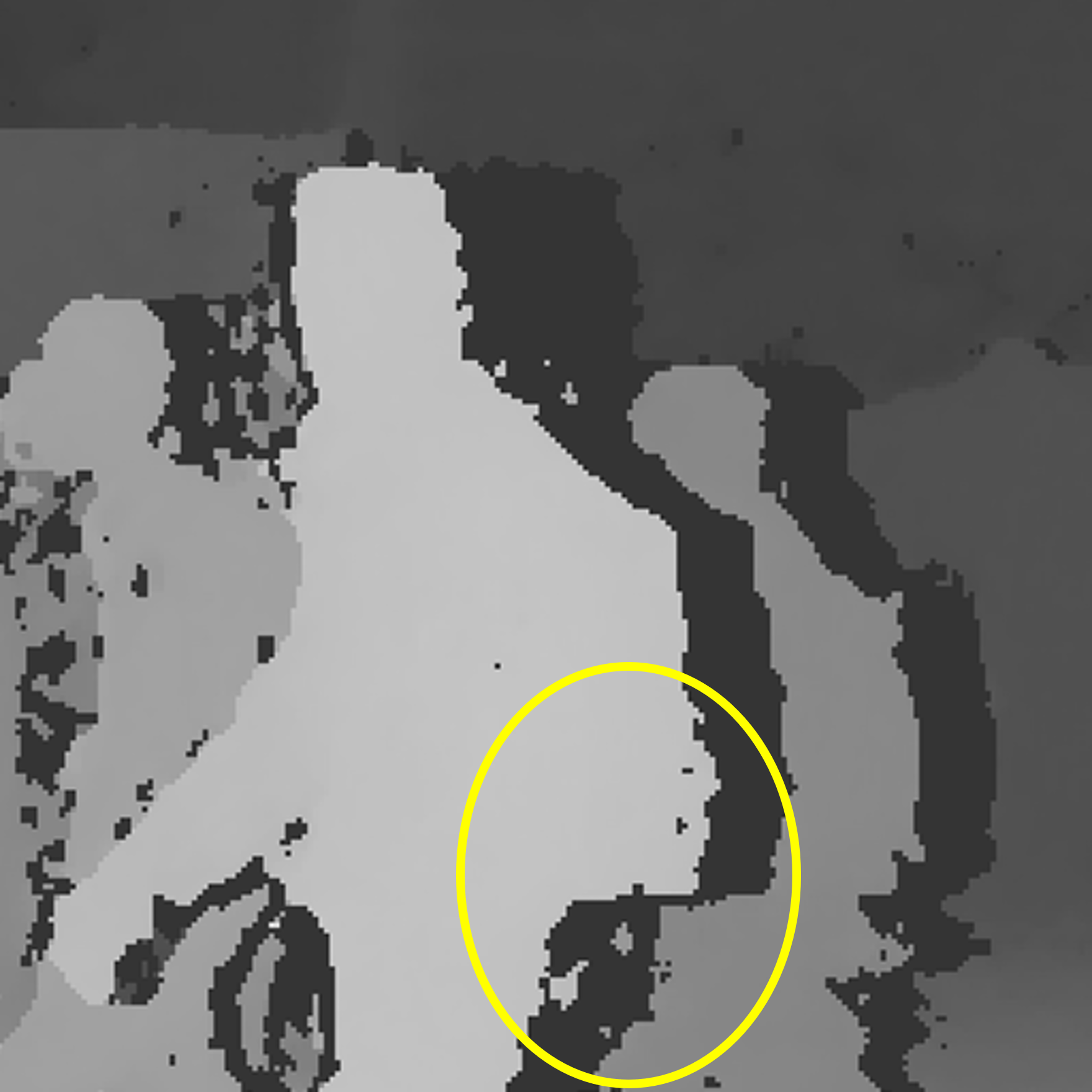}
\end{subfigure} 
\begin{subfigure}[b]{\fs\textwidth}
\includegraphics[width=\textwidth]{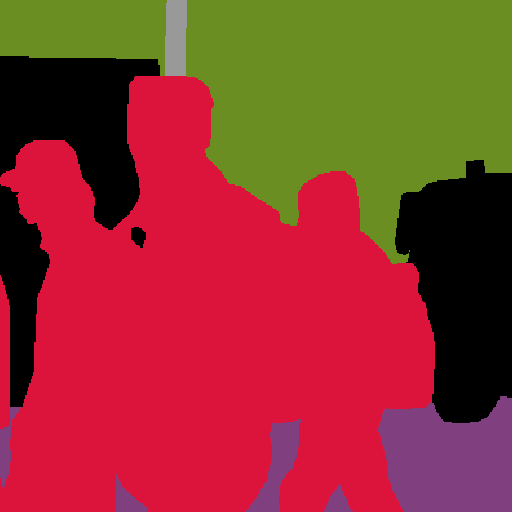}
\end{subfigure}
\begin{subfigure}[b]{\fs\textwidth}
\includegraphics[width=\textwidth]{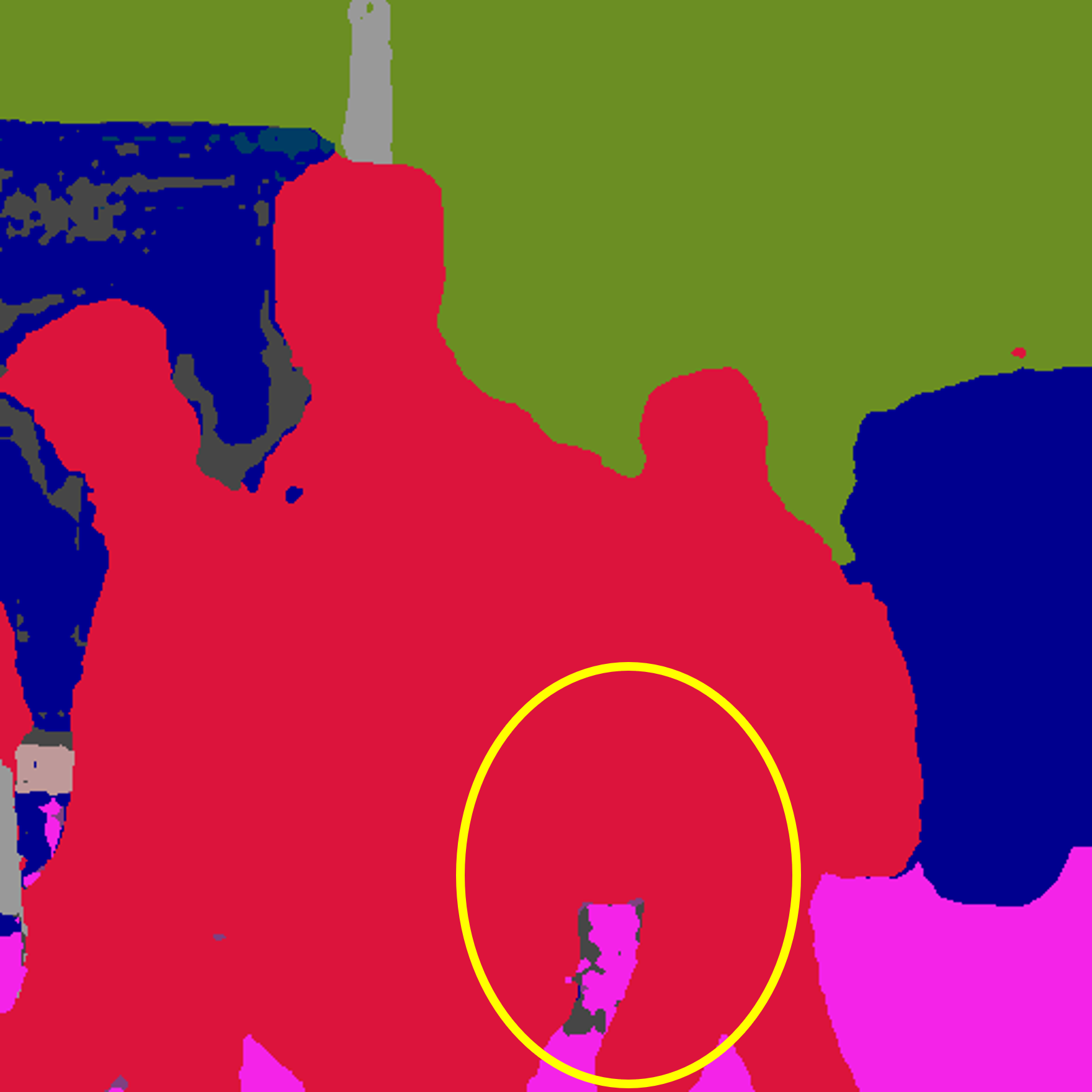}
\end{subfigure}
\begin{subfigure}[b]{\fs\textwidth}
\includegraphics[width=\textwidth]{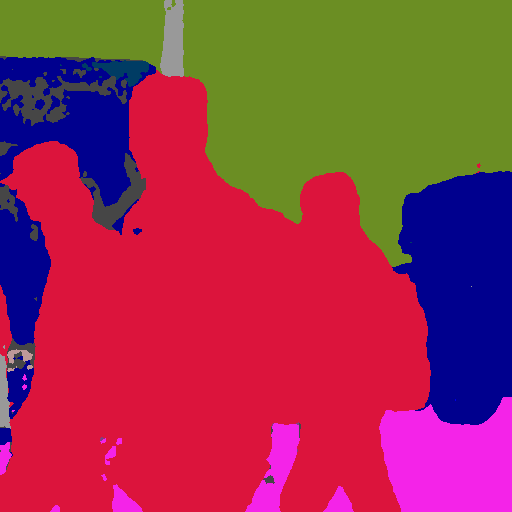}
\end{subfigure}
\end{subfigure}

\begin{subfigure}[b]{\textwidth}
\centering
\begin{subfigure}[b]{\fs\textwidth}
\includegraphics[width=\textwidth]{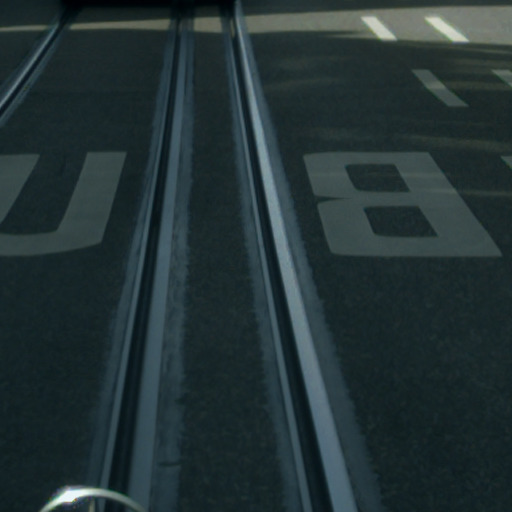}
\end{subfigure} 
\begin{subfigure}[b]{\fs\textwidth}
\includegraphics[width=\textwidth]{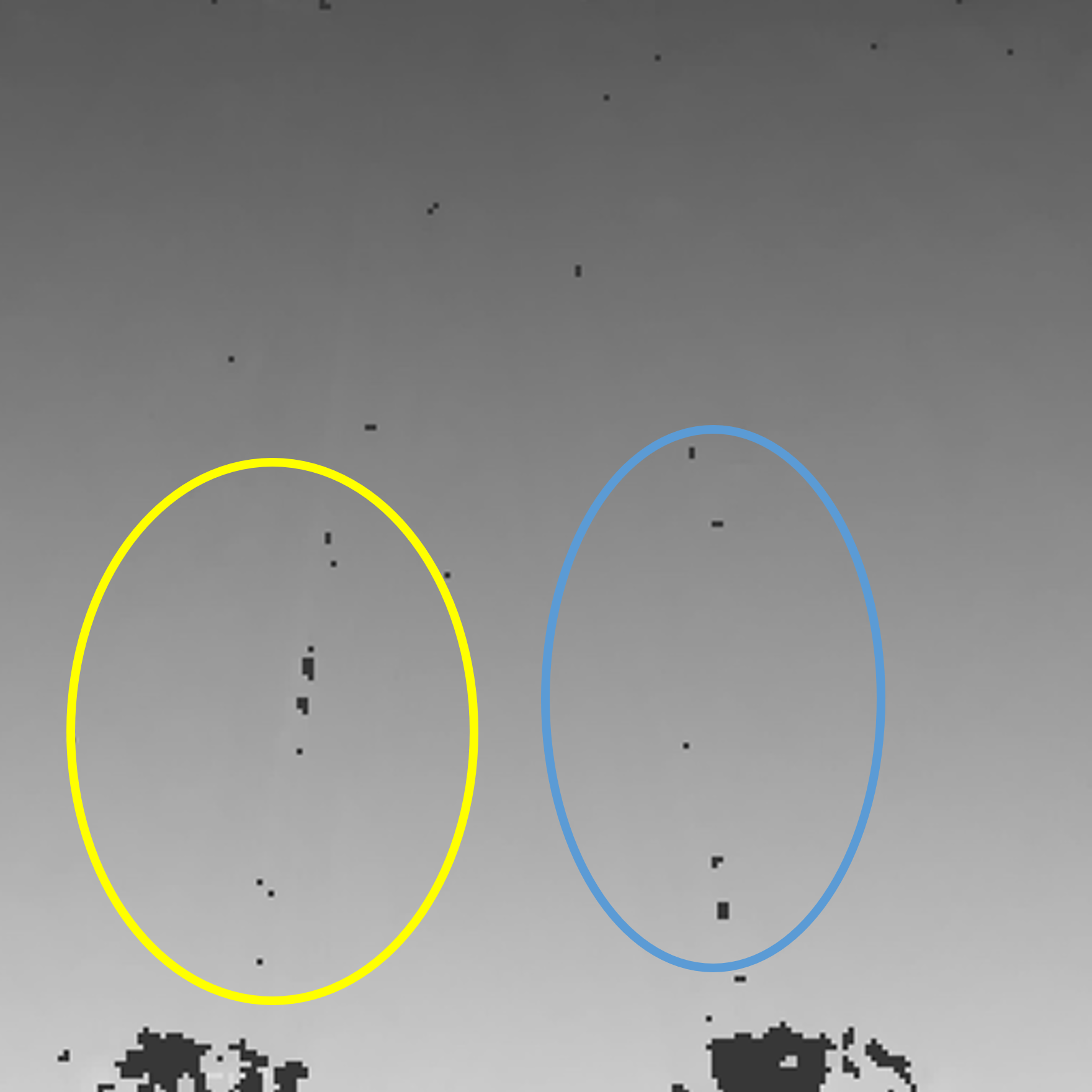}
\end{subfigure} 
\begin{subfigure}[b]{\fs\textwidth}
\includegraphics[width=\textwidth]{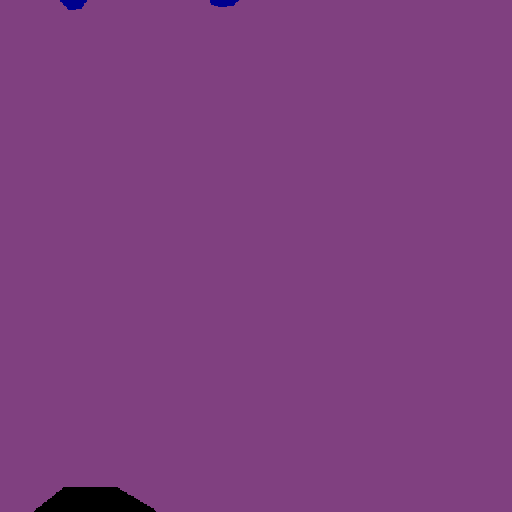}
\end{subfigure}
\begin{subfigure}[b]{\fs\textwidth}
\includegraphics[width=\textwidth]{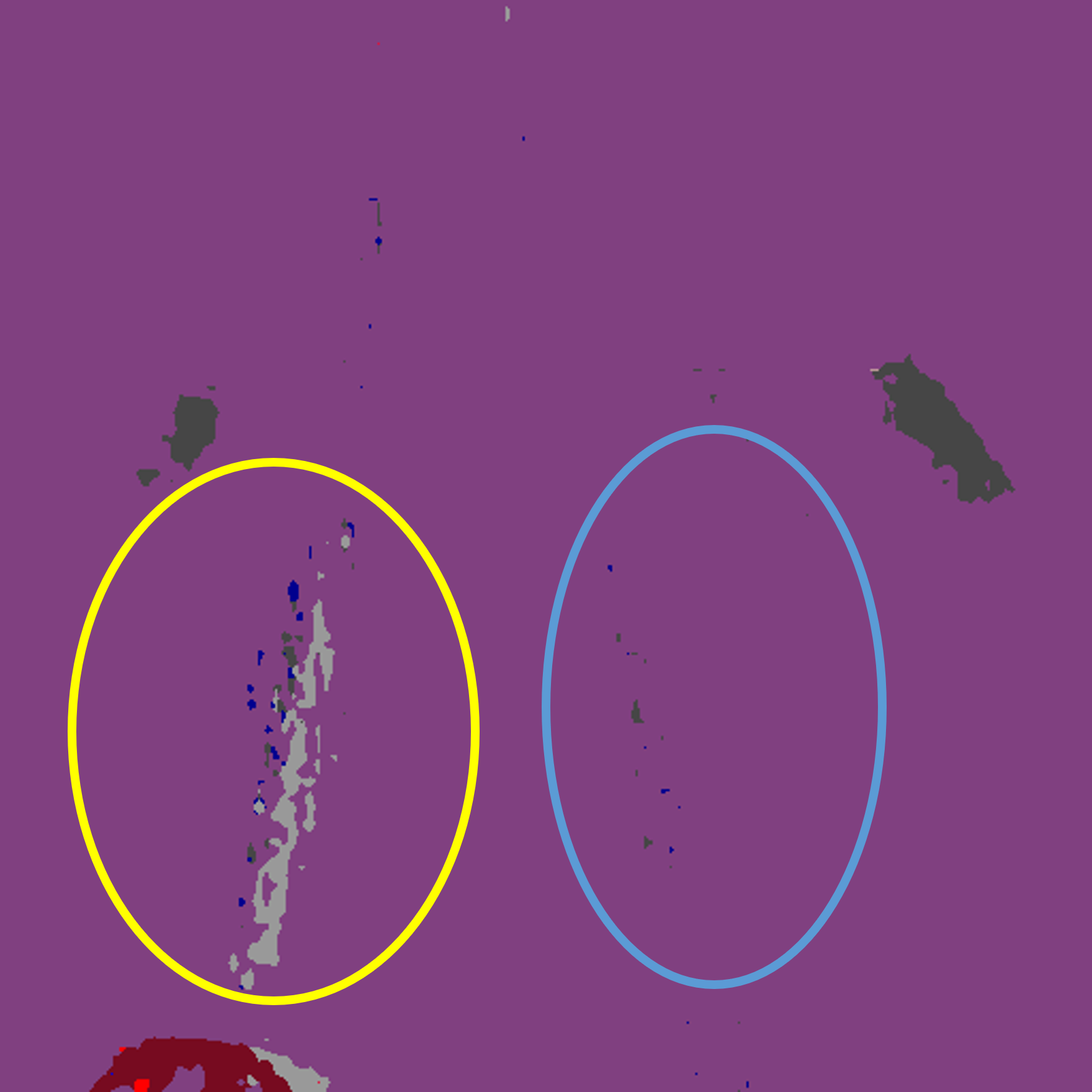}
\end{subfigure}
\begin{subfigure}[b]{\fs\textwidth}
\includegraphics[width=\textwidth]{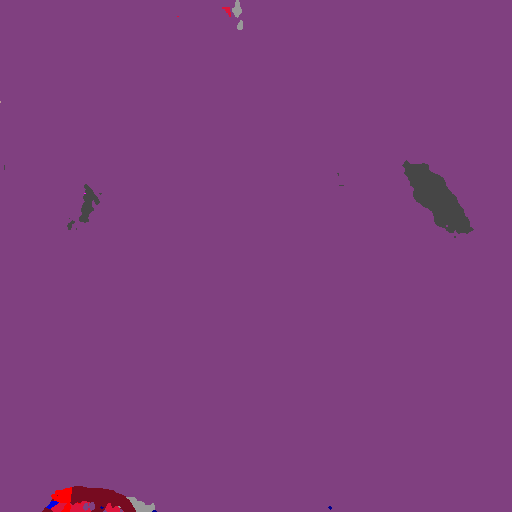}
\end{subfigure}
\end{subfigure}
\caption{Self-training with depth masking allows masking areas where the depth fails to capture the scene content.}
\label{fig:depth_masking}
\end{figure*}
}

\afterpage{
\begin{figure*}[ht]
\begin{subfigure}[b]{\textwidth}
\centering
\begin{subfigure}[b]{\fs\textwidth}
\caption*{RGB}
\includegraphics[width=\textwidth]{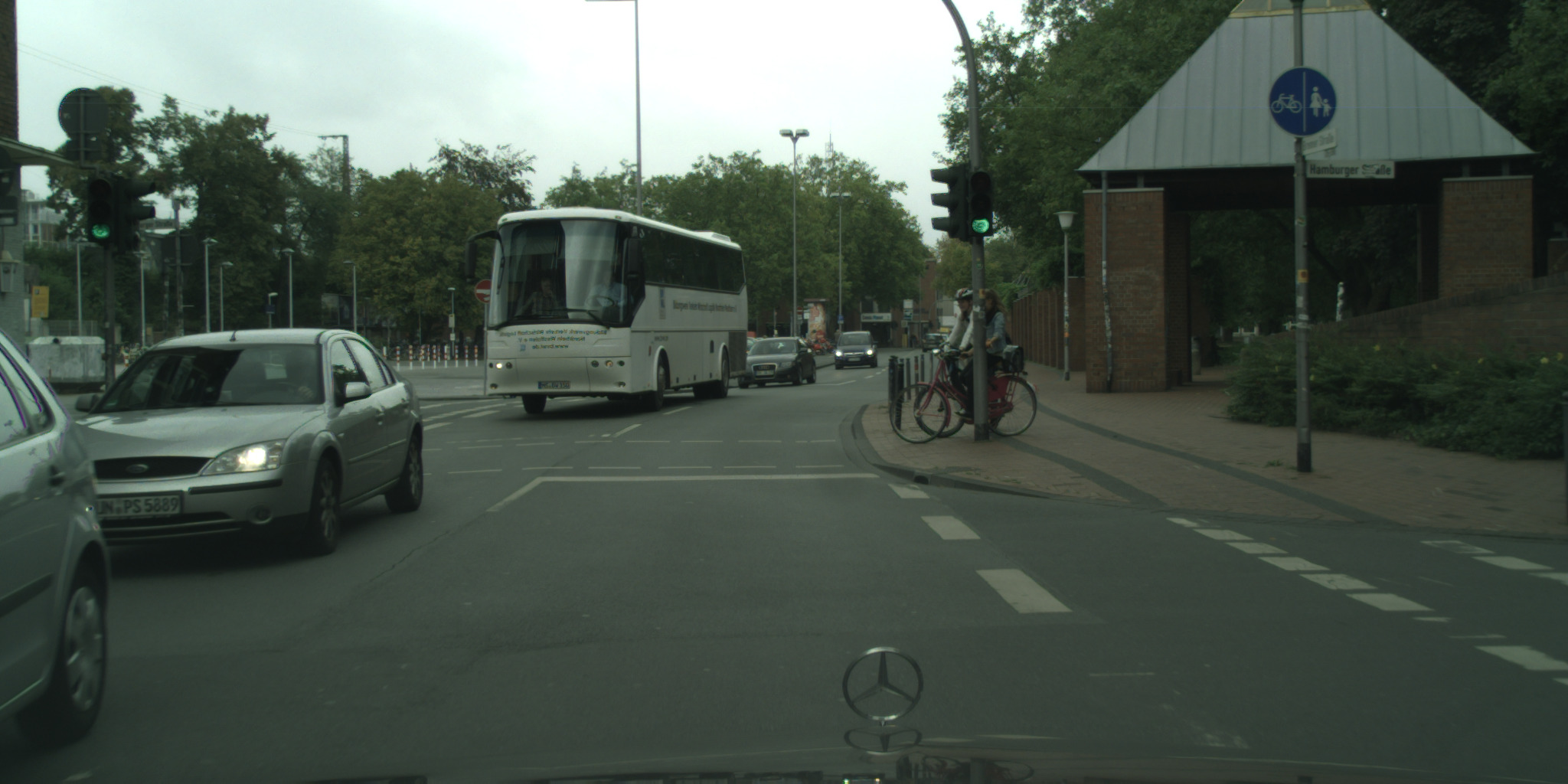}
\end{subfigure} 
\begin{subfigure}[b]{\fs\textwidth}
\caption*{Depth}
\includegraphics[width=\textwidth]{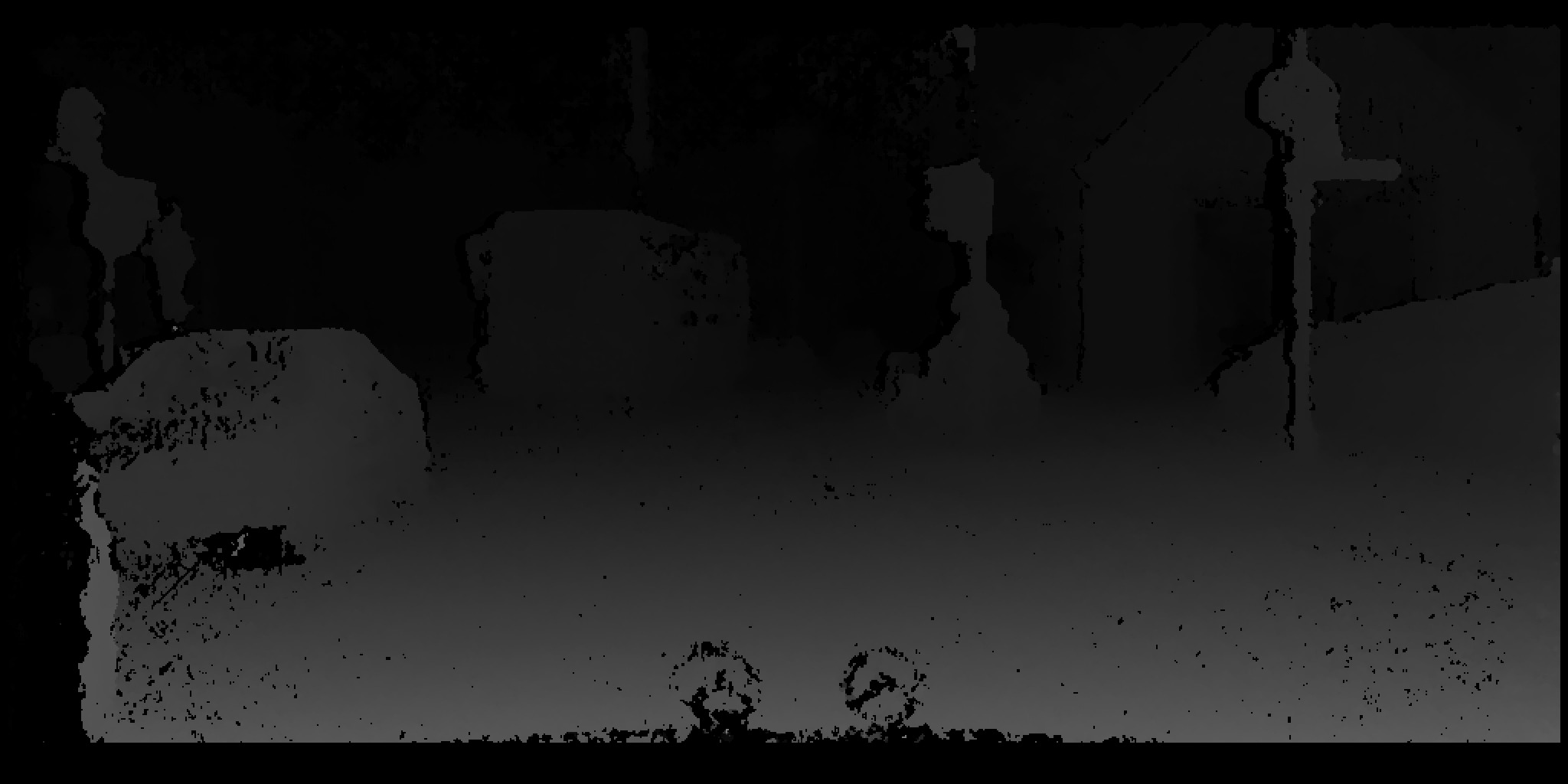}
\end{subfigure} 
\begin{subfigure}[b]{\fs\textwidth}
\caption*{GT}
\includegraphics[width=\textwidth]{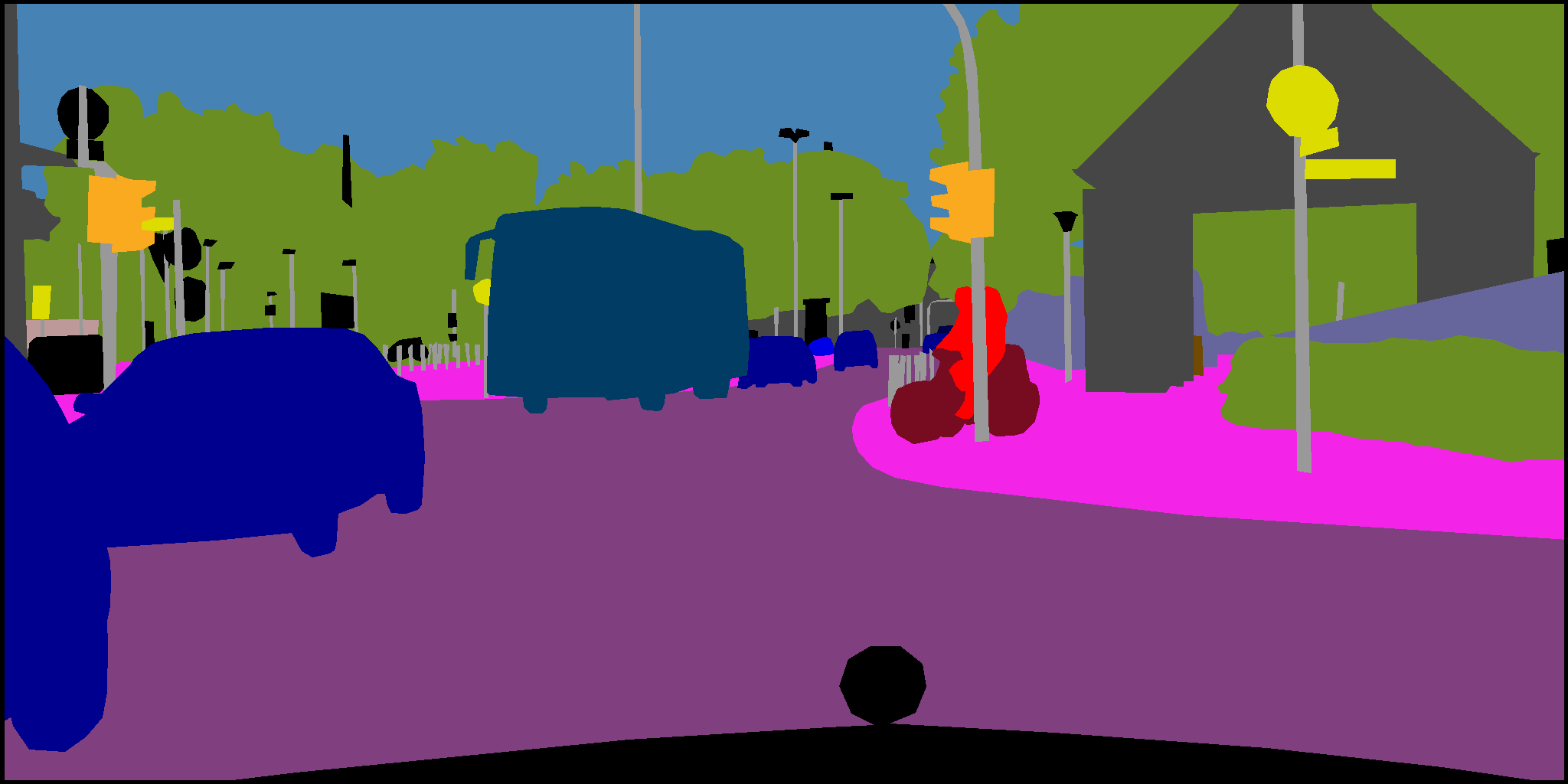}
\end{subfigure}
\begin{subfigure}[b]{\fs\textwidth}
\caption*{RGB+FDA\cite{yang2020fda}}
\includegraphics[width=\textwidth]{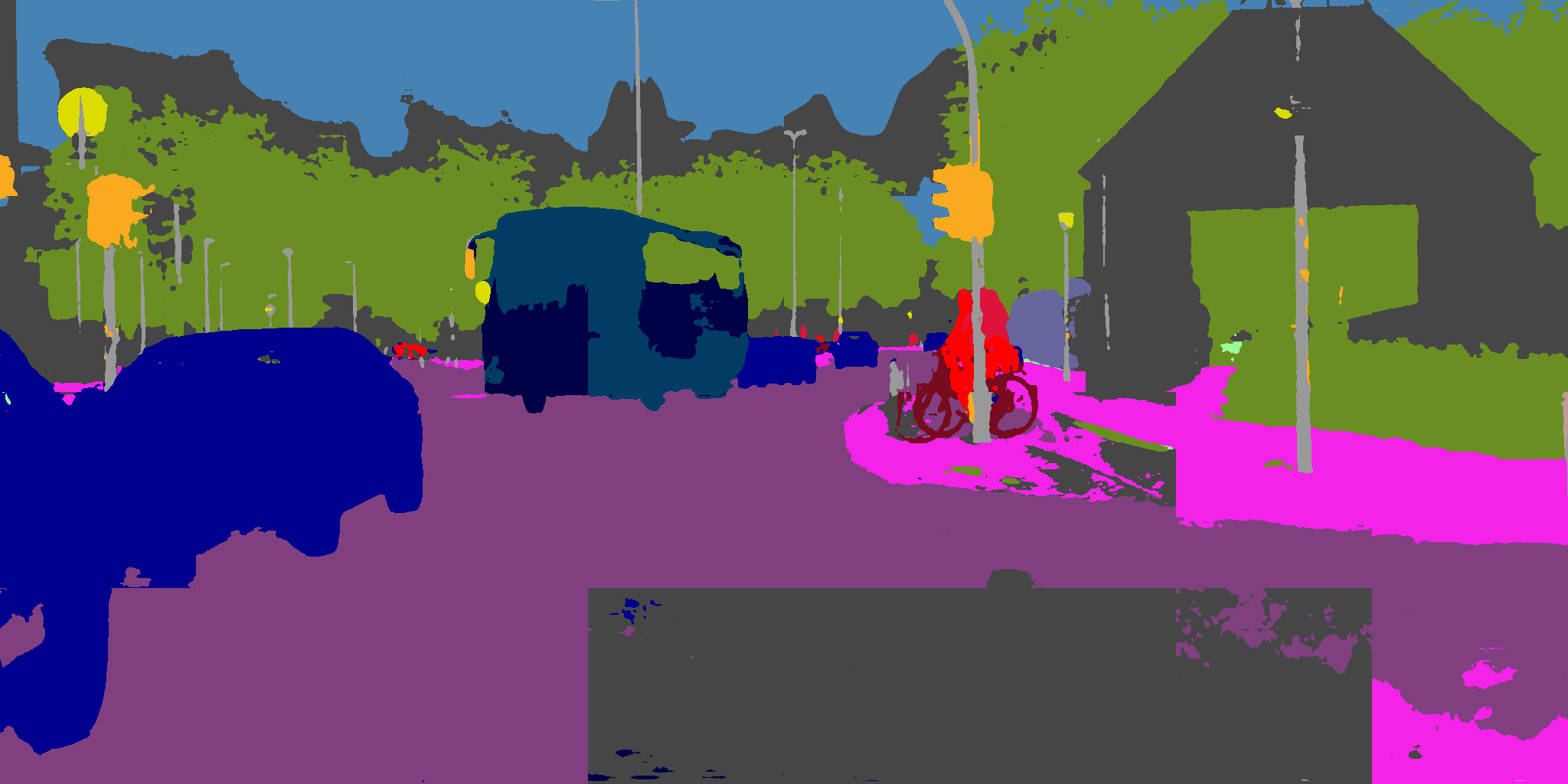}
\end{subfigure}
\begin{subfigure}[b]{\fs\textwidth}
\caption*{MISFIT\tiny{w/o$\mathcal{L}_{depth-ent}$}}
\includegraphics[width=\textwidth]{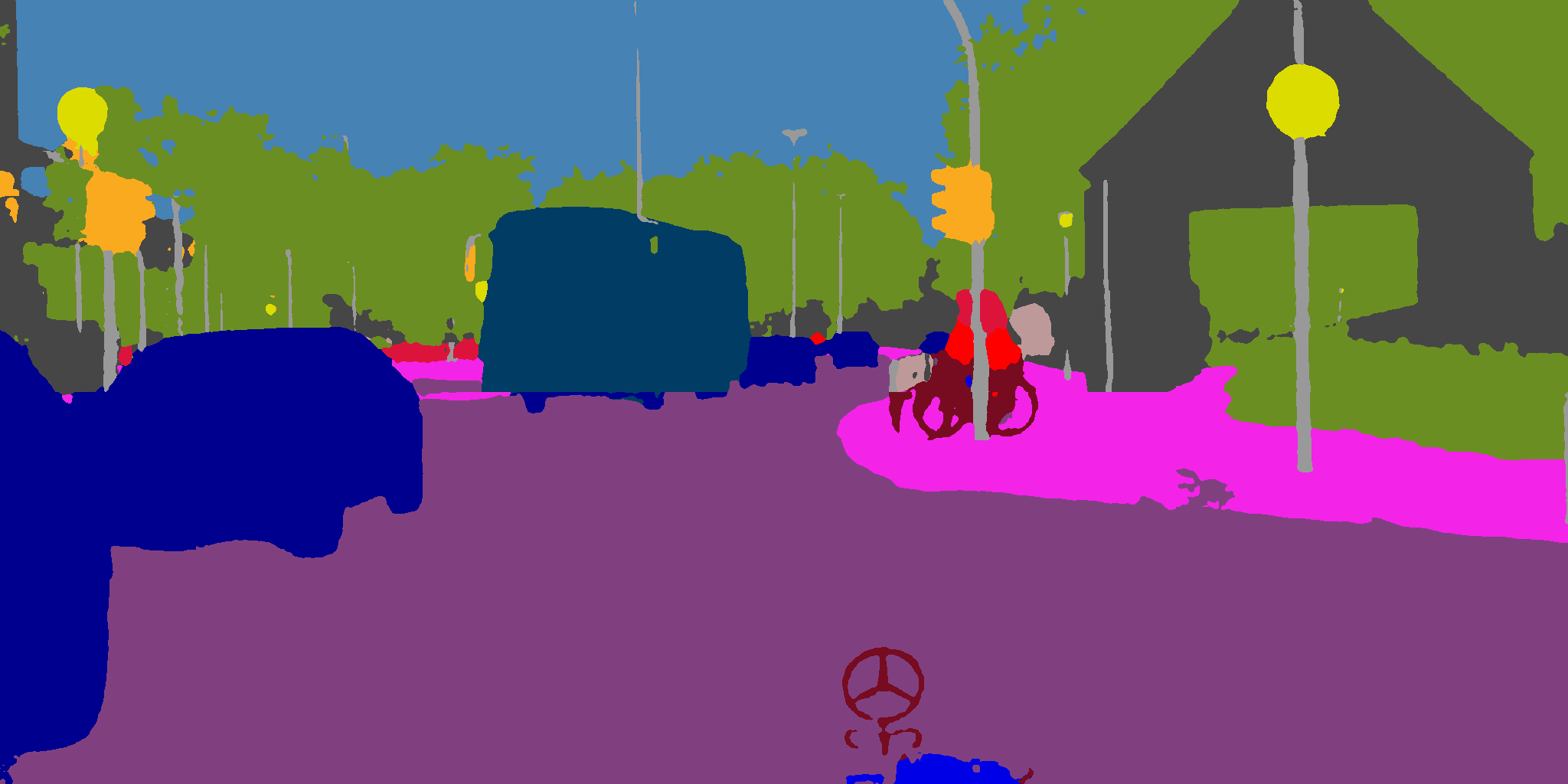}
\end{subfigure}
\begin{subfigure}[b]{\fs\textwidth}
\caption*{MISFIT}
\includegraphics[width=\textwidth]{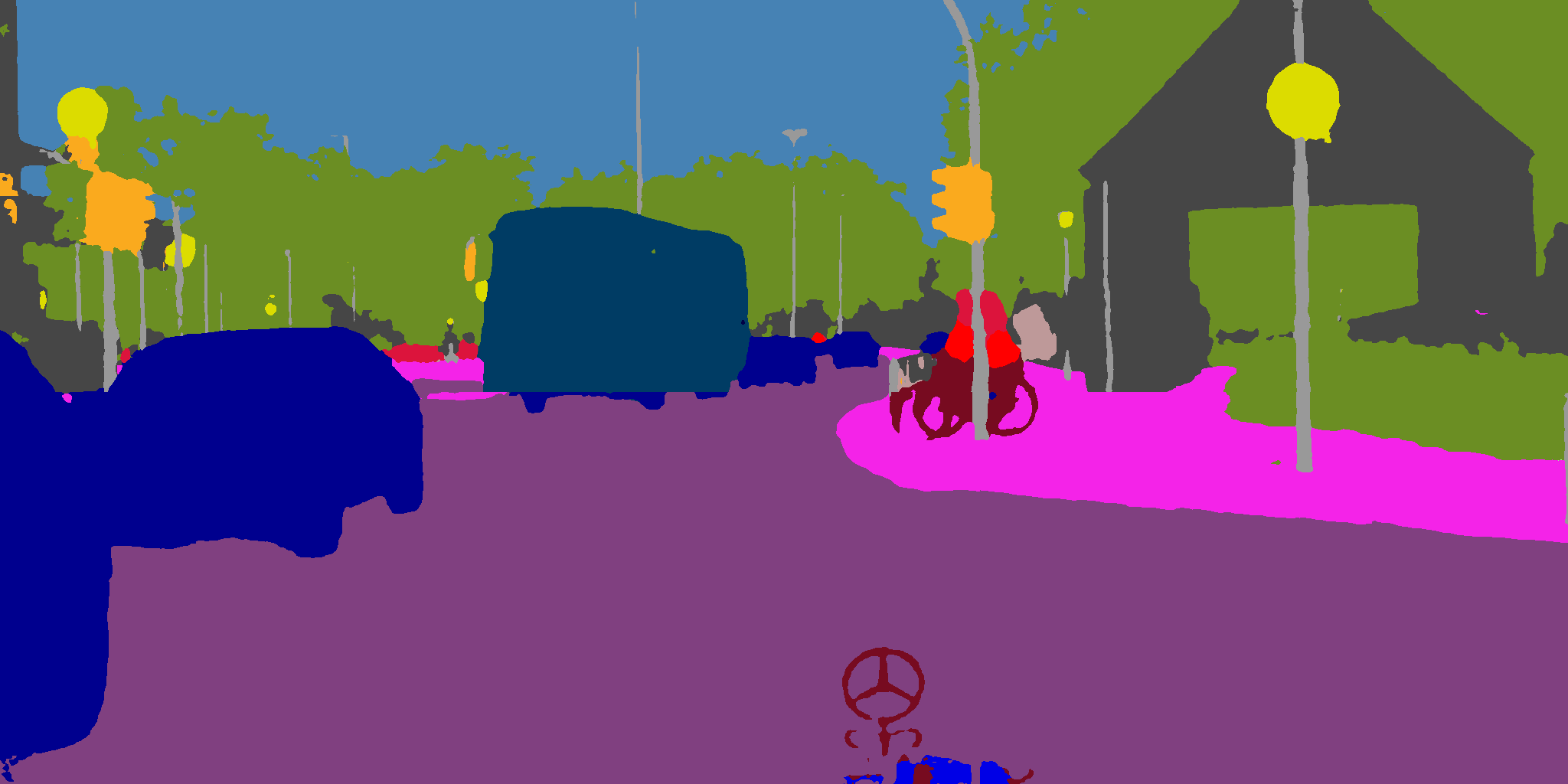}
\end{subfigure}
\end{subfigure}
\begin{subfigure}[b]{\textwidth}
\centering
\begin{subfigure}[b]{\fs\textwidth}
\includegraphics[width=\textwidth]{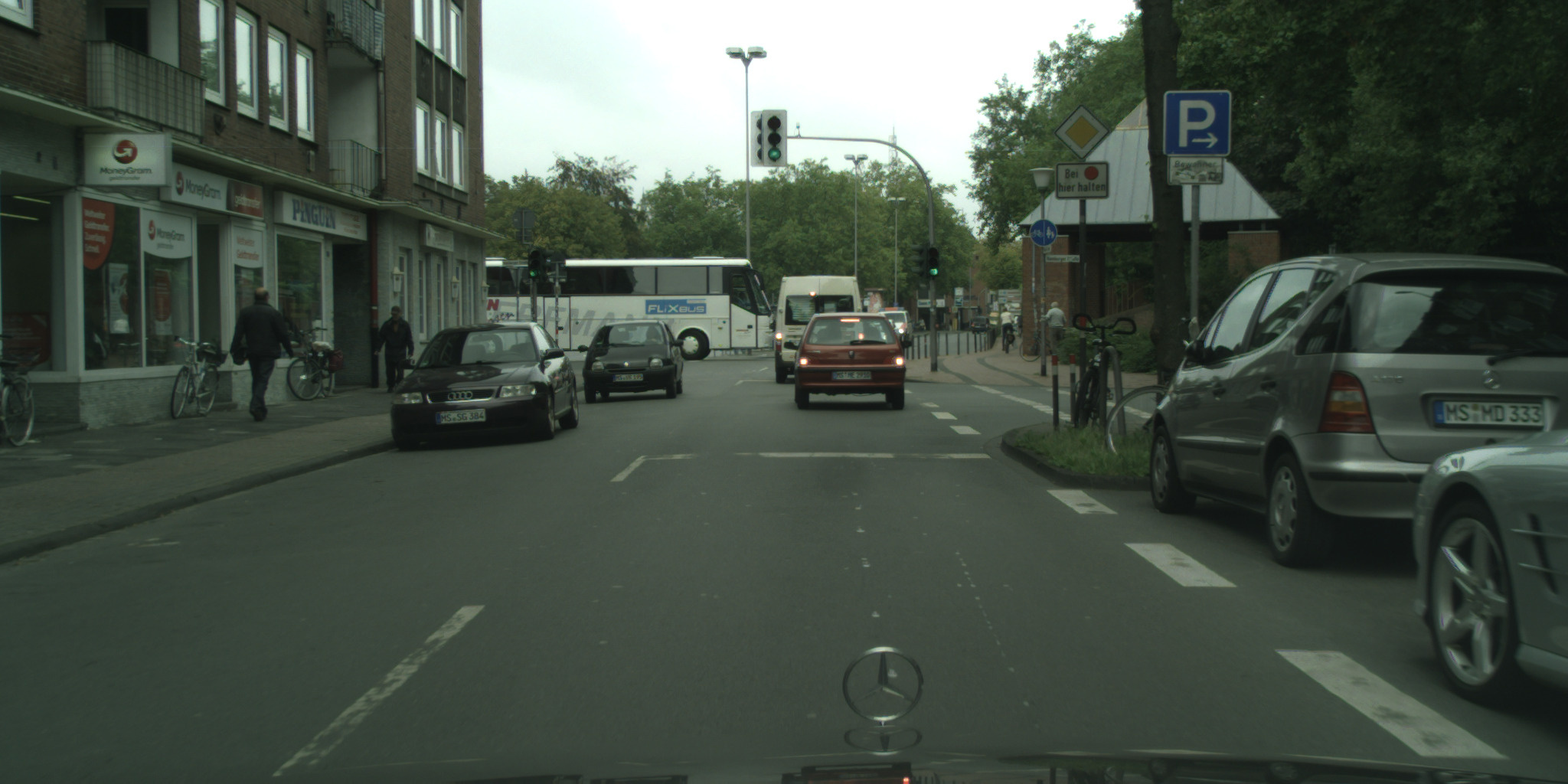}
\end{subfigure} 
\begin{subfigure}[b]{\fs\textwidth}
\includegraphics[width=\textwidth]{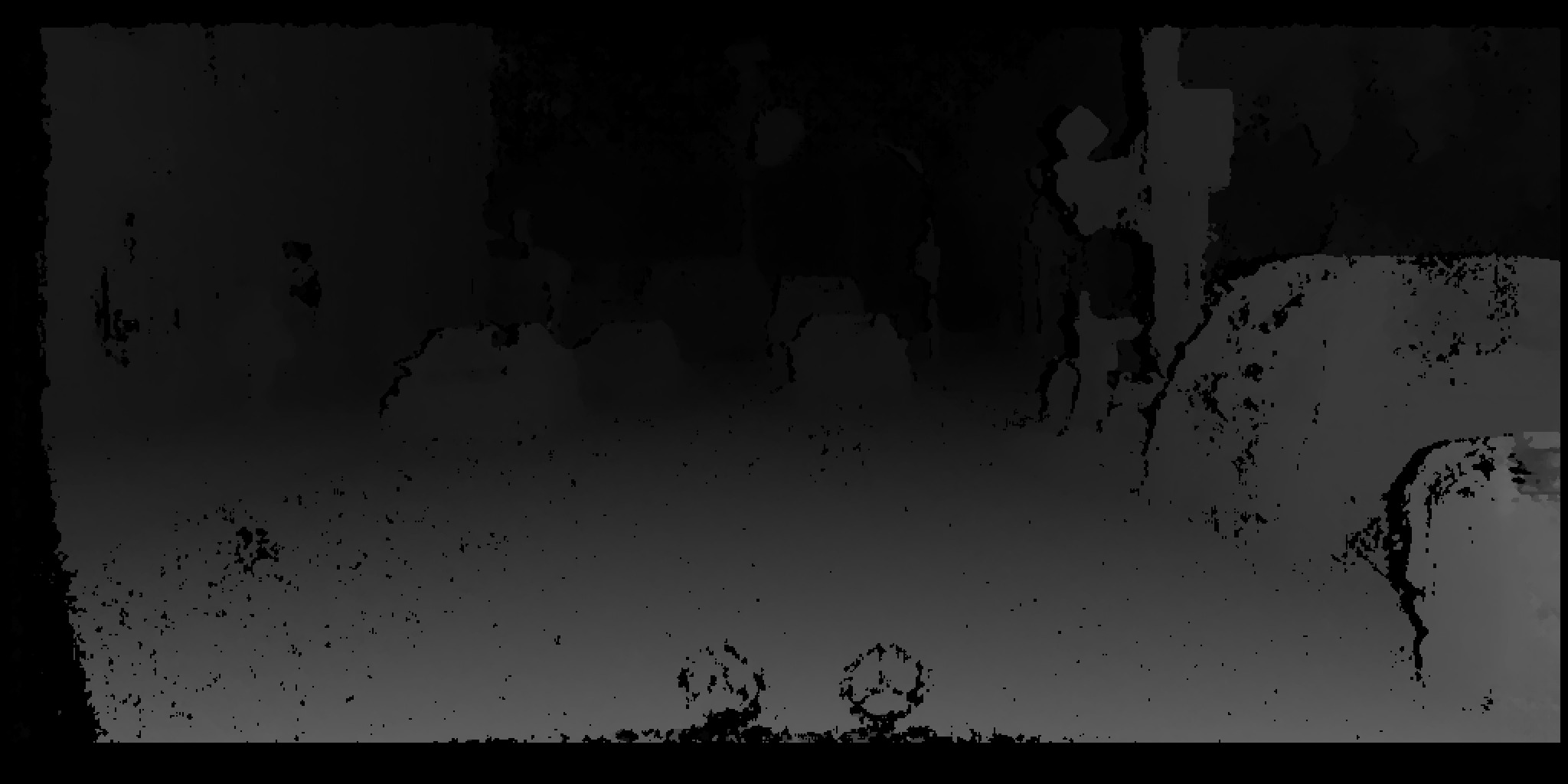}
\end{subfigure} 
\begin{subfigure}[b]{\fs\textwidth}
\includegraphics[width=\textwidth]{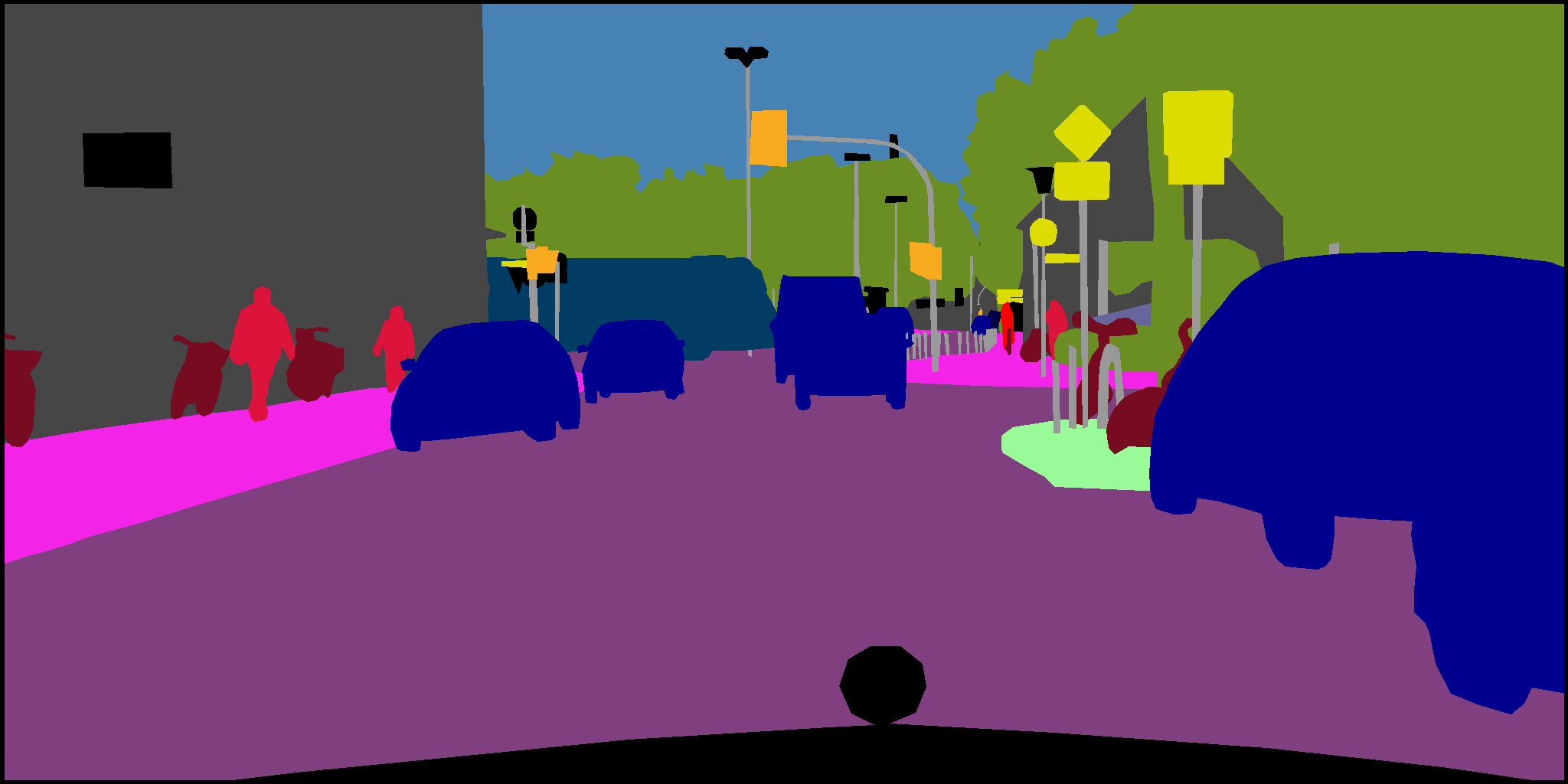}
\end{subfigure}
\begin{subfigure}[b]{\fs\textwidth}
\includegraphics[width=\textwidth]{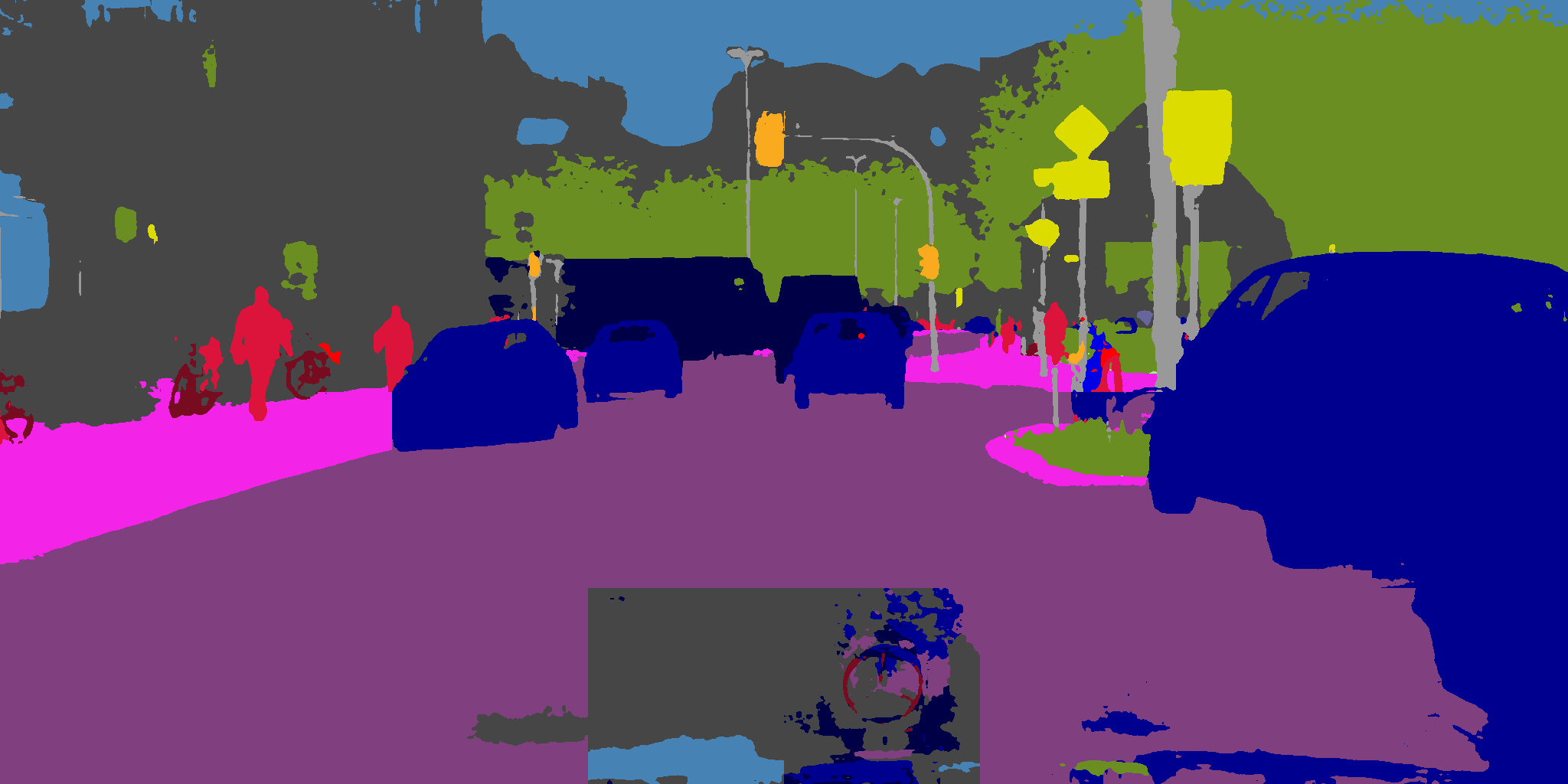}
\end{subfigure}
\begin{subfigure}[b]{\fs\textwidth}
\includegraphics[width=\textwidth]{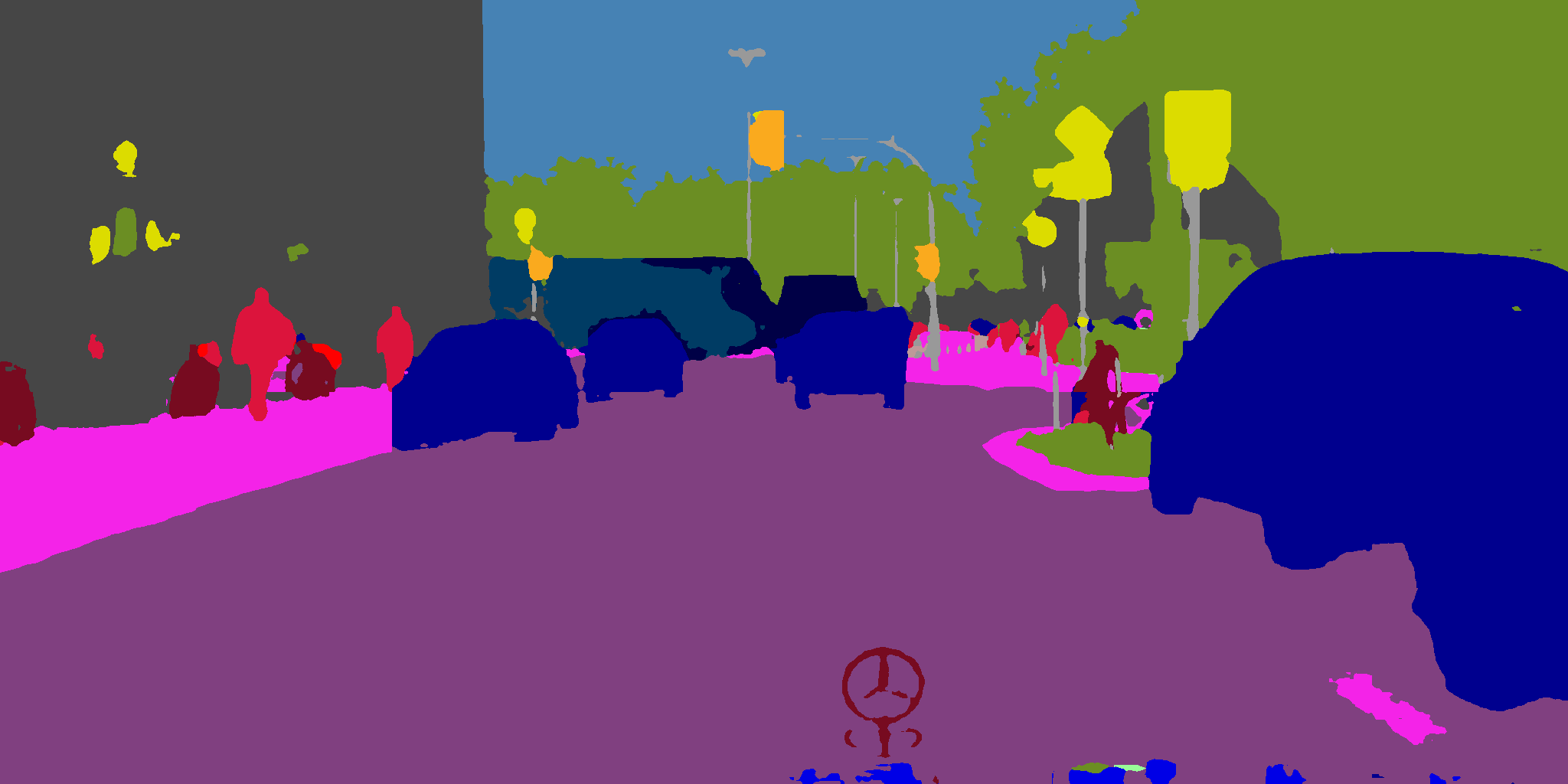}
\end{subfigure}
\begin{subfigure}[b]{\fs\textwidth}
\includegraphics[width=\textwidth]{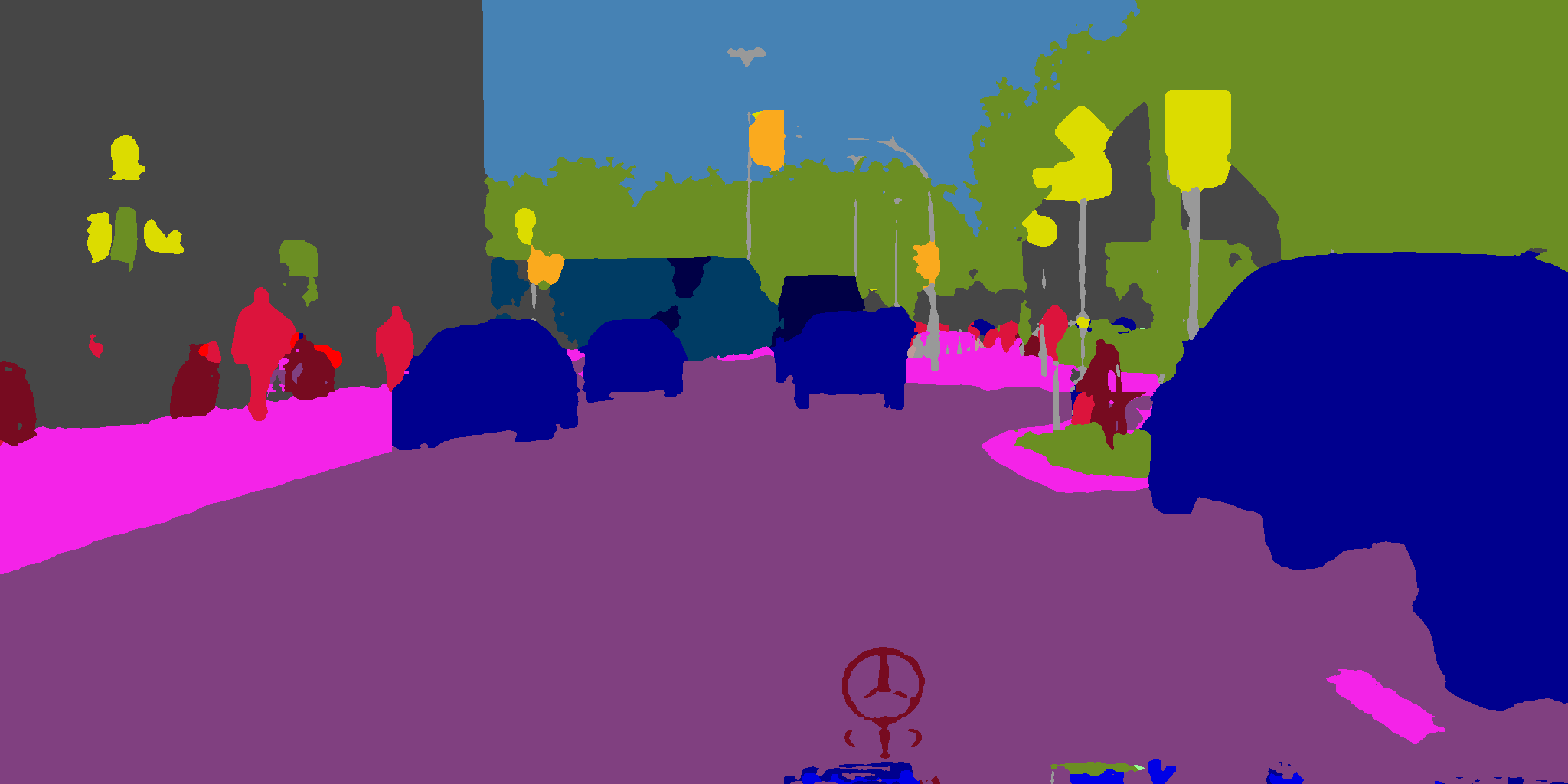}
\end{subfigure}
\end{subfigure}
\begin{subfigure}[b]{\textwidth}
\centering
\begin{subfigure}[b]{\fs\textwidth}
\includegraphics[width=\textwidth]{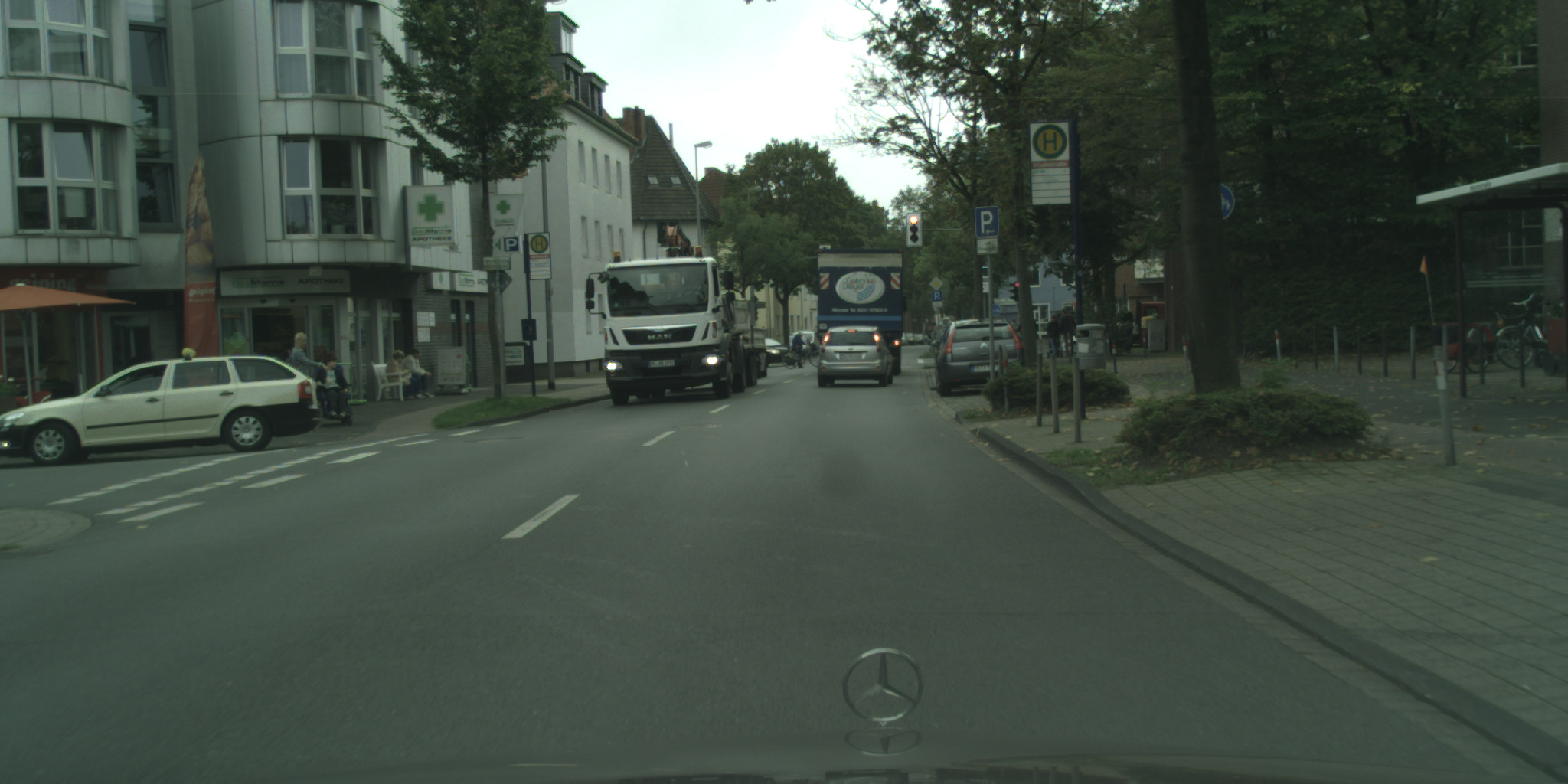}
\end{subfigure} 
\begin{subfigure}[b]{\fs\textwidth}
\includegraphics[width=\textwidth]{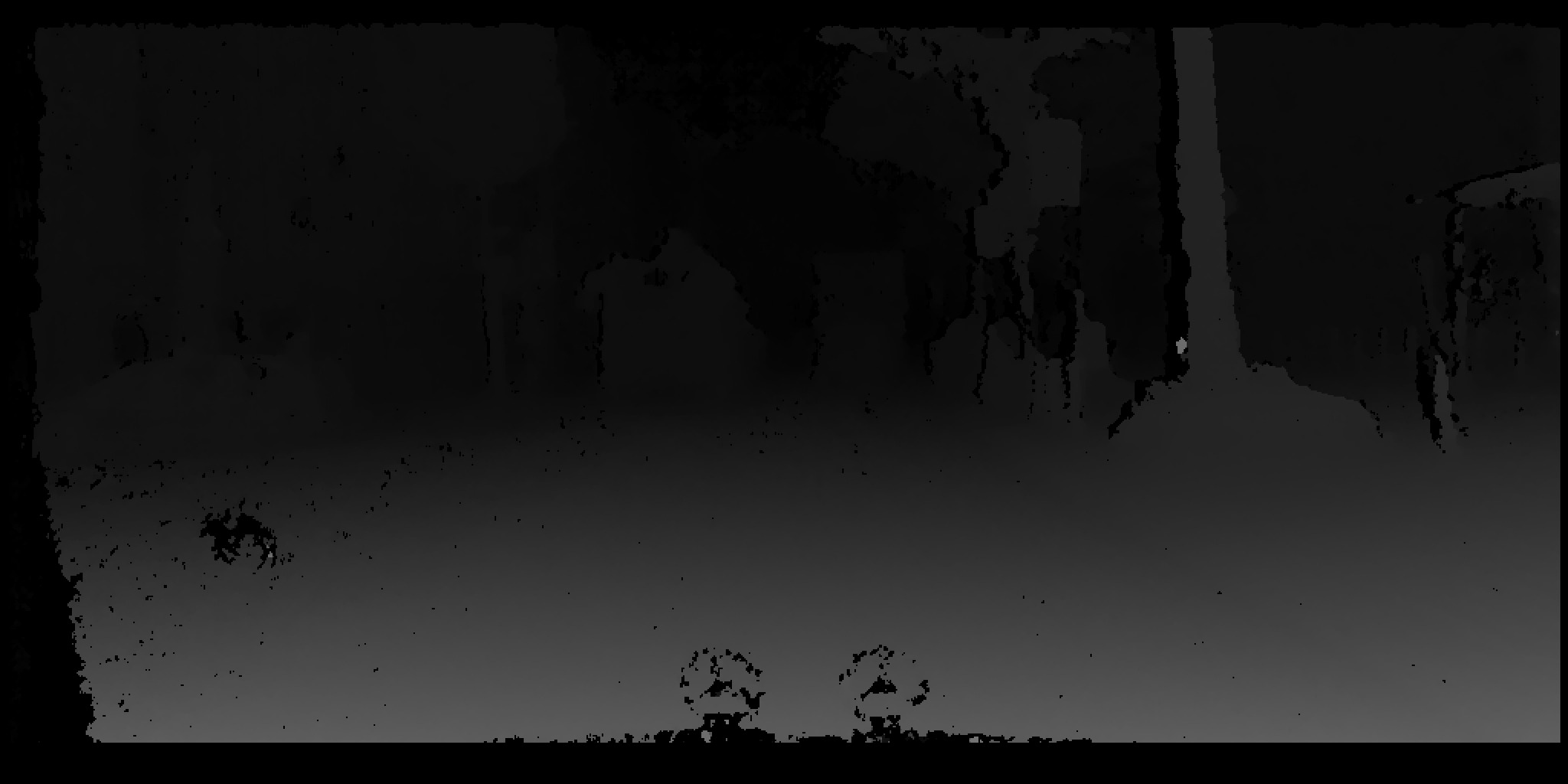}
\end{subfigure} 
\begin{subfigure}[b]{\fs\textwidth}
\includegraphics[width=\textwidth]{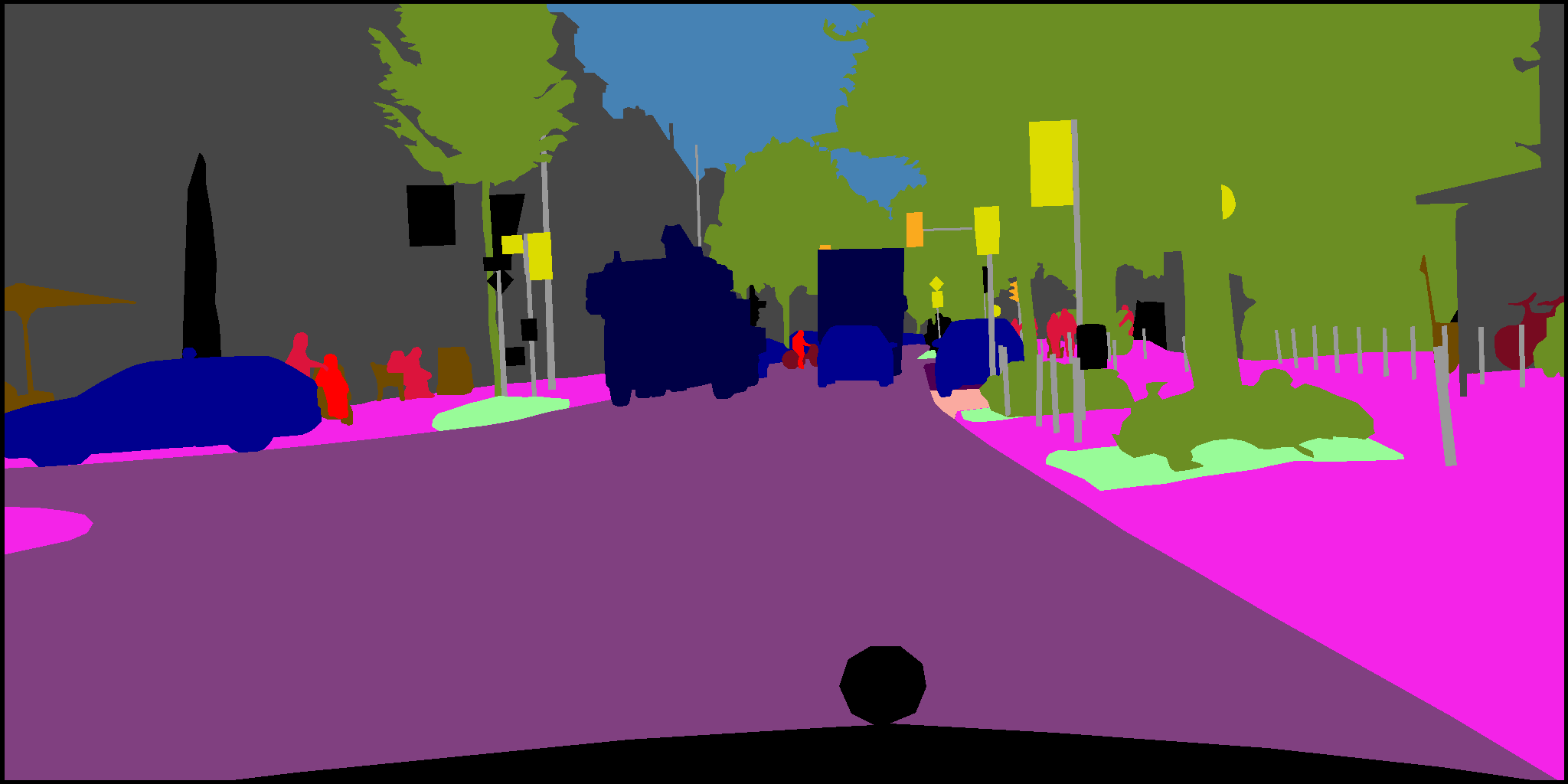}
\end{subfigure}
\begin{subfigure}[b]{\fs\textwidth}
\includegraphics[width=\textwidth]{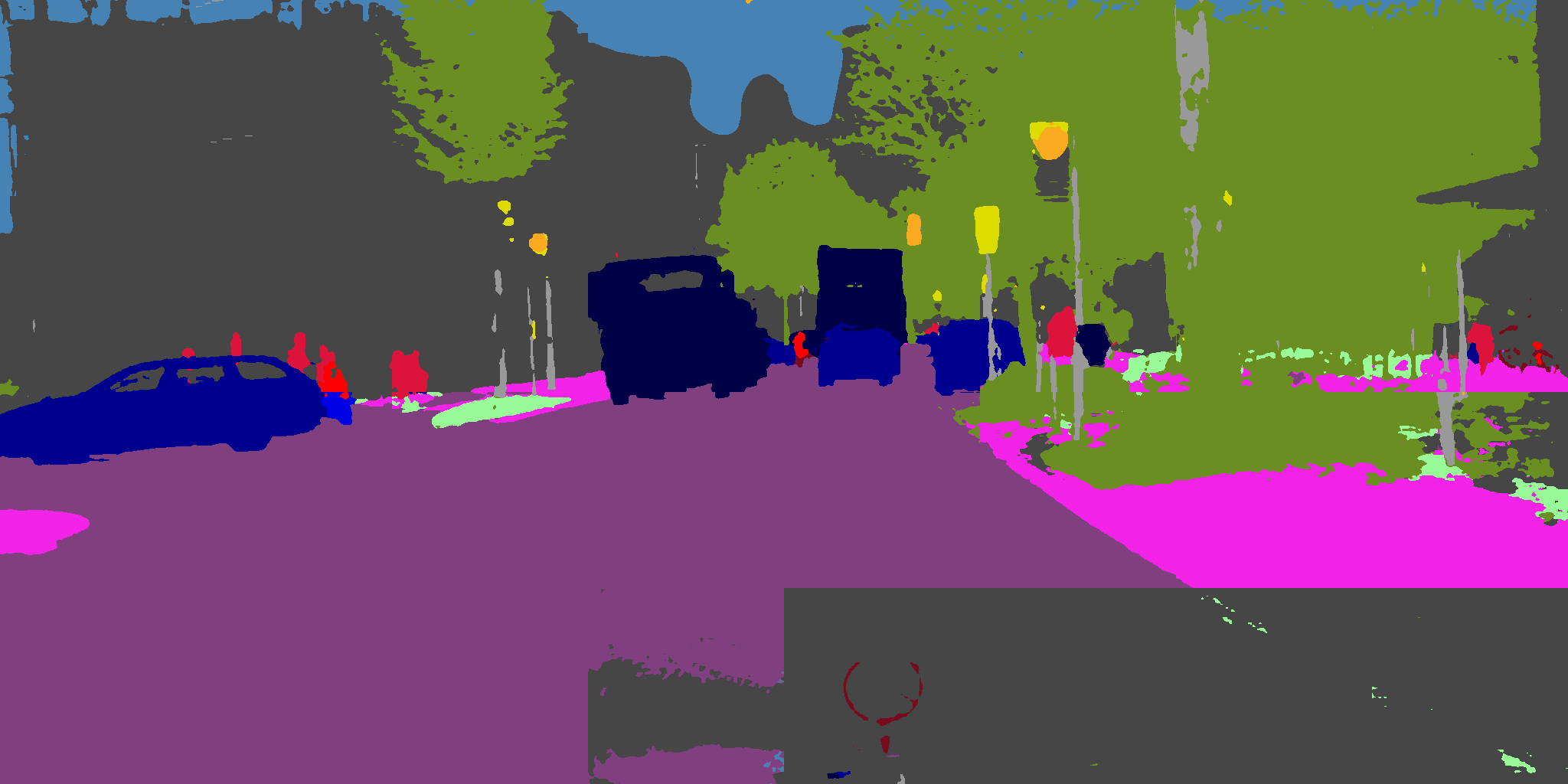}
\end{subfigure}
\begin{subfigure}[b]{\fs\textwidth}
\includegraphics[width=\textwidth]{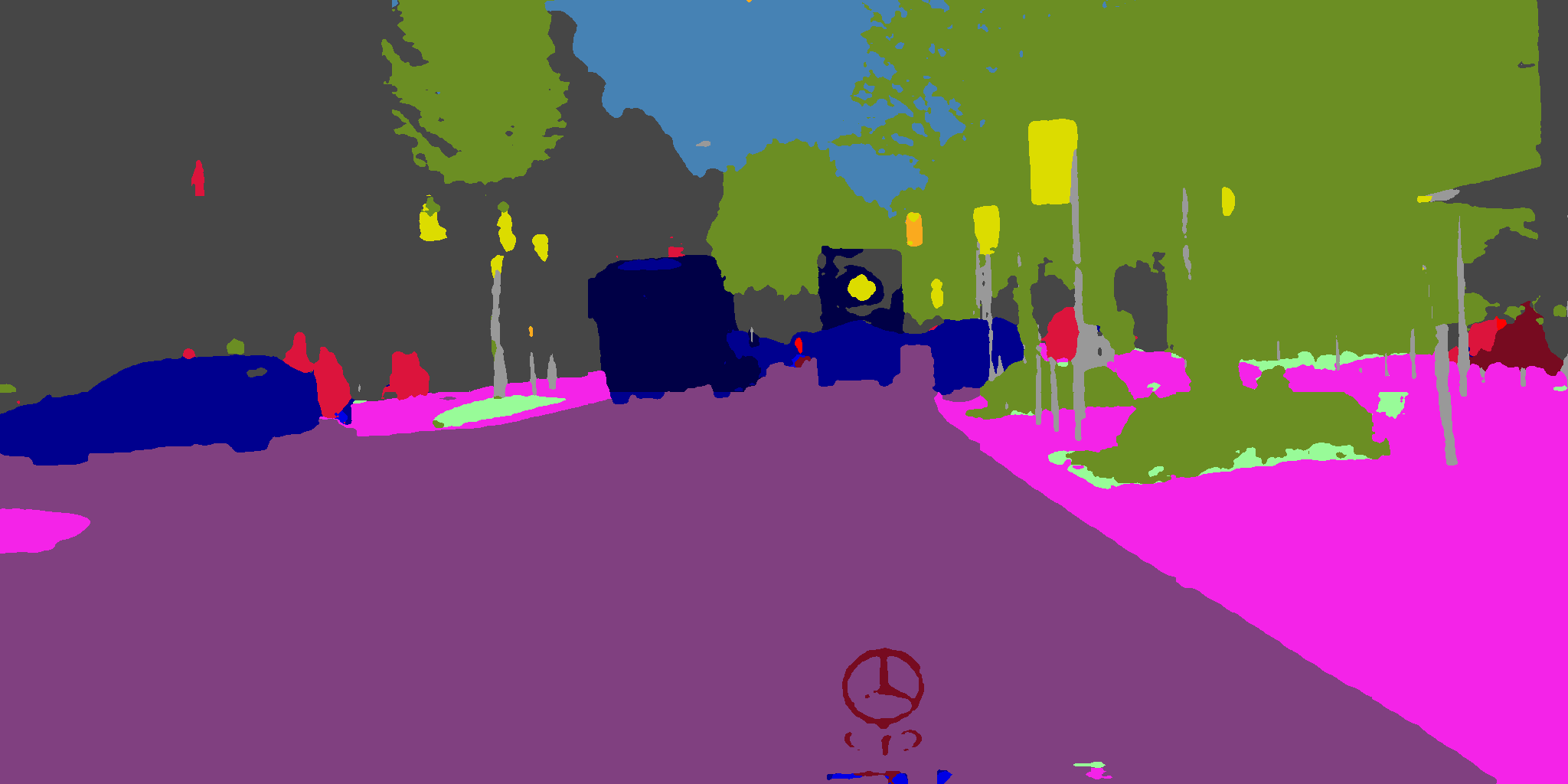}
\end{subfigure}
\begin{subfigure}[b]{\fs\textwidth}
\includegraphics[width=\textwidth]{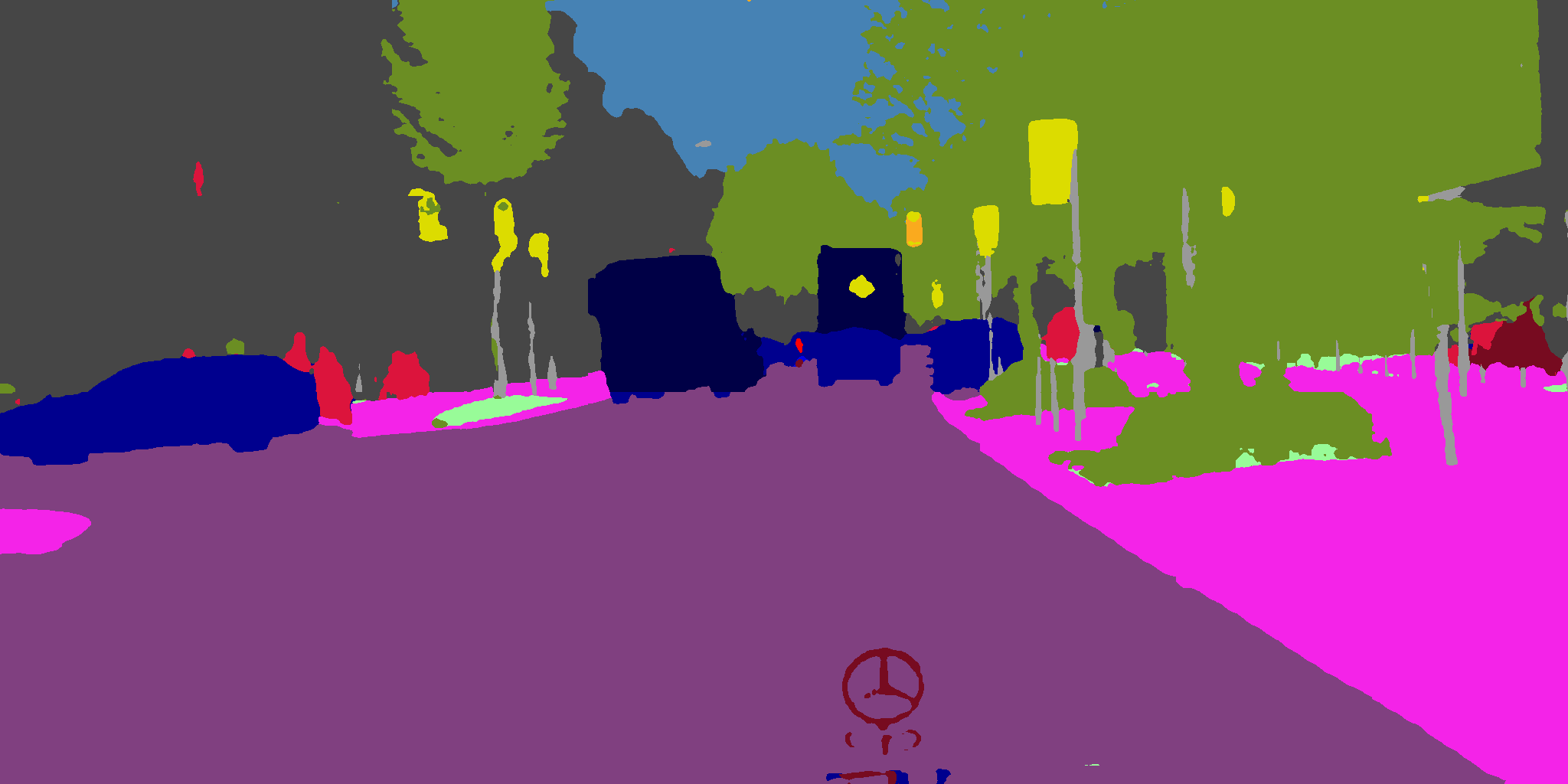}
\end{subfigure}
\end{subfigure}
\caption{Qualitative semantic segmentation results for the SELMA-to-Cityscapes adaptation task.}
\label{fig:qualitative}
\end{figure*}
}

\begin{table}[htbp]
\centering
\begin{tabular}{cccccc}
\toprule
  \multirow{2}{*}{\textbf{Dataset}}  & \multicolumn{5}{c}{$\mathbf{\beta}$}\\
  \cmidrule(lr){2-6} & 0.00 & 0.01 & 0.05 & 0.09 & 0.12 \\
 \midrule
SYNTHIA & 39.79 & 39.60 & 40.66 & \textbf{40.82} & 38.58 \\
SELMA & 38.25 & 39.52 & 39.17 & \textbf{40.80} & 38.60 \\
\bottomrule
\end{tabular}
\caption{Performance with different values of $\beta$.}
\label{tab:abl_beta}
\end{table}

\vspace{-0.1cm}

\section{Depth-dependent Entropy Loss Weighting}
Another design choice that needs to be properly evaluated is the depth-dependent weight for the entropy loss. Various models can be selected for this task, following the rationale that closer regions should get a higher weight since they have a better image resolution and depth accuracy.
We considered 4 possible options, where $d$ is the normalized disparity in the $[0,1]$ range:
\begin{enumerate}
    \item Directly using disparity $w=d$ as the weight; 
    \item Using a piecewise linear function to crop the weights range, set to $w=\frac{1}{\sqrt{10}}$ for $d<0.1$, to $w=\sqrt{10}$ for $d\geq 1$ and with a linear disparity-dependent increase in the middle;
    \item Using the square root of disparity, i.e., $w=\sqrt{d}$;
    \item Using an exponential function of the disparity value, i.e., $w=e^{d}$.
\end{enumerate}
Table \ref{tab:abl_lambda} shows how directly using the disparity as the weight led to the best performances. 
\begin{table}[ht]
\centering
\begin{tabular}{cc}
\hline
\textbf{Weighting  Scheme} %
 & \textbf{mIoU} \\
 \hline
(1) & \textbf{54.52} \\
(2) & 53.05\\
(3) & 52.93 \\
(4) & 52.80 \\
\bottomrule
\end{tabular}
\caption{Ablation on different weighting schemes.}
\label{tab:abl_lambda}
\end{table}

\section{Ablation on Self-training with depth}
As mentioned in the main paper, pseudo-label filtering has the capability to help the adaptation process considerably. In Figure \ref{fig:depth_masking}, we show that masking pixels with missing or distorted depth data improves the prediction in critical regions. 
In the first row, the depth of the car is noisy and has discontinuities determining misleading predictions in the pseudo-labels, as evidenced by the circles. Therefore, it is important to discard the wrong supervision of the labels corresponding to those regions, by masking the pseudo-labels. 
In this way, the model learns to handle the gaps coherently. Analogously, in the second row, holes in the depth are introduced by the movement of the people, and effectively masking them, makes the employed pseudo-labels more consistent.
Finally, in the last row, noisy artifacts in the depth are introduced by railway tracks on the road, compromising significantly the pseudo-labels.

Furthermore, in Table \ref{tab:ablation_self-train_depth}, a comparative evaluation is conducted between standard self-training and depth-aware self-training.  The depth usage allows for a noticeable improvement over the RGB and RGB-D self-training. In particular with the standard self-training, the approach starts the learning properly,
still, it becomes too confident about the wrong self-predictions in regions like the ones of Figure~\ref{fig:depth_masking}. %
Using instead the depth guidance the corrupted pseudo-labels are discarded and the learning continues to improve up to convergence at around $52.5\%$.

\begin{table}[ht]
   \centering
   \begin{tabular}{c|cccc|c}
   \toprule
        \textbf{Input}  & \textbf{ST} & $\textrm{\textbf{ST}}_d$  & \textbf{mIoU} \\
       \midrule
       RGB &  &  & 36.93\\
       RGB & \cmark &  & 40.59\\
       \midrule
       RGB+D &  & &  39.79\\
       RGB+D & \cmark & &  41.29\\
       RGB+D &  & \cmark &  \textbf{52.50}\\
   \bottomrule
   \end{tabular}
       \caption{Improvement of the self-training using depth data. ST corresponds to the standard self-training, while $\textrm{ST}_d$ is the depth-guided self-training.} 
   \label{tab:ablation_self-train_depth}
\end{table}

\section{Ablation on different backbones}
With the aim of testing the generalization ability of the transformer architecture, we tested MISFIT over different backbones (see Table \ref{tab:abl_back}). Even with the utilization of less complex backbones (MiT-B2 and MiT-B4), the results obtained using the model of \cite{xie2021segformer} are comparable to the state-of-the-art performance. This observation suggests that the architecture exhibits a level of robustness and efficiency, enabling competitive outcomes with lower parameter counts.

\begin{table}[ht]
    \centering
    \begin{tabular}{ccc}
        \toprule
        \textbf{Encoder} & \textbf{(M)Params.} & \textbf{mIoU} \\
        \midrule
        MiT-B2 & 27.7 & 43.2 \\
        MiT-B4 & 64.1 & 52.55 \\
        MiT-B5 & 84.7 & \textbf{54.5} \\
        \bottomrule
    \end{tabular}
    \caption{Ablation on different backbones.}
    \label{tab:abl_back}
\end{table}

\section{Additional Visual Results}
Since in the main paper visual results are shown only for the SYNTHIA-to-Cityscapes setting, in Figure \ref{fig:qualitative}, we provide some qualitative results on the SELMA-to-Cityscapes setting.
In the first row, the feature-level adaptation technique enhances the accuracy of car shapes and eliminates segmentation artifacts on the road. Furthermore, incorporating the entropy output level loss refines the shape of the sidewalk on the right side.
In the second row, in addition to similar improvements observed on the road and the car on the right, it is evident that using only input-level adaptation results in a complete failure to detect the bus in the background. In contrast, our approach properly detects it. %
In the third row, the approach is able to refine the shape of the vegetation region on the right side and removes artifacts on the road.

\end{document}